\documentclass{article}
\usepackage{PRIMEarxiv}

\usepackage[utf8]{inputenc} 
\usepackage[T1]{fontenc}    
\usepackage{url}            
\usepackage{booktabs}       
\usepackage{amsfonts}       
\usepackage{nicefrac}       
\usepackage{microtype}      
\usepackage{xcolor}         

\usepackage[colorlinks, linkcolor=orange, anchorcolor=blue, citecolor=blue]{hyperref}

\usepackage{algorithm}
\usepackage{algorithmic}

\usepackage{amsmath}
\usepackage{amssymb}
\usepackage{mathtools}
\usepackage{amsthm}

\usepackage{cleveref}

\usepackage{def}
\usepackage{color}
\usepackage{enumitem}
\usepackage{comment}
\usepackage{bm}
\usepackage{breqn}
\usepackage{array}
\usepackage{subfigure}
\usepackage{graphicx}
\usepackage{multirow}
\usepackage{wrapfig}
\usepackage{color-edits}

\usepackage{makecell}   
\usepackage{tabularx} 
\usepackage{caption}
\usepackage{subcaption}

\usepackage[round]{natbib}

\newcounter{module}

\makeatletter
\newenvironment{module}[1][htb]{%
  \refstepcounter{module}
  \renewcommand{\ALG@name}{Module}
  \begin{algorithm}[#1]%
}{%
  \end{algorithm}%
}
\makeatother

\newcounter{vfunction}

\makeatletter
\newenvironment{vfunction}[1][htb]{%
  \refstepcounter{vfunction}
  \renewcommand{\ALG@name}{Value Function}
  \begin{algorithm}[#1]%
}{%
  \end{algorithm}%
}
\makeatother

\newcounter{framework}
\makeatletter

\usepackage{xspace}
\newcommand{\ouralg}{TreeG\xspace}
\newcommand{\xtgrad}{TreeG-G\xspace}
\newcommand{\xtsampling}{TreeG-SC\xspace}
\newcommand{\xcleansampling}{TreeG-SD\xspace}

\theoremstyle{plain}

\usepackage{setspace}
\AtBeginDocument{%
  \addtolength\abovedisplayskip{-0.35\baselineskip}%
  \addtolength\belowdisplayskip{-0.35\baselineskip}%
  \addtolength\abovedisplayshortskip{-0.35\baselineskip}%
  \addtolength\belowdisplayshortskip{-0.35\baselineskip}%
}

\title{\Large \bf Training-Free Guidance Beyond Differentiability: \\
Scalable Path Steering with Tree Search in Diffusion and Flow Models}

\usepackage{authblk}

\author[1]{Yingqing Guo$^\ast$}
\author[1]{Yukang Yang$^\ast$}
\author[1]{Hui Yuan\thanks{Equal Contribution. 
Corresponding to: {\texttt {yg6736@princeton.edu}}.} $\,$}
\author[1]{Mengdi Wang}

\affil[1]{
Princeton University
}

\begin{document}
\maketitle

\begin{abstract}
Training-free guidance enables controlled generation in diffusion and flow models, but most methods rely on gradients and assume differentiable objectives. This work focuses on training-free guidance addressing challenges from non-differentiable objectives and discrete data distributions. We propose \textbf{\ouralg}:  \underline{Tree} Search-Based Path Steering \underline{G}uidance, applicable to both continuous and discrete settings in diffusion and flow models.
 \ouralg offers a unified framework for training-free guidance by proposing, evaluating, and selecting candidates at each step, enhanced with tree search over active paths and parallel exploration.  We comprehensively investigate the design space of \ouralg over the candidate proposal module and the evaluation function, instantiating \ouralg into three novel algorithms. Our experiments show that \ouralg consistently outperforms top guidance baselines in symbolic music generation, small molecule design, and enhancer DNA design with improvements of $29.01\%, 16.6\%,$ and $18.43\%$. Additionally, we identify an inference-time scaling law showing \ouralg's scalability in inference-time computation.

\end{abstract}
\section{Introduction}\label{sec:intro}

During the inference process of diffusion and flow models, guidance methods steer generations toward desired objectives, achieving remarkable success in vision \citep{dhariwal2021diffusion,ho2022classifier}, audio \citep{kim2022guided}, biology \citep{nisonoff2024unlocking,zhang2024generalized}, and decision making \cite{ajay2022conditional,chi2023diffusion}. 
In particular, training-free guidance offers high applicability by directly controlling the generation process with off-the-shelf objective functions without extra training \citep{song2023loss, zhao2024mitigating,bansal2023universal, he2023manifold}. Most training-free guidance methods are gradient-based, leveraging the objective’s gradient to steer inference, and thus assume the objective function is differentiable \citep{ye2024tfg, guo2024gradient}.

However, recent advances have pushed the boundaries of guided generation beyond differentiability: guidance objectives have expanded to include non-differentiable goals \citep{huang2024symbolic, ajay2022conditional}; diffusion and flow models have shown strong performance on discrete data \citep{austin2021structured, vignac2022digress}, where objectives are inherently non-differentiable unless approximated via differentiable features \citep{li2015improving, yap2011padel}. In such settings, training-free methods designed for differentiable objectives face fundamental limitations. Yet, the design space beyond differentiability remains under-explored: only a few methods exist \citep{lin2025tfgflow, huang2024symbolic}, they differ significantly and appear disconnected from prior guidance principles \citep{lin2025tfgflow, chung2022diffusion}. This underscores the need for a unified perspective and comprehensive study of guidance beyond differentiability. To this end, we propose an algorithmic framework \textbf{\ouralg}: \underline{Tree} Search-Based Path Steering \underline{G}uidance, designed for both diffusion and flow models across continuous and discrete data spaces.

\textbf{\ouralg is based on path steering guidance, providing a unified perspective on training-free guidance beyond differentiability.} Let $\boldsymbol{x}_0, \cdots, \boldsymbol{x}_t, \boldsymbol{x}_{t+\Delta t}, \cdots, \boldsymbol{x}_1$ denote the inference trajectory of a diffusion or flow model.  While gradients of the objective, when available, can offer a precise direction to steer inference at each step, an alternative is to search for a favorable path: propose multiple next state candidates $\boldsymbol{x}_{t+\Delta t}$ (via \texttt{BranchOut} module), evaluate them using a value function (denoted by $V$) that reflects the objective, and select the best candidate to proceed. This structured search enables \ouralg using only zeroth-order information, making it applicable beyond  differentiability.

\begin{figure}
    \centering
    \subfigure[Tree-Search Inference Process]{\includegraphics[width=0.525\linewidth]{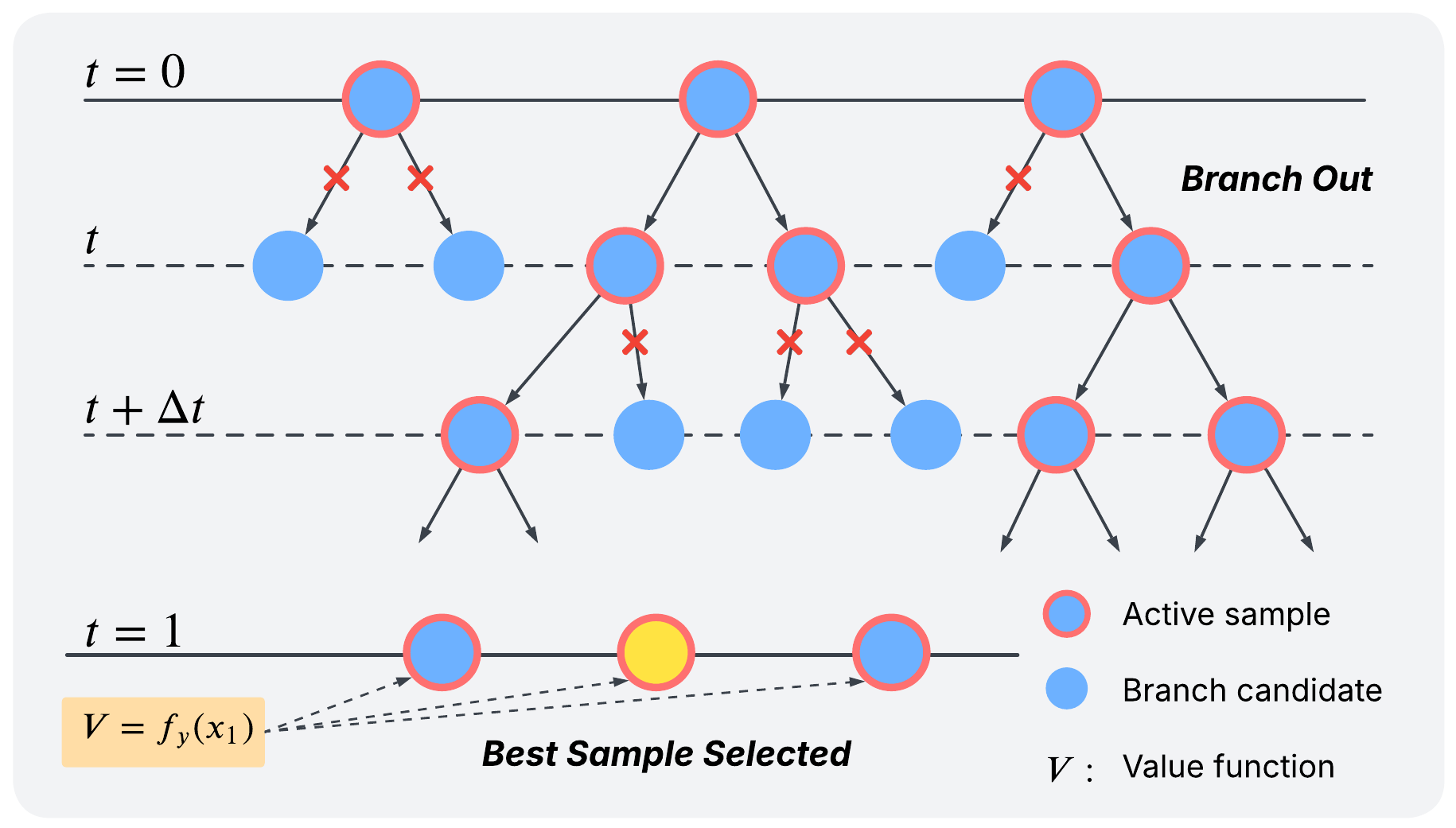}}\label{fig:tree}
    \hspace{0.002\linewidth} 
    \subfigure[ Designs for \texttt{BranchOut} and Value Function $V$, gradient-based guidance
    ]{\includegraphics[width=0.455\linewidth]{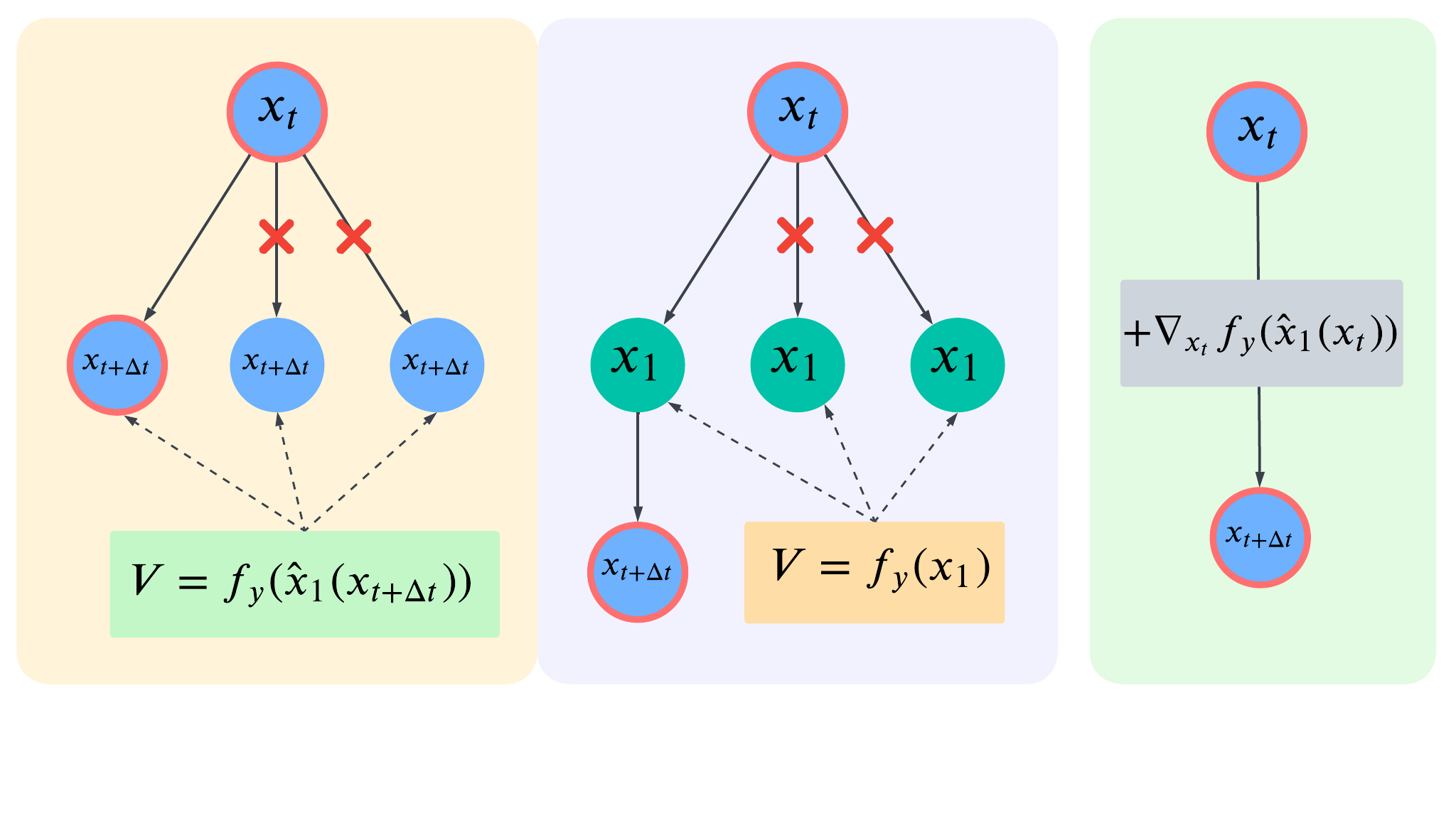}}
    \caption{\ouralg Overview: (a) An \textbf{active set} of size $A$ is maintained, where each sample branches into $K$ candidates. At each step, the top $A$ candidates are retained, and the best sample is selected at the final step. (b) Left: The \textbf{current state}-based (\texttt{BranchOut}, $V$) evaluates candidates via a lookahead estimate of the clean sample (Sec.~\ref{subsec:xtsampling}). Middle: branching and selection occur in the destination state space (Sec.~\ref{subsec:x1sampling}). Right: Gradient-based guidance can be applied to the current state
    branch-out module when a differentiable objective predictor is available (Sec.~\ref{subsec:xtgrad}).
    }
    \label{fig:inference demonstration}
    \vspace{-10pt}
\end{figure}

 \textbf{\ouralg adopts a tree-search mechanism to explore multiple trajectories, further enhancing the effectiveness of path steering guidance.} An \textbf{\textit{active set}} of size $A$ is maintained at each inference step  (Fig.~\ref{fig:inference demonstration}(a)). Each sample in the active set branches into multiple candidate next states, from which the top $A$ candidates, ranked by the value function $V$, are selected for the next step. This process iterates until the final step, where the best sample from the active set is chosen as the output. 
 By enlarging $A$, \ouralg achieves higher objective values while flexibly adapting to the computational budget.

\textbf{\ouralg's design space is over the branching-out module and value function, offering flexible and comprehensive configurations.} Effective search requires both efficient exploration and reliable evaluation. As shown in Fig.~\ref{fig:inference demonstration}(b), we propose two compatible pairs of $(\texttt{BranchOut}, V)$: one based on the current state ($\boldsymbol{x}_t$), the other on the predicted destination ($\hat{\boldsymbol{x}}_1$). The former uses the original diffusion model to generate multiple next states and evaluates them via a lookahead estimate of the clean sample. The latter generates candidate destinations, which indicate the orientation of the next state, and selects the optimal using an off-the-shelf objective function. The top-ranked destination determines the next state. In addition,  \ouralg introduces a novel gradient-based algorithm for guiding discrete flow models using gradients when a differentiable objective predictor is available.

\textbf{Contributions.} Our methodological contributions are as follows:
\begin{itemize}[itemsep=0pt, parsep=0pt, topsep=0pt, leftmargin=5pt]
\item We propose a novel tree search framework \ouralg of training-free guidance. It applies to both continuous and discrete, diffusion and flow models (Sec.~\ref{sec:propose sps}), and supports non-differentiable objectives.
\item We instantiate three novel algorithms within this framework, and provide theoretical guarantees showing that they recover the posterior conditional distribution (Sec.~\ref{sec: design space}).
\item We show that existing sampling-based methods \citep{huang2024symbolic,li2024derivative} are special cases of \ouralg (Sec.~\ref{subsec:xtsampling}), limited to exploration at the current state. Our novel exploration strategy at destination states, combined with a comprehensive design space for candidate proposals and value functions, enhances the effectiveness and versatility. 
\end{itemize}

Empirically, we demonstrate that \ouralg:
\begin{itemize}[itemsep=0pt, parsep=0pt, topsep=0pt, leftmargin=5pt]
\item Outperforms existing guidance methods on diverse tasks: symbolic music generation (continuous diffusion with {\textit{non-differentiable}} objectives), molecular design, and enhancer DNA design (both on {\textit{discrete}} flow models). Path steering guidance, the special case of \ouralg with the active set size as 1, consistently outperforms the strongest guidance baseline, yielding improvements of $29.01\%, 16.6\%,$ and $18.43\%$ respectively (Sec.~\ref{sec: exp guidance results}). %
\item Exhibits an \emph{inference-time scaling law}, where performance consistently improves as inference-time computation increases by larger active set and branch-out sizes (Fig.~\ref{fig:results overview} and Sec.~\ref{sec: exp scalability}).

\begin{figure}
    \centering
    \subfigure[Symbolic Music (Loss $\downarrow$)]{\includegraphics[width=0.32\linewidth]{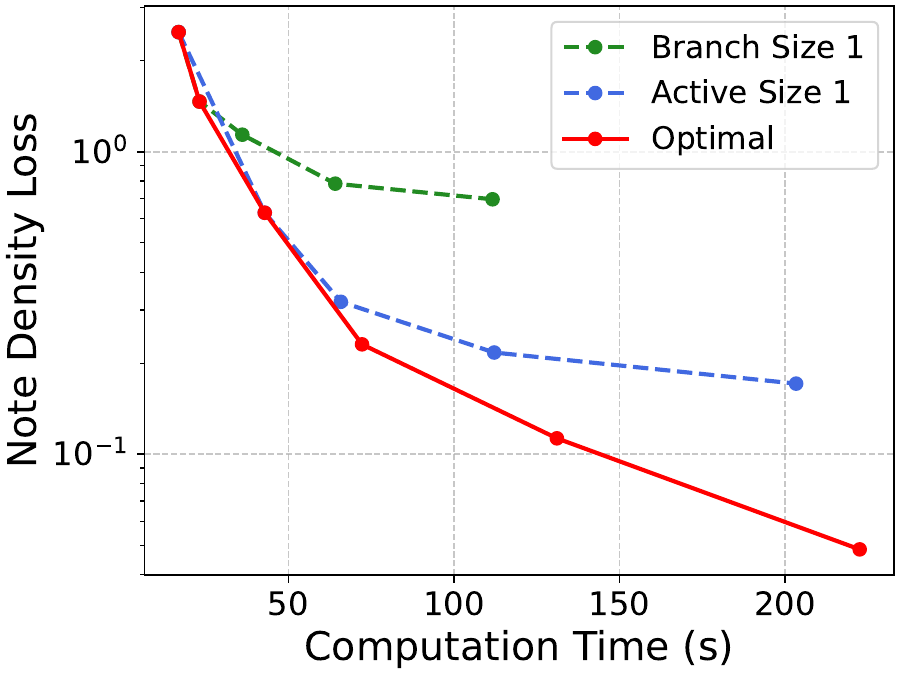}}
    \hspace{0.005\linewidth}
     \subfigure[Small Molecule  (QED $\uparrow$)]
     {\includegraphics[width=0.32\linewidth]{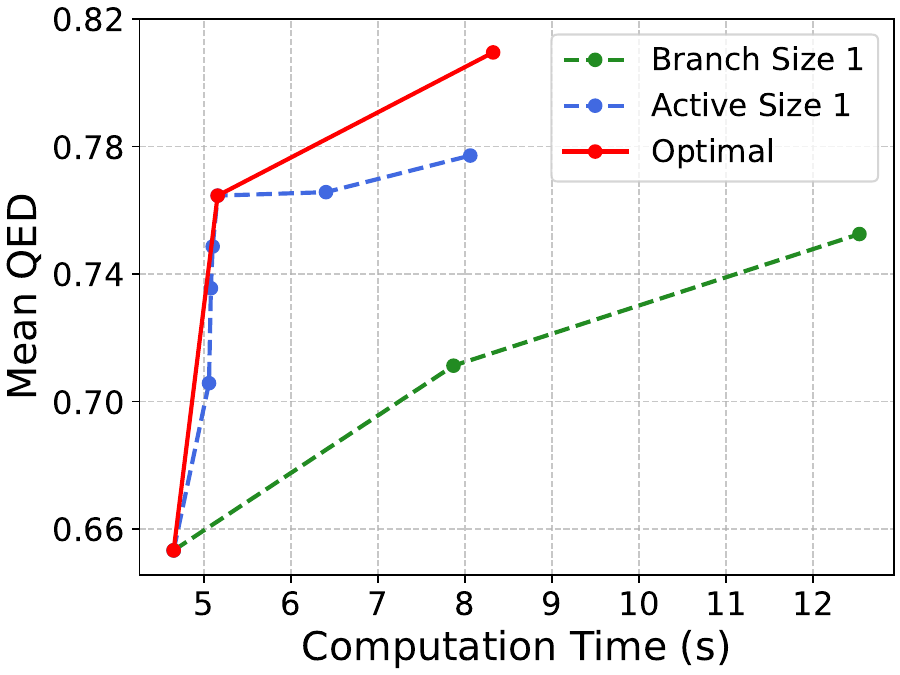}}
    \hspace{0.005\linewidth}
    \subfigure[Enhancer DNA  (Prob $\uparrow$)]{\includegraphics[width=0.32\linewidth]{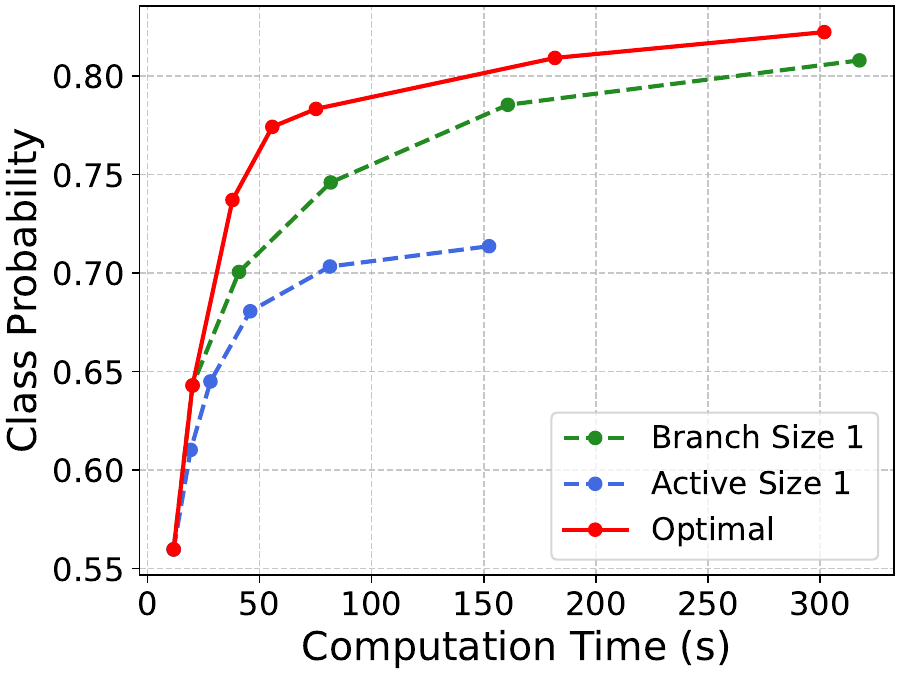}}
\caption{\textbf{Inference-Time Scaling Law of \ouralg.} ``Optimal'' refers to the best-performing combination of active set and branch-out sizes under the same inference time. The results demonstrate the scalability of \ouralg, validating the effectiveness of its multi-active-path and branch-out design.\protect\footnotemark}
    \label{fig:results overview}
\end{figure}

\footnotetext{Please refer to App.~\ref{app:exp results overview} for the experimental setup of Fig.~\ref{fig:results overview}.}

\end{itemize}

\section{Related Work}
We review the most relevant work on \emph{inference-time guidance} here; see App.~\ref{app:related work} for additional work.


\textbf{Derivative-Free.} \citet{huang2024symbolic} and \citet{li2024derivative} guide sampling toward non-differentiable objectives using soft values derived from optimal control. These are special cases of our \xtsampling with an active set size of one. Notably, our \xcleansampling (Sec.\ref{subsec:x1sampling}) introduces a novel strategy that proposes candidates at the \emph{end} of the sampling trajectory in the clean space, unexplored by existing derivative-free methods. This novel perspective, along with our flexible design for candidate proposals and value functions, underscores the broader versatility and applicability of our framework.

\textbf{On Discrete Models.} \citet{nisonoff2024unlocking} extends classifier(-free) guidance \citep{dhariwal2021diffusion,ho2022classifier} to discrete settings, requiring training time-dependent classifiers. \citet{lin2025tfgflow} reweights the rate matrix using objective values. All three of our algorithms apply to discrete models, including a gradient guidance method we propose for cases with a differentiable objective predictor.

\textbf{Inference-Time Scaling.} The idea of searching across multiple inference paths has also been touched upon by two concurrent works \citep{ma2025inference, uehara2025inference}. Our \ouralg provides the first systematic study of the design space of tree search, offering a novel methodology for both exploration and evaluation.

\section{Preliminaries}
\textbf{Notations.} Bold symbols \( \boldsymbol{x} \) denote high-dimensional vectors, while \( x \) indicates scalars. Superscripts like \( x^{(d)} \) refer to the \( d \)-th dimension, whereas \( x^i \) or \( x^{i,j} \) denote independent samples indexed by \( i \) and \( j \). \( p_t \) is the density of intermediate training distributions. For inference,  \( \mathcal{T}_t \) represents the sample.

\subsection{Diffusion and Flow Models} 
Diffusion and flow models learn to reverse a transformation from a data distribution \( p_{\text{data}} \) to a noise distribution \( p_0 \). Let \( p_1 = p_{\text{data}} \), and define intermediate distributions \( p_t \) for \( t \in (0,1) \) that progressively corrupt \( p_1 \) into \( p_0 \) over \( T \) uniform timesteps (\( \Delta t = 1/T \)). Generation starts by sampling \( \boldsymbol{x}_0 \sim p_0 \), then iteratively samples \( \boldsymbol{x}_t \sim p_t \) for \( t = i/T \), where \( i \in [T] \), ultimately yielding \( \boldsymbol{x}_1 \sim p_1 \).

Diffusion and flow models are equivalent~\citep{lipman2024flow, domingo2024adjoint}. We focus on the widely used continuous diffusion models and discrete flow models, denoted \( u_\theta^{\textit{diff}} \) and \( u_\theta^{\textit{flow}} \), and refer to both as \( u_\theta \) or diffusion models when clear from context. We next review the two models in more detail.

\textbf{Diffusion Models.} For diffusion models applied to continuous data, given a data sample $\boldsymbol{x}_1 \sim p_1 $, the noisy sample at timestep $t=i /T$ ($i \in [T]$) is constructed as $\boldsymbol{x}_t = \sqrt{\bar{\alpha}_t}\boldsymbol{x}_1 + \sqrt{1-\bar{\alpha}_t}\epsilon $, where $\epsilon \sim  \mathcal{N}\rbr{\boldsymbol{0}, \boldsymbol{I}}$ and $\cbr{\bar{\alpha}_t} $ are pre-defined monotonically increasing parameters that control the noise level. The diffusion model  $u_\theta^{\textit{diff}}: \mathcal{X}\times [0,1] \to \mathcal{X}$ parameterized by $\theta$, estimates the noise added to $\boldsymbol{x}_t$, it's equivalent to learning the score function of $p_t(\boldsymbol{x}_t)$ \citep{ho2020denoising, song2019generative}:
\begin{equation}\label{eq:diffusion train obj}
    u_\theta^{\textit{diff}} = \argmin_{u_\theta} \EE_{\boldsymbol{x}_1 \sim p_1, \epsilon \sim \mathcal{N}\rbr{\boldsymbol{0}, \boldsymbol{I}}} \norm{ u_\theta\rbr{\boldsymbol{x}_t,t}- \epsilon }^2 = - \sqrt{1-\bar{\alpha}_t} \nabla \log p_t
\end{equation} 
For sampling, we begin with $\boldsymbol{x}_0 \sim \mathcal{N}\rbr{\boldsymbol{0}, \boldsymbol{I}}$  and iteratively apply the DDPM sampling step \citep{ho2020denoising}:
\begin{equation}\label{eq:diffusion ddpm sampling}
    \boldsymbol{x}_{t+\Delta t}= \frac{1}{\sqrt{\alpha_t}} \rbr{\boldsymbol{x}_t + \rbr{1 - \alpha_t} \nabla \log p_t (\boldsymbol{x}_t)  } + \sigma_t \epsilon = \frac{1}{\sqrt{\alpha_t}} \rbr{\boldsymbol{x}_t - \frac{1 - \alpha_t}{\sqrt{1-\bar{\alpha}_t}} u_\theta^{\textit{diff}}\rbr{\boldsymbol{x}_t,t}  } + \sigma_t \epsilon,
\end{equation}
where $\epsilon \sim \mathcal{N}\rbr{\boldsymbol{0}, \boldsymbol{I}}$, $\alpha_t = \bar{\alpha}_t/\bar{\alpha}_{t+\Delta t}$ and $\sigma_t = \sqrt{1-\alpha_t}$. This step \eqref{eq:diffusion ddpm sampling} can be restated as sampling from a distribution centered on a linear interpolation of $\boldsymbol{x}_t$ and $\boldsymbol{x}_{1|t}$:
\begin{equation}\label{eq:linear interpolation sampling}
   \boldsymbol{x}_{t+\Delta t}= c_{t,1}\boldsymbol{x}_{t}+c_{t,2} \boldsymbol{x}_{1|t} + \sigma_t \epsilon,
\end{equation}
where $c_{t,1}$ and $c_{t,2}$ are constants\footnote{ We have $c_{t,1}= {\sqrt{\alpha_t}(1-\bar{\alpha}_{t+\Delta t})}/\rbr{1-\bar{\alpha}_t},  c_{t,2} = {\sqrt{\bar{\alpha}_{t+\Delta t}}(1-\alpha_t)}/\rbr{1-\bar{\alpha}_t}$.}, and with $\boldsymbol{x}_{1|t}$ being an estimation of the conditional expectation $\EE \sbr{\boldsymbol{x}_1 \mid \boldsymbol{x}_t}$ based on Tweedie’s formula \citep{efron2011tweedie}:
\begin{equation}\label{eq:pred_x1_continuous}
    \boldsymbol{x}_{1|t} := \frac{1}{\sqrt{\bar{\alpha}_t}} \rbr{\boldsymbol{x}_t - {\sqrt{1-\bar{\alpha}_t}} u_\theta^{\textit{diff}}\rbr{\boldsymbol{x}_t,t}}.
\end{equation}


\textbf{Flow Models.} For flow models applied to discrete data \citep{campbell2024generative}, suppose data space is $\mathcal{X}  = [S]^D$, where $D$ is the dimension and $S$ is the number of states per dimension.  An additional mask state $M$ is introduced as the noise prior distribution. Given a data sample $\boldsymbol{x}_1$, the intermediate distributions are constructed by $p_{t|1}(\boldsymbol{x}_t| \boldsymbol{x}_1) = \Pi_{d=1}^D p_{t|1}(x_t|x_1) $ with $p_{t|1}(x_t|x_1) = t \delta \cbr{x_1, x_t} + (1-t)\delta \cbr{M, x_t} $. The flow model $u_\theta^{\textit{flow}}$ estimates the true denoising distribution $p_{1|t}(\boldsymbol{x}_1|\boldsymbol{x}_t)$. Specifically, it's defined as $u_\theta^{\textit{flow}} = (u_\theta^{(1)},\ldots,u_\theta^{(D)})$, where each component $u_\theta^{(d)}(x_1|\cdot)$ is a function $ \mathcal{X} \times [0,1]\to \Delta([S]) $. Here, $\Delta([S])$ represents the probability distribution over the set $[S]$ The training objective is:
\begin{equation}\label{eq:flow train obj}
   u_\theta^{\textit{flow}}= \argmin_{u_\theta} \EE_{\boldsymbol{x}_1 \sim p_1, \boldsymbol{x}_t \sim p_{t|1}}\sbr{\log u_{\theta}^{(d)}(x_1^{(d)}|\boldsymbol{x}_t)}.
\end{equation}
For generation, it requires the rate matrix:
\begin{equation}\label{eq:compute rate matrix}
     R_{t}^{(d)} \rbr{\boldsymbol{x}_t, j}= \EE_{{x}_1^{(d)} \sim p_{1|t}^{(d)}}\sbr{R_t\rbr{x_t^{(d)}, j|x_1^{(d)}}} = \EE_{{x}_1^{(d)} \sim u_{\theta}^{(d)}\rbr{{x}_1\mid \boldsymbol{x}_t}}\sbr{R_t\rbr{x_t^{(d)}, j|x_1^{(d)}}},
\end{equation}
where the pre-defined conditional rate matrix can be chosen as the popular: $R_t(x_t,j|x_1) = \frac{\delta \cbr{j,x_1}}{1-t}\delta \cbr{x_t, M}$. The generation process can be simulated via Euler steps \citep{sun2022score}:
\begin{equation}\label{eq:discrete euler sampling}
    {x}_{t+\Delta t}^{(d)} \sim  \text{Cat}\rbr{\delta \{{x}_t^{(d)}, j \} + R_{t}^{(d)}\rbr{\boldsymbol{x}_t, j} \Delta t },
\end{equation}
where $ \delta\cbr{k, j}$  is the Kronecker delta which is $1$ when $k = j$ and is otherwise $0$.

\subsection{Objective}
The objective is to sample from the conditional distribution $p(\boldsymbol{x} \mid y)$, where $y$ denotes a desired property, using a pre-trained diffusion model that generates samples from unconditional distribution $p(\boldsymbol{x})$. The extent to which a sample satisfies the property $y$ is quantified by an objective function $f_y: \mathcal{X} \to \mathbb{R}$, where $f_y(\boldsymbol{x}) = \log p(y \mid \boldsymbol{x})$ and $\boldsymbol{x}$ in the clean data space.  We aim to sample from:
$$
    p(\boldsymbol{x} \mid y) \propto p(\boldsymbol{x}) \, p(y \mid \boldsymbol{x}) \propto p(\boldsymbol{x}) \, \exp(f_y(\boldsymbol{x})).
$$

Given the training objectives defined in \eqref{eq:diffusion train obj} and \eqref{eq:flow train obj} for conditional counterparts  $\boldsymbol{x}_1 \sim p_1(\boldsymbol{x} \mid y)$, the objective translates to estimating the conditional score or rate matrix: 
\begin{equation}\label{eq:objective condition score/rate}
    \nabla_{\boldsymbol{x}_t} \log p_t\rbr{\boldsymbol{x}_t \mid y}; \quad  R_{t}^{(d)} \rbr{\boldsymbol{x}_t, j \mid y}= \EE_{{x}_1^{(d)} \sim p_{1|t}^{(d)}\rbr{\boldsymbol{x}_1|y}}\sbr{R_t\rbr{x_t^{(d)}, j|x_1^{(d)}}},
\end{equation}
which enables the sampling step \eqref{eq:diffusion ddpm sampling} or \eqref{eq:discrete euler sampling} to use the conditional counterparts and generate the next state $\boldsymbol{x}_{t + \Delta t} $ following the conditional transition distribution $ \mathcal{T}_t(\boldsymbol{x}_{t + \Delta t} \mid \boldsymbol{x}_t, y)$.

We assume a \emph{perfect} pre-trained diffusion model, where $u_\theta$ exactly minimizes the training objectives in \eqref{eq:diffusion train obj} or \eqref{eq:flow train obj}. Our focus is on \emph{\textbf{training-free}} methods that neither fine-tune $u_\theta$ nor train a time-dependent classifier aligned with the diffusion noise schedule.

\section{\ouralg: \underline{Tree} Search-Based Path Steering \underline{G}uidance}
\label{sec:propose sps}
While gradients provide precise direction when available \citep{guo2024gradient}, search offers an alternative when they are not: proposing multiple candidates, evaluating them via a value function aligned with the objective, and selecting the best to proceed. We demonstrate the tree search framework in this section.

\subsection{Algorithmic Framework}
Let $\boldsymbol{x}_0, \cdots, \boldsymbol{x}_1$ denote the inference path from pure noise to a clean sample. At each step (Alg.~\ref{alg:framework}), the \texttt{BranchOut} module proposes $K$ candidate states, among which some candidates are selected into the active set. While the basic approach tracks a single path, the active set size $A$ can be increased to explore multiple paths via tree search.
The selection step can be implemented as ranking candidates by their values and selecting the top, or resampling candidates with probabilities proportional to their values. In following sections, we will show that \ouralg with resampling enjoys theoretical guarantees, while empirically, experiments (please refer to App.~\ref{app:rank resampling}) demonstrate that selection by ranking is superior to resampling.




\begin{algorithm}
\begin{algorithmic}[1]
\caption{\ouralg: {Tree} Search-Based Path Steering {G}uidance}
\label{alg:framework}
\STATE {\bf Input:} diffusion model $u_{\theta}$, branch out policy and value function $ (\texttt{BranchOut}, V)$, objective function $f_y$, active set size $A$, branch out sample size $K$.
\STATE {\bf Initialize:} $t=0$,  $\mathcal{A} = \cbr{  \boldsymbol{x}_0^{1}, \ldots, \boldsymbol{x}_0^{A} }$,  $\boldsymbol{x}_0^{i} \sim p_0$.
\WHILE{$t < 1$}
\STATE {\bf Propose candidates for next step:} For $\boldsymbol{x}_{t}^{i} \in \mathcal{A}$, 
$\boldsymbol{x}_{t+\Delta t}^{i, j} \sim  \texttt{BranchOut}\rbr{\boldsymbol{x}_t^{i}, u_\theta}$, $ j \in [K]$. 
\STATE {\bf Select:} select $A$ candidates (by ranking or resampling \footnotemark) with respect to the value function $V\rbr{\boldsymbol{x}_{t+\Delta t}^{i, j}, t, f_y}$, $i \in [A], j \in [K] $: $\boldsymbol{x}_{t+\Delta t}^{i_1, j_1},\ldots,\boldsymbol{x}_{t+\Delta t}^{i_A, j_A}$.
\STATE {\bf Update the active set:}  $ \mathcal{A} = \cbr{  \boldsymbol{x}_{t+\Delta t}^{i_1,j_1}, \ldots, \boldsymbol{x}_{t+\Delta t}^{i_A, j_A} }$
\STATE $t \leftarrow t + \Delta t $.
\ENDWHILE
\STATE {\bf Output}: $\boldsymbol{x}_1^\ast = \argmax_{\boldsymbol{x}_1 \in \mathcal{A}}f_y\rbr{\boldsymbol{x}_1}$.
\end{algorithmic}
\end{algorithm}

In Alg.\ref{alg:framework}, the \texttt{BranchOut} module and value function $V$ are two core components that require careful design, for which we propose novel designs in the next section. Those specifications of \texttt{BranchOut} and $V$ lead to new algorithms that outperform existing baselines (Sec.\ref{sec:exp}).  




\footnotetext{Given a set of candidates \( \{x_1, x_2, \ldots, x_n\} \) with associated nonnegative values \( V_1, V_2, \ldots, V_n \), ranking involves selecting the top \( A \) candidates with the highest \( V_i \) values; Resampling defines \( P(x_i) = {V_i}/\rbr{\sum_{j=1}^{n} V_j} \), and then samples \( A \) candidates according to $P(x_i)$.}


\vspace{-1pt}
\section{Design Space of \ouralg}\label{sec: design space}
\vspace{-1pt}
This section explores the design space of \ouralg, focusing on the $(\texttt{BranchOut}, V)$ pair. We introduce two compatible pairs that operate by sampling and selecting from either the current state or the predicted destination. Additionally, we present a gradient-based discrete guidance method.

\subsection{Sample-then-Select on Current States}
\label{subsec:xtsampling}
The target conditional score and rate matrix in \eqref{eq:objective condition score/rate} relate to their unconditional counterparts as follows:
\begin{equation*}
\begin{aligned}
\nabla_{\boldsymbol{x}_t} \log p_t(\boldsymbol{x}_t \mid y) &= \nabla_{\boldsymbol{x}_t} \log p_t(\boldsymbol{x}_t) + \nabla_{\boldsymbol{x}_t} \log p_t(y \mid \boldsymbol{x}_t), \quad \text{(by Bayes' rule)} \\
R_t(\boldsymbol{x}_t, \boldsymbol{x}_t^\prime \mid y) &= \frac{p_t(y \mid \boldsymbol{x}_t^\prime)}{p_t(y \mid \boldsymbol{x}_t)} \cdot R_t(\boldsymbol{x}_t, \boldsymbol{x}_t^\prime), \quad \text{(by \cite{nisonoff2024unlocking})}
\end{aligned}
\end{equation*}
Both expressions depend on $p_t(y \mid \boldsymbol{x}_t)$, which is key to adapting the unconditional score or rate matrix, estimated by a pre-trained diffusion model, to their conditional counterparts. 

The idea is to use the original backward process to generate multiple candidate states at time $t$, then prioritize samples with higher $p_t(y \mid \boldsymbol{x}_t)$ using it as a value function. We approximate $p_t(y \mid \boldsymbol{x}_t)$ as:
\begin{equation}\label{eq:prob y given xt}
     p_t(y \mid \boldsymbol{x}_t) = \EE_{\boldsymbol{x}_1 \sim p_{1|t} } p(y\mid \boldsymbol{x}_1)  = \EE_{\boldsymbol{x}_1 \sim p_{1|t} } \exp \rbr{f_y(\boldsymbol{x}_1)}\simeq \frac{1}{N}\sum_{i=1}^N \exp \rbr{f_y({\hat{\boldsymbol{x}}_1^{i}})}.
\end{equation}
Based on this, we propose the $(\texttt{BranchOut}, V)$ pair operating on current states as follows:
\begin{figure}[!htb]
\vspace{-16pt}
  \centering
  \begin{minipage}[t]{0.35\textwidth}
    \begin{module}[H]
      \centering
      \begin{algorithmic}[1]
\STATE {\bf Input:} $\boldsymbol{x}_t$, diffusion model $u_\theta$, time step $t$.
\STATE Sample the next state by the original generation process:
$\boldsymbol{x}_{t + \Delta t} \sim$ \eqref{eq:diffusion ddpm sampling} or \eqref{eq:discrete euler sampling}.
\STATE {\bf Output}: { $\boldsymbol{x}_{t + \Delta t}$}
\end{algorithmic}
\caption{\texttt{BranchOut}-Current}
\label{mol:xt_sampling}
    \end{module}
  \end{minipage}
  \hspace{0.02\textwidth}
  \begin{minipage}[t]{0.6\textwidth}
    \begin{vfunction}[H]
      \centering
   \begin{algorithmic}[1]
\STATE {\bf Input:} $\boldsymbol{x}_t$, diffusion model $u_\theta$, objective function $f_y$, time step $t$, (optional) Monte-Carlo sample size $N$.
\STATE {\bf Predict the clean sample:}\\
\quad (continuous) $\hat{\boldsymbol{x}}_1 = \boldsymbol{x}_{1|t}$ in \eqref{eq:pred_x1_continuous}.\\
\quad (discrete) $\hat{\boldsymbol{x}}_1^i \sim \text{Cat} \rbr{u_\theta( \boldsymbol{x}_{t}, t)}, i \in [N]$.
\STATE {\bf Evaluate:} $V(\boldsymbol{x}_t) = \frac{1}{N}\sum_{i=1}^N \exp \rbr{f_y({\hat{\boldsymbol{x}}_1^{i}})}$. 
\STATE {\bf Output:} { $V(\boldsymbol{x}_t)$}
\end{algorithmic}
\caption{$V:$ Current State Evaluator}
\label{vfunc:noisy}
    \end{vfunction}
  \end{minipage}
  \vspace{-9pt}
\end{figure}

We refer to instantiating Alg.~\ref{alg:framework} with Module~\ref{mol:xt_sampling} and Value Function~\ref{vfunc:noisy} as \textbf{\ouralg-Sampling Current}, abbreviated as \textbf{\xtsampling}. We demonstrate that stochastic control guidance (SCG) \cite{huang2024symbolic} and soft value decoding guidance (SVDD) \cite{li2024derivative} are special cases of \xtsampling (App.~\ref{app sec: compare to existing work}).

The following theorem (proof in App.~\ref{prof:xt_sampling}) shows that \xtsampling yields a distribution $\hat{\mathcal{T}}$ that closely approximates the target conditional transition distribution $\mathcal{T}(\cdot \mid \boldsymbol{x}_{t}, y)$.
\begin{theorem}\label{thm:xt_sampling}
Consider \ouralg-Sampling Current at time $t$, with an active set of size one and selection performed via resampling. Then, for any $\varepsilon, \delta > 0$, it holds with probability $1-\delta$:
\begin{equation*}
        \norm{\hat{\mathcal{T}} - \mathcal{T}(\cdot \mid \boldsymbol{x}_{t}, y)}_{\ell} \, < \varepsilon,
\end{equation*}
 provided one of the following conditions is satisfied:\\
\textbf{(a, Continuous)} It holds under  $\ell = 1$ norm. Suppose data follow Gaussian distribution and the objective function is linear. Branch-Out size  $K = \Theta(\frac{\log (1/\delta)}{\varepsilon^2})$ and timestep satisfies $\alpha_t =1-O\rbr{\varepsilon^2}$.\\
\textbf{(b, Discrete)} It holds under $\ell = \infty$ norm. The Branch-Out size is $K = \Theta(\frac{\log (|\mathcal{X}|/\delta)}{\varepsilon^2})$, Monte Carlo size is $N = \Theta(\frac{\log (|\mathcal{X}|/\delta)}{\varepsilon^2})$, and the timestep $\Delta t = O(\varepsilon)$, where $\mathcal{X}$ is the data space.
\end{theorem}

\vspace{-6pt}
\subsection{Sample-then-Select on Destination States}\label{subsec:x1sampling}
\vspace{-2pt}
During inference, the transition probability at each step is determined by the current state and the \emph{end} state of the path, with the latter estimated by the diffusion model, as stated in the following lemma (proof in App.~\ref{prof:transition prob}).
\vspace{-2pt}
\begin{lemma} \label{lemma:transition prob}
In both continuous and discrete cases, the transition probability during inference at timestep $t$ satisfies:
\begin{equation*}
\begin{aligned}
    \mathcal{T}(\boldsymbol{x}_{t + \Delta t} \mid \boldsymbol{x}_{t}) = \EE_{{\hat{\boldsymbol{x}}}_1} \sbr{\mathcal{T}^{\star}(\boldsymbol{x}_{t + \Delta t} \mid \boldsymbol{x}_{t}, \hat{{\boldsymbol{x}}}_1)},
\end{aligned}
\end{equation*}
where the expectation is taken over a distribution estimated by $u_\theta$, with $\mathcal{T}^{\star}$ being the true posterior distribution predetermined by the noise schedule.
\end{lemma}
\vspace{-5pt}
In diffusion and flow models, $\mathcal{T}^{\star}$ is centered at a linear interpolation between its inputs: 
the current state $\boldsymbol{x}_{t}$, and the predicted destination state $\hat{{\boldsymbol{x}}}_1$, indicating that the orientation of the next state is partially determined by $\hat{\boldsymbol{x}}_1$, as it serves as one endpoint of the interpolation. If the $\hat{\boldsymbol{x}}_1$ has a high objective value, then its corresponding next state will be more oriented to a high objective. Accordingly, we introduce the $(\texttt{BranchOut}, V)$ pair to operate on destination states as follows.
\begin{figure}[!htb]
\vspace{-16pt}
  \centering
  \begin{minipage}[t]{0.66\textwidth}
    \begin{module}[H]
      \centering
      \begin{algorithmic}[1]
\STATE {\bf Input:} $\boldsymbol{x}_t$, diffusion model $u_\theta$, time step $t$, \\(optional) tuning parameter $\rho_t, \tau_t$.
\STATE {\bf Sample destination state candidates:}
\\
\quad {(continuous)} $\hat{\boldsymbol{x}}_1 \sim \mathcal{N}(\boldsymbol{x}_{1|t}, \rho_t \boldsymbol{I})$, \, $\boldsymbol{x}_{1|t}$ in \eqref{eq:pred_x1_continuous}.\\
\quad {(discrete)} $\hat{\boldsymbol{x}}_1 \sim \text{Cat}\rbr{u_\theta( \boldsymbol{x}_{t},t)}$. \\
\STATE {\bf Compute the next state:}
\\
\quad (continuous) $\boldsymbol{x}_{t+\Delta t} \sim \mathcal{N}\rbr{c_{t,1} \boldsymbol{x}_t + c_{t,2} \hat{\boldsymbol{x}}_1, \tau_t I}$.\\
\quad(discrete) ${x}_{t+\Delta t}^{(d)} \sim \text{Cat}\rbr{\delta\{x_t^{(d)}, j \} + R_t\rbr{{x}_t^{(d)}, j \mid \hat{x}_1^{(d)} } \Delta t }$. \\
\STATE {\bf Output}: { $(\boldsymbol{x}_{t + \Delta t}, \boldsymbol{\hat{x}}_1)$}
\end{algorithmic}
\caption{\texttt{BranchOut}-Destination}
\label{mol:x1_sampling}
    \end{module}
  \end{minipage}
  \hspace{0.03\textwidth}
  \begin{minipage}[t]{0.28\textwidth}
    \begin{vfunction}[H]
      \centering
   \begin{algorithmic}[1]
\STATE {\bf Input:} $(\boldsymbol{x}_{t + \Delta t}, \boldsymbol{\hat{x}}_1)$, \\ objective function $f_y$.
\STATE Evaluate on the clean sample: \\
$V\rbr{(\boldsymbol{x}_{t + \Delta t}, \boldsymbol{\hat{x}}_1)} $ $= \exp \rbr{f_y\rbr{\boldsymbol{\hat{x}}_1}}$.
\STATE {\bf Output:} { $V\rbr{(\boldsymbol{x}_{t + \Delta t}, \boldsymbol{\hat{x}}_1)}$}
\end{algorithmic}
\caption{$V:$ \\ Destination State Evaluator}
\label{vfunc:clean}
    \end{vfunction}
  \end{minipage}
\end{figure}

\vspace{-8pt}
We name Alg.~\ref{alg:framework} with Module~\ref{mol:x1_sampling} and Value Function~\ref{vfunc:clean} by \textbf{\ouralg-Sampling Destination (\xcleansampling)}. \xcleansampling outputs $\hat{\mathcal{T}}$ that closely approximates the conditional transition distribution, as shown below (proof in App.~\ref{prof:x1_sampling}).

\vspace{-3pt}
\begin{theorem}\label{thm:x1_sampling}
Consider \ouralg-Sampling Destination at time $t$, with an active set of size one and selection performed via resampling. Then, for any $\varepsilon, \delta > 0$, it holds with probability $1-\delta$:
\begin{equation*}
        \norm{\hat{\mathcal{T}} - \mathcal{T}(\cdot \mid \boldsymbol{x}_{t}, y)}_{\ell} \, < \varepsilon,
\end{equation*}
provided one of the following conditions is satisfied:\\
\textbf{(a, Continuous)} It holds under the $\ell = 1$ norm. Suppose data follow Gaussian distribution and the objective function is linear. The Branch-Out size is $K = \Theta(\frac{\log (1/\delta)}{\varepsilon^2})$.\\
\textbf{(b, Discrete)} It holds under the $\ell = \infty$ norm. The Branch-Out size is $K = \Theta(\frac{\log (|\mathcal{X}|/\delta)}{\varepsilon^{2/D}})$  where $\mathcal{X}$ is the data space and $D$  is its dimension.
\end{theorem}

\vspace{-3pt}
\subsection{Gradient-Based Guidance with Differentiable Objective Predictor}\label{subsec:xtgrad}
\vspace{-3pt}
Previously in this section, we derived two algorithms that do not rely on the gradient of the objective. Though we do not assume the true objective is differentiable, leveraging its gradient as guidance is still a feasible option when a differentiable objective predictor is available. Therefore, in what follows, we propose a novel gradient-based training-free guidance for discrete flow models, which also fits into the \ouralg framework as a special case with $K=1$.

For discrete flow models, we aim for the conditional rate matrix \citep{nisonoff2024unlocking}:
\begin{equation*}
    R_t(\boldsymbol{x}_t, j \mid y) = \frac{p_t(y \mid \boldsymbol{x}_t^{\backslash d}(j) )}{p_t(y \mid \boldsymbol{x}_t)} \cdot R_t(\boldsymbol{x}_t, j),
\end{equation*}
where $\boldsymbol{x}_t^{\backslash d}$ matches $\boldsymbol{x}_t$  except at dimension $d$, and  $\boldsymbol{x}_t^{\backslash d}(j)$ has its $d$-dimension set to $j$. While \cite{nisonoff2024unlocking} requires training a time-dependent predictor to estimate $p_t(y \mid \boldsymbol{x})$, we propose to estimate it using \eqref{eq:prob y given xt} in a training-free way. However, computing this estimation over all possible $ \boldsymbol{x}_t^{\backslash d}$'s is computationally expensive. As suggested by \cite{nisonoff2024unlocking, vignac2022digress}, we can approximate the ratio using Taylor expansion:
\begin{equation}
\begin{aligned}
    \log \frac{p_t(y \mid \boldsymbol{x}_t^{\backslash d} )}{p_t(y \mid \boldsymbol{x}_t)} &= {\log p_t(y \mid \boldsymbol{x}_t^{\backslash d} )}-{\log p_t(y \mid \boldsymbol{x}_t)} \\
    & \simeq (\boldsymbol{x}_t^{\backslash
       d} - \boldsymbol{x}_t)^{\top} \nabla_{\boldsymbol{x}_t} \log p_t(y\mid \boldsymbol{x}_t) 
    \label{eq: taylor expansion}
\end{aligned}
\end{equation}
We use the Straight-Through Gumbel-Softmax trick \citep{jang2016categorical} to enable gradient backpropagation through the sampling process (details are in App.~\ref{appendix:imp_discret}). We also show that this approximation achieves high accuracy compared to computing \eqref{eq:prob y given xt} for all $\boldsymbol{x}_t^{\backslash d}$, while offering greater efficiency (see App.~\ref{app:ablation discrete}).

A backward sampling step, utilizing the estimated conditional rate matrix, can be viewed as a \texttt{BranchOut} operation, termed \texttt{BranchOut}-Gradient (detailed in App.~\ref{app:gradient module}). With $K=1$, \texttt{BranchOut}-Gradient reduces to gradient-based guidance methods. For $K>1$, it aligns with Value Function~\ref{vfunc:noisy}. We denote this algorithm as \textbf{\ouralg Gradient (\xtgrad)}.

\vspace{-1pt}
\section{Experiments}
\vspace{-1pt}
\label{sec:exp}
This section evaluates \ouralg on one continuous and two discrete models across diverse tasks. Sec.~\ref{sec: exp setting} introduces the comparison methods; Sec.~\ref{sec: exp guidance results} details the tasks and results; Sec.~\ref{sec: exp scalability} validates framework scalability; and Sec.~\ref{sec: exp discuss guidance design} discusses configuration choices for different scenarios.

\vspace{-1pt}
\subsection{Settings}\label{sec: exp setting}
\vspace{-1pt}
Below are the methods we would like to compare: \\
\textbf{For continuous models:}  \textbf{DPS \citep{chung2022diffusion}}, a training-free classifier gradient guidance requiring surrogate neural network predictors for non-differentiable objective functions; \textbf{TDS \citep{wu2023practical}}, a sequential Monte Carlo method based on gradient guidance; \textbf{SCG \citep{huang2024symbolic}}; \textbf{SVDD \citep{li2024derivative}}; and \textbf{\xcleansampling (Sec.~\ref{subsec:x1sampling})}.\\
\textbf{For discrete models:} \textbf{DG \cite{nisonoff2024unlocking}}, a training-based classifier guidance with a predictor trained on noisy inputs, implemented with Taylor expansion and gradients; \textbf{TFG-Flow \cite{lin2025tfgflow}}, a training-free method estimating the conditional rate matrix; \textbf{SVDD \citep{li2024derivative}}; \textbf{\xtgrad (Sec.~\ref{subsec:xtgrad})}, which trains a predictor on clean data for non-differentiable objectives; \textbf{\xtsampling (Sec.~\ref{subsec:xtsampling})}; and \textbf{\xcleansampling (Sec.~\ref{subsec:x1sampling})}.

\subsection{Guided Generation} \label{sec: exp guidance results}
For comparison to guidance baselines in this section, the active size of \ouralg is set to $A=1$. 
\subsubsection{Symbolic Music Generation} 
We follow the setup of \cite{huang2024symbolic}, using a continuous diffusion model pre-trained on several piano midi datasets, detailed in App.~\ref{app:exp detail music}. The branch-out size for \xcleansampling is $K=16$; SCG and SVDD use a sample size of $16$, the temperature of SVDD is $0.01$, and TDS uses $4$ particles.

\textbf{Guidance Target.} Our study focuses on three types of targets: pitch histogram, note density, and chord progression. The objective function is $f_{y}(\cdot) = - \ell \rbr{y, \texttt{Rule}(\cdot)}$, where $\ell$ is the loss function. Notably, the rule function $\texttt{Rule}(\cdot)$ is \textit{non-differentiable} for note density and chord progression.

\textbf{Evaluation Metrics.} For each task, we evaluate performance on 200 targets as formulated by \cite{huang2024symbolic}. Two metrics are used: (1) Loss, which measures how well the generated samples adhere to the target rules. (2) Average Overlapping Area (OA), which assesses music quality by comparing the similarity between the distributions of the generated and ground-truth music \cite{yang2020evaluation}.

\begin{table}[ht]
    \centering
     \caption{Evaluation on Music Generation. \xcleansampling reduces loss by average $29.01 \%$. Results of no guidance, Classifier, and DPS are copied from \cite{huang2024symbolic}. 
     Best results are \textbf{bold}, second-best \underline{underlined}.
    }
    \label{tab:music guidance}
    \resizebox{1\textwidth}{!}{
    \begin{tabular}{l|cccccc}
    \toprule
      \multirow{2}{*}{Method}   & \multicolumn{2}{c}{Pitch Histogram} & \multicolumn{2}{c}{Note Density} & \multicolumn{2}{c}{Chord Progression}
     \\  & Loss $\downarrow$  & OA $\uparrow$  & Loss $\downarrow$  & OA $\uparrow$ & Loss $\downarrow$  & OA $\uparrow$ \\
      \midrule
      No Guidance & $0.0180 \pm 0.0100$ & $0.842 \pm  0.012$ & $2.486 \pm 3.530$ & $0.830 \pm 0.016$ & $ 0.831 \pm 0.142$ & {$ 0.854 \pm 0.026$} \\ 
      \midrule
      Classifier & $0.0050 \pm 0.0040$ &  $0.855 \pm 0.020$ & $0.698 \pm 0.587$ & ${0.861 \pm 0.025}$ & $0.723 \pm 0.200$ &$0.850 \pm 0.033$ \\
      DPS & \underline{$0.0010 \pm 0.0020$} &  $0.849 \pm 0.018$ &  $1.261 \pm 2.340$ &$0.667 \pm 0.113$ &  $0.414 \pm 0.256$  &$0.839 \pm 0.039$  \\
      TDS & $0.0027 \pm 0.0055$ & $0.845 \pm 0.017$ & $0.218 \pm 0.241$ & $0.875 \pm 0.023$ & $0.714 \pm 0.187$ & $0.857 \pm 0.028$ \\
            SCG & $0.0036 \pm 0.0057$ & ${0.862 \pm 0.008}$ &  $\mathbf{0.134 \pm 0.533 }$  & {$0.842 \pm 0.022$} & \underline{$0.347 \pm 0.212$} & $0.850 \pm 0.046$ \\
            SVDD & $0.0085\pm0.0100$ & $0.846\pm0.013$ & $0.445 \pm 0.437$ & $0.835 \pm 0.028$ & $0.528 \pm 0.189$ & $0.865 \pm 0.014$ \\
            \midrule
    \textbf{\xcleansampling} & $\mathbf{0.0002 \pm 0.0003}$ & {$0.860 \pm 0.016$} & \underline{$0.142 \pm 0.423 $} & $0.832 \pm 0.023$ & $\mathbf{0.301 \pm 0.191}$& $ {0.856 \pm 0.032}$ \\
    \bottomrule
    \end{tabular}}
\end{table}

\textbf{Results.} As shown in Table~\ref{tab:music guidance}, our \xcleansampling outperforms baselines while preserving comparable sample quality. For differentiable rules (pitch histogram), \xcleansampling outperforms DPS, which gradient-free methods like SCG and SVDD cannot achieve. For non-differentiable rules (note density and chord progression), \xcleansampling matches or exceeds SCG and significantly outperforms others.

\subsubsection{Small Molecule Generation} 

We validate our methods on the generation of small molecules with discrete flow models. Following \cite{nisonoff2024unlocking}, the small molecules are represented as simplified molecular-input line-entry system (SMILES) strings. These \textit{discrete} sequences are padded to 100 tokens and there are 32 possible token types including one pad and one mask token $M$. We adopt the same unconditional flow model and Euler sampling curriculum as \cite{nisonoff2024unlocking}.

\textbf{Guidance Target.} 
Following the benchmarks in~\cite{gao2022sample,li2024derivative}, 
we deploy guidance to maximize three targets: Quantitative Estimate of Drug-likeness (QED), synthetic accessibility (SA), and binding score with dopamine type 2 receptor (DRD2). A pretrained target predictor $f(x)$ is directly used as the objective function. We also guide the number of rings $N_r$ towards target values $N_r^*\in[0,6]$. 
The objective function is formalized as
$f_y(\boldsymbol{x})= -\frac{(y-f(\boldsymbol{x}))^2}{2\sigma^2}$. Please refer to~\ref{app:molecule setup} for details.

\textbf{Evaluation Metrics.} We measure four molecule properties with RDKit and TDC~\cite{Huang2021tdc} for generated {valid} unique sequences. For $N_r$, we report the mean absolute error (MAE) against each target value. We evaluate molecular \emph{diversity} through average
Tanimoto similarity (TS) across molecules.


\begin{table}[ht]
    \centering
    \caption{Evaluation on Small Molecule Generation. 
    The Tanimoto similarity (TS) of unguided generated sequences is $0.12\pm0.02$. Full results of $N_r^*$ are shown in App.~\ref{app:tab2 add}.
    }
    \label{tab:smg}
    \resizebox{1.0\textwidth}{!}{
    \begin{tabular}{l|ccccccccccccccccc}
    \toprule
       \multirow{2}{*}{Method}  & 
       & \multicolumn{2}{c}{QED}  
       & \multicolumn{2}{c}{SA} 
       & \multicolumn{2}{c}{DRD2}  
       & \multicolumn{2}{c}{$\mathrm{N_r}^*=1$}  
         \\
         & & VAL $\uparrow$ & TS $\downarrow$ 
        & VAL $\uparrow$ & TS $\downarrow$  & VAL $\uparrow$ & TS $\downarrow$  & MAE $\downarrow$ & TS $\downarrow$ \\
         \midrule
          No Guidance & 
         & $0.61\pm0.19$ & \rule{0.4cm}{0.2mm} %
         & $0.79\pm0.10$ & \rule{0.4cm}{0.2mm} %
        & $0.06\pm0.17$ & \rule{0.4cm}{0.2mm} %
        & $2.09\pm1.16$ & \rule{0.4cm}{0.2mm} %
         \\ 
         \midrule
          DG & 
         & $0.62\pm0.20$ & 
          $0.12\pm0.02$
         & $0.80\pm0.10$ & 
         $0.12\pm0.02$
         & $0.17\pm0.32$ & 
         $0.12\pm0.02$
         & \underline{$0.11\pm0.33$} & 
        $0.14\pm0.03$
         \\
         TFG-Flow & 
         & $0.61\pm0.20$ & 
          $0.12\pm0.02$
         & $0.79\pm0.10$ & 
          $0.12\pm0.02$
         & $0.09\pm0.22$ & 
          $0.12\pm0.02$
         & $0.28\pm0.65$ & $0.13\pm0.02$

         \\
         SVDD & 
         & {$0.71\pm0.17$}
         & $0.13\pm0.02$
         & {$0.81\pm0.09$} & 
         $0.13\pm0.02$
         & \underline{$0.64\pm0.42$} & 
         $0.17\pm0.04$
         & 
         $0.35\pm1.14$ 
         & $0.14\pm0.03$ 
\\
           \midrule

       \textbf{\xtsampling} ($K=4$) & 
     & $\textbf{0.79}\pm\textbf{0.12}$ & $0.12\pm0.02$ 
     & $\textbf{0.90}\pm\textbf{0.07}$ & $0.21\pm0.05$ 
     & $\textbf{0.77}\pm\textbf{0.33}$ & $0.20\pm0.04$ 
     & 
     $\textbf{0.01}\pm\textbf{0.07}$ 
     & $0.14\pm0.02$
     \\
    \textbf{\xcleansampling} ($K=200$) & 
     & \underline{$0.77\pm0.14$} & $0.12\pm0.02$ 
     & \underline{$0.86\pm0.10$} & $0.16\pm0.04$ 
     & $0.45\pm0.41$ & $0.14\pm0.03$ 
     &  $0.11\pm0.38$ & $0.12\pm0.02$
     \\
     
    \textbf{\xtgrad} & 
     & $0.64\pm0.19$ & $0.12\pm0.02$ 
     & $0.79\pm0.10$ & $0.12\pm0.02$ 
     & $0.22\pm0.37$ & $0.13\pm0.02$ 
     &  $0.44\pm1.19$ 
     & 
    $0.13\pm0.02$       
     \\
    \bottomrule
 
    \end{tabular}}
\end{table}

\textbf{Results.} 
\ouralg consistently outperforms guidance baselines DG and TFG-Flow (Tab.~\ref{tab:smg}). Our best guidance method \xtsampling beats SVDD and achieves a $16.6\%$ average improvement over four targets compared to the best baseline methods (Tab.~\ref{tab:smg_rel}).

\subsubsection{Enhancer DNA Design}
We follow the experimental setup of \cite{stark2024dirichlet}, using a discrete flow model pre-trained on DNA sequences of length 500, each labeled with one of 81 cell types \cite{janssens2022decoding, taskiran2024cell}. For inference, we apply 100 Euler sampling steps. The branch-out size for \xtgrad is set $K=1$. Setup details are in App.~\ref{app sec: setup dna}.


\textbf{Guidance Target.} The goal is to generate enhancer DNA sequences that belong to a specific target cell type. The guidance target predictor is provided by an oracle classifier $f$ from \cite{stark2024dirichlet}. The objective function of given cell class $y$ is $f_y(\cdot) = \log  f(y\mid \cdot)$. 
\begin{table}[ht]
    \centering
     \caption{Evaluation of guidance methods for Enhancer DNA Design at varying guidance strength levels $\gamma_t = \gamma$ in Module~\ref{mol:grad guidance}. \xtgrad consistently achieves significantly higher target class probabilities.}
    \label{tab:enhancer guidance res}
    \resizebox{1\textwidth}{!}{
    \begin{tabular}{lc|ccccccccc}
    \toprule
         \multicolumn{2}{c|}{\multirow{2}{*}{Method  { \small (strength $\gamma$)}} }& \multicolumn{3}{c}{Class 1} & \multicolumn{3}{c}{Class 2} & \multicolumn{3}{c}{Class 3}\\
   & & Prob  $\uparrow$ & FBD  $\downarrow$ & Div  $\uparrow$ & Prob  $\uparrow$ & FBD  $\downarrow$ & Div  $\uparrow$  & Prob  $\uparrow$ & FBD  $\downarrow$ & Div  $\uparrow$ \\
   \midrule
    No Guidance & \rule{0.4cm}{0.2mm} & $0.021 \pm 0.079$ & $242$ & $373$ & $0.008 \pm 0.052$ & $602$ &$373$ & $0.007 \pm 0.053$ & $910$ & $373$\\
         \midrule
            \multirow{3}{*}{DG} &  {\footnotesize $20$} &  $0.359 \pm 0.188$ & {${109}$} & $351$  & $ 0.627 \pm 0.340$ & ${120}$ & $359$ & $0.693 \pm 0.264$ & ${102}$ & $357$ \\
          & {\footnotesize $100$} & $0.372 \pm 0.237$ & $116$ & $352$ & $0.571 \pm 0.356$ & {$212$} & $343$ & $0.173 \pm 0.256$ & $347$ & $349$ \\
          & {\footnotesize $200$} & $0.251 \pm 0.171$ & $157$ & $331$ & $0.350 \pm 0.351$ & $294$ & $335$ & $0.064 \pm 0.143$ & $514$ & $343$ \\
         \midrule          
           \multirow{1}{*}{TFG-Flow} &  {\footnotesize $200$}  & $0.054 \pm 0.129$ & $151$ & $375$ & $0.012 \pm 0.076$ & $352$ & $375$ & $0.004 \pm 0.029$ & $617$ & $375$ \\ \midrule
           \multirow{1}{*}{SVDD} &  \rule{0.4cm}{0.2mm}  & $0.155 \pm 0.173$ & $101$ & $364$ & $0.088 \pm 0.193$ & $364$ & $367$ & $0.061 \pm 0.168$ & $504$ & $363$ \\ \midrule
           \multirow{3}{*}{\textbf{\xtgrad}} &  {\footnotesize $ 20$} &  $0.236 \pm 0.178$ & {$111$} & $355$  & $0.313 \pm 0.343$ & $279$ & $347$ & $0.247 \pm 0.280$ & $314$ & $324$ \\
           & {\footnotesize $ 100$} & \underline{$0.509 \pm 0.242$} & $189$ & $353$ & \underline{$0.915 \pm 0.154$} & $318$ & $332$ & \underline{$0.745 \pm 0.217$} & {$125$} & $306$ \\
          & {\footnotesize $200$} & $\mathbf{0.560 \pm 0.258}$ & $179$ & $358$ & $\mathbf{0.951 \pm 0.110}$ & $337$& $331$ & $\mathbf{0.894 \pm 0.136}$ & $213$ & $321$ \\    
         \bottomrule
    \end{tabular}}
\end{table}

\textbf{Evaluation Metrics.} We generate 1000 DNA sequences conditioned on cell type and evaluate performance using three metrics. Target Class Probability, provided by the oracle classifier, where higher probabilities indicate better guidance. Frechet Biological Distance (FBD) measures the distributional similarity between the generated samples and the training data for a specific target class. Diversity is measured by the average pairwise Hamming distance between sequences.


\textbf{Results.} As shown in Tab.~\ref{tab:enhancer guidance res}, our  \xtgrad consistently achieves the highest target probabilities as guidance strength increases, with an average improvement of $18.43\%$ compared to DG over eight test classes \citep{nisonoff2024unlocking}, and significantly exceeding other training-free baselines. See App.~\ref{app:additional res} for details.



\subsection{Scalability on Inference-Time Computation} \label{sec: exp scalability}

\begin{figure}
    \centering
    \subfigure[ND Music]{\includegraphics[width=0.245\linewidth]{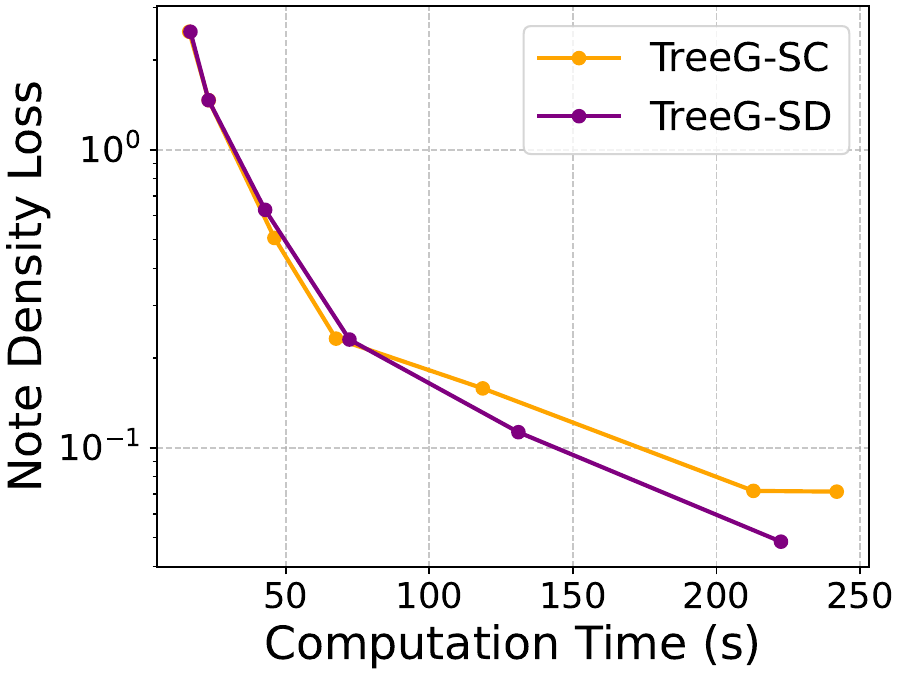}}
    \subfigure[PH  Music]{\includegraphics[width=0.245\linewidth]{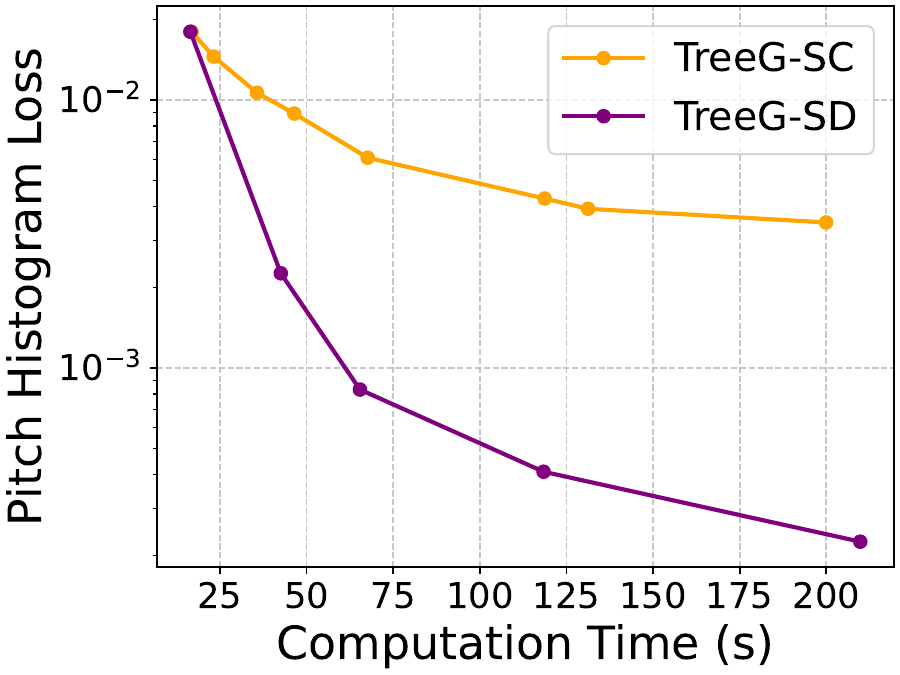}}
    \subfigure[Molecular ($N_r^*=5$)]{\includegraphics[width=0.245\linewidth]{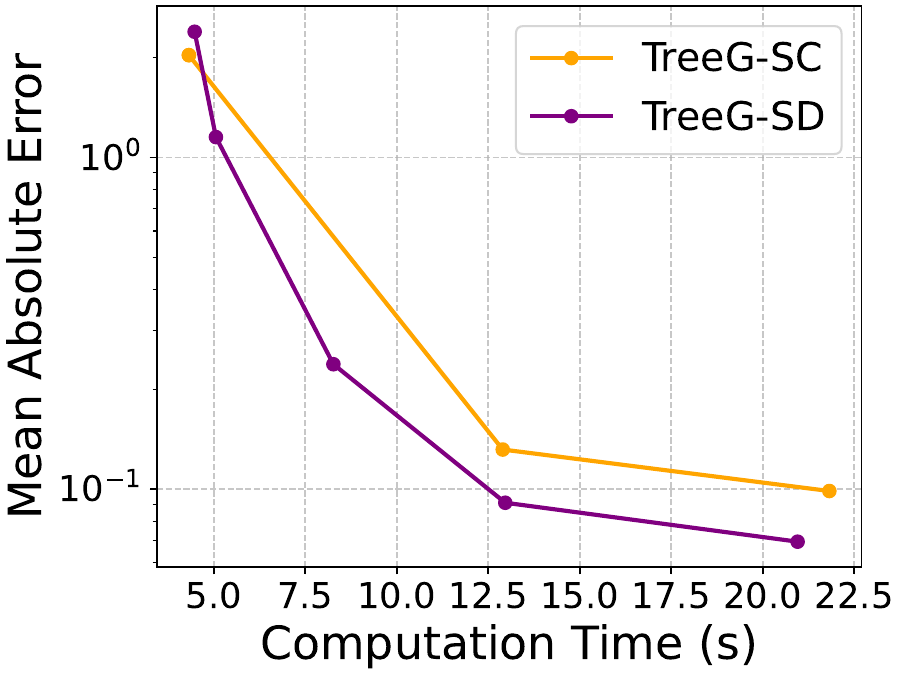}}
    \subfigure[Enhancer (Class 4)]{\includegraphics[width=0.245\linewidth]{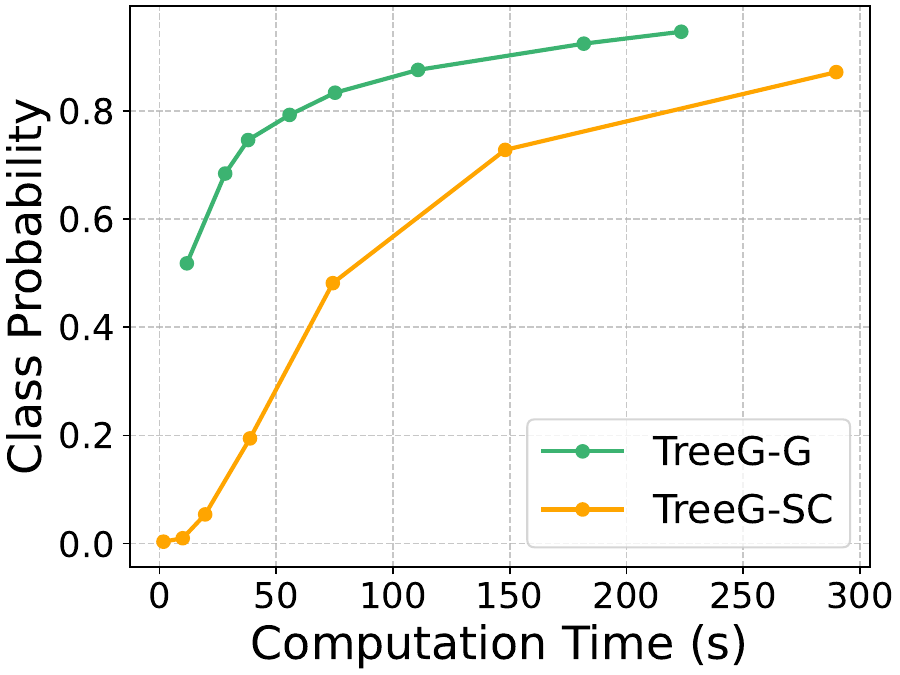}}
    \vspace{-7pt}
    \caption{\textbf{Inference Time Scaling Behavior:} As the active set size and branch-out size increase, the optimization effect of the objective function scales with inference time. This trend is consistently observed across all algorithms and tasks. The inference time is measured with a batch size of 1 for music and 100 for molecule and DNA design. For DNA design, $\gamma=20$ for \xtgrad.}
    \label{fig:scaling law}
\end{figure}


\ouralg is scalable to the active size $A$ (i.e., the number of generation paths) and the branch-out size $K$. It is compatible with all guidance methods.  When increasing the active set size and branch-out size, the computational cost of inference rises. We investigate the performance frontier to optimize the objective function concerning inference time. The results reveal an inference-time scaling law, as illustrated in Fig.~\ref{fig:scaling law}. Our findings indicate consistent scalability across all algorithms and tasks, with App.~\ref{fig:scaling law} showcasing four examples. Additional results refer to App.~\ref{app:additional res}.

\subsection{\ouralg Configuration Analysis}\label{sec: exp discuss guidance design}
This section analyzes the configuration of \ouralg, focusing on selecting the instantiated algorithms and balancing $A$ and $K$ under a fixed computational budget.

\begin{wraptable}{r}{0.47\textwidth}  
  \centering
  \small  
  \captionsetup{font=small}
  \caption{Computation Complexity of \ouralg}
     \label{tab:computation complexity}
     \vspace{-2pt}
    \begin{tabular}{lc}
    \toprule
   Methods & Computation \\
    \midrule
    \xtsampling  & $AC_{\text{model}}+AK(C_{\text{model}}+ NC_{\text{pred}})$  \\
    \xcleansampling & $AC_{\text{model}}+AKC_{\text{pred}}$\\
    \xtgrad & $ AK\rbr{C_{\text{model}}+ NC_{\text{pred}}}+AC_{\text{backprop}}$ \\
    \bottomrule
    \end{tabular}
\vspace{-5pt}
\end{wraptable}
We analyze the computational complexity of \ouralg, summarized in Tab.~\ref{tab:computation complexity}, The cost units are: $C_{\text{model}}$ for a forward pass through the diffusion model, $C_{\text{pred}}$ for a predictor call, and $C_{\text{backprop}}$ for backpropagation through both. $N$ denotes the Monte Carlo sample size used in Value Function~\ref{vfunc:noisy}.



\textbf{Design Axes Comparison.} We evaluate \ouralg designs along two axes. \textbf{Gradient-free vs. gradient-based:} \xtgrad requires an accurate objective predictor to be effective. If available, predictor latency guides the choice—faster predictors favor \xtsampling and \xcleansampling, while slower ones make \xtgrad more practical. \textbf{\xtsampling (current state) vs. \xcleansampling (destination state):} As Tab.~\ref{tab:computation complexity} shows, \xcleansampling is more efficient, costing only $A$ for the diffusion model forward pass, versus $AK$ for alternatives. Experiments show \xcleansampling outperforms \xtsampling in continuous diffusion, while \xtsampling is better for discrete flow. Please refer to App.~\ref{app: exp discuss guidance design} for detailed discussion.

\begin{wrapfigure}{r}{0.32\textwidth}
\centering
\vspace{-5pt}
\includegraphics[width=0.31\textwidth]{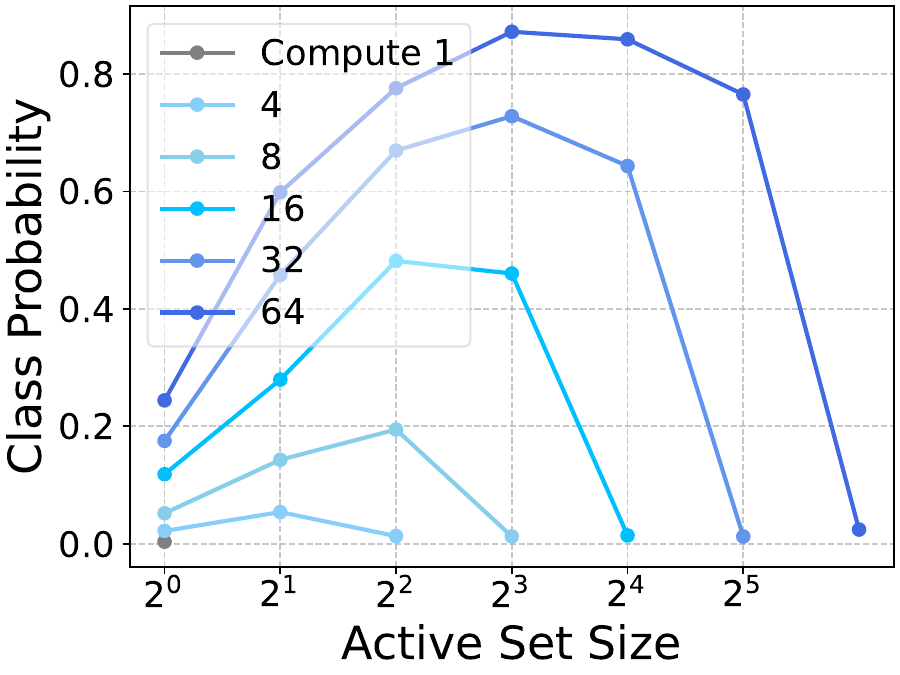}
\captionsetup{font=small}
\caption{Trade-off between Active Set Size $A$ and Branch-out Size $K$ with Fixed Compute. The results are for \xtsampling DNA (Class 4). }
\vspace{-23pt}
\label{fig:trade-off A K}
\end{wrapfigure}

\textbf{Trade-off between Active Set Size $A$ and  Branch-out Size $K$.} Tab.~\ref{tab:computation complexity} shows that the computational complexity of \xtsampling and \xtgrad using \texttt{BranchOut}-Current is $O(AK)$. With a fixed product $A \cdot K$ (i.e., fixed inference cost; see Fig.~\ref{fig:time with fixed AK} in App.~\ref{app:scalability of DNA}), we explore how to best balance $A$ and $K$. As shown in Fig.\ref{fig:trade-off A K}, performance peaks when both values are moderate. In the special case where  $K\!=\!1$
 inference paths do not interact, often leading to suboptimal performance.

\section{Conclusion}
We proposed the framework \ouralg based on inference path search, along with three novel instantiated algorithms: \xtsampling, \xcleansampling, and \xtgrad, which guide diffusion models toward the posterior conditional distribution and address the non-differentiability challenge in training-free guidance. Experimental results demonstrated the improvements of \ouralg against existing methods. Furthermore, we identified an inference-time scaling law that highlights \ouralg's scalability in inference-time computation.


\clearpage
\appendix

\section{Additional Related Work}\label{app:related work}
In this section, we provide an overview of related work on aligning diffusion models with downstream objectives. The methods in this domain can be broadly categorized into inference-time alignment and post-training/fine-tuning approaches. We begin by reviewing these two main categories, followed by a discussion of other relevant methods, including alternative alignment techniques for diffusion models and discrete diffusion and flow models.

\textbf{Inference-Time Alignment.} Classifier guidance uses a time-dependent classifier to provide directional signals throughout the generation process \cite{song2020score, dhariwal2021diffusion}. A growing body of work explores training-free guidance methods that leverage gradients derived from objective functions \cite{chung2022diffusion, bansal2023universal, ye2024tfg, song2023loss, he2023manifold, yu2023freedom,shen2024understanding}. A line of work \cite{wu2023practical, phillips2024particle, dou2024diffusion, cardoso2023monte,skreta2025feynman} combined diffusion models with Sequential Monte Carlo methods. \cite{li2024derivative, huang2024symbolic} use value-based selection and importance sampling When objectives are non-differentiable. A concurrent work \cite{singhal2025general} proposes an SMC-based framework for inference-time scaling using value resampling. A widely used and straightforward approach is Best-of-N sampling, where the model generates multiple candidates and selects those with the highest objective values \cite{stiennon2020learning, nakano2021webgpt, beirami2024theoretical}.

\textbf{Fine-tuning Diffusion Models.} Fine-tuning diffusion models involves directly adjusting model parameters to align with downstream objectives. One common method involves direct fine-tuning through gradient backpropagation across the sampling process \cite{black2023training,uehara2024fine,prabhudesai2023aligning}. To ensure stable training and prevent divergence from the original distribution, Kullback-Leibler (KL) regularization has been introduced in \cite{uehara2024fine, domingo2024adjoint}. Reinforcement learning has also emerged as a powerful tool for fine-tuning \cite{black2023training, fan2023reinforcement, zhao2024adding, uehara2024understanding}. \cite{wallace2024diffusion} proposes direct preference optimization for aligning diffusion models with human preferences. Additionally, recent work \cite{wang2024fine} has extended fine-tuning techniques to discrete diffusion models.

\textbf{Other Diffusion Alignment Methods.} Beyond guidance and fine-tuning approaches, an alternative line of training-free methods focuses on optimizing the initial latent state of the reverse diffusion process \citep{wallace2023end, ben2024d, karunratanakul2024optimizing}.  These methods typically use an ODE solver to backpropagate the objective gradient directly to the initial latent state, making them a gradient-based version of the Best-of-N strategy. Additionally, adjoint-based methods have been proposed for gradient estimation in diffusion models \cite{marion2024implicit, pan2023adjointdpm, blasingame2024adjointdeis}. Other recent work explores diffusion-based alignment techniques and applications of diffusion models in biology \cite{watson2023novo, avdeyev2023dirichlet, shen2024non, jolicoeur2309generating}.



\textbf{Discrete Diffusion and Flow Models.} \citet{austin2021structured} and \citet{hoogeboom2021argmax} pioneered diffusion in discrete spaces by introducing a corruption process for categorical data. \citet{campbell2022continuous} extended discrete diffusion models to continuous time, while \citet{lou2023discrete} proposed learning probability ratios. Discrete Flow Matching \citet{campbell2024generative,gat2024discrete} further advances this field by developing a Flow Matching algorithm for time-continuous Markov processes on discrete state spaces, commonly known as Continuous-Time Markov Chains (CTMCs). \citet{lipman2024flow} presents a unified perspective on flow and diffusion.

\section{Limitations and Broader Impacts}\label{app sec:impacts}

\textbf{Limitations.} We do not observe significant limitations in our methods. However, in cases where the objective function is non-differentiable, very expensive to evaluate, and lacks an effective surrogate neural network, our methods can become relatively time-consuming. These represent inherently difficult scenarios, where a trade-off between computational cost and optimization effectiveness is unavoidable.

\textbf{Broader Impacts.} This paper aims to explore inference-time alignment methods for diffusion models, advanced control techniques in generative AI, and contribute to the broader field of artificial intelligence. This work holds promise for improving the accuracy and personalization of AI systems in diverse domains, such as image synthesis and drug discovery. Nonetheless, the same techniques may also be exploited to produce harmful content.

\section{Discussion on Design Axes} \label{app: exp discuss guidance design}
We compare the guidance designs: \xtsampling, \xcleansampling and \xtgrad based on experimental results, to separate the effect of guidance design from the effect of tree search, we set $A=1$. A side-to-side comparison on the performance of the three methods are provided in \cref{tab:discrete internal comparison}.


\paragraph{Gradient-based v.s. Gradient-free: depends on the predictor.} The choice between gradient-based and gradient-free methods largely depends on the characteristics of the predictor. 

The first step is determining whether a reliable, differentiable predictor is available. If not, sampling methods should be chosen over gradient-based approaches. For example, in the chord progression task of music generation, the ground truth reward is obtained from a chord analysis tool in the music21 package \cite{cuthbert2010music21}, which is non-differentiable. Additionally, the surrogate neural network predictor achieves only $33\%$ accuracy \cite{huang2024symbolic}. As shown in \cref{tab:music guidance}, in cases where no effective differentiable predictor exists, the performance of gradient-based methods (e.g., DPS) is significantly inferior to sampling-based methods (e.g., \xcleansampling and SCG).

If a good differentiable predictor is available, the choice depends on the predictor's forward pass time. Our experimental tasks illustrate two typical cases: In molecule generation, where forward passes are fast as shown in \cref{tab:basic time}, sampling approaches efficiently expand the candidate set and capture the reward signal, yielding strong results (\cref{tab:discrete internal comparison}). In contrast, for enhancer DNA design, where predictors have slow forward passes, increasing the sampling candidate set size to capture the reward signal becomes prohibitively time-consuming, making gradient-based method more effective (\cref{tab:discrete internal comparison}).

\paragraph{\xtsampling v.s. \xcleansampling.} Experiments on continuous data and discrete data give divergent results along this axis. In the continuous task of music generation (\cref{tab:music guidance}), \xcleansampling achieves equal or better performance than SCG (equivalent to \xtsampling) with the same candidate size $K=16$ and similar time cost (details in \cref{app:additional music}). Thus, \xcleansampling is preferable in this continuous setting. Conversely, for discrete tasks, \xcleansampling requires significantly more samples, while \xtsampling outperforms it, as shown in \cref{tab:discrete internal comparison}.

\begin{table}[ht]
    \centering
    \begin{minipage}[t]{0.75\textwidth}
    \caption{Comparison of results across \xtgrad, \xtsampling and \xcleansampling. For molecule generation, the target is specified as the number of rings $N_r = 2$. For enhancer DNA design, the results correspond to Class 3.}
    \label{tab:discrete internal comparison}

    \resizebox{0.88\textwidth}{!}{
    \begin{tabular}{lcccc}
    \toprule
       &  & \xtgrad & \xtsampling &  \xcleansampling  \\
       \midrule 
       \multirow{5}{*}{\rotatebox{90}{Molecule}}  
       & MAE $\downarrow$ & $0.09\pm0.54$ & $0.02\pm0.14$ & $0.10\pm0.33$  \\
         & Time $\downarrow$ & 13.5s & 12.9s & 11.2s \\
          & $N$ & 30 & 30 & \rule{0.35cm}{0.2mm} \\
           &  $K$ & \rule{0.35cm}{0.2mm} & 2 & 200 \\
    \midrule 
       \multirow{6}{*}{\rotatebox{90}{Enhancer}}  
       & Prob $\uparrow$ 
       & $0.89 \pm 0.14$ & $0.13 \pm 0.39$ & $0.002 \pm 0.000$ \\
        & FBD $\downarrow$ & $213$ & $384$ & $665$\\
         & Div $\uparrow$ & $321$ & $375$ & $376$ \\
         & Time $\downarrow$ & 10.3s & 285.1s & 189.7s \\
          & $N$ & 20  & 20 & \rule{0.35cm}{0.2mm}   \\
           &  $K$ & \rule{0.35cm}{0.2mm} & 64 & 1024 \\
    \bottomrule
    \end{tabular}
    }

    \end{minipage}
    
\end{table}

\begin{table}[ht]
    \centering
    \caption{Computation time per basic unit (ms)}
    \label{tab:basic time}
    \begin{tabular}{cccc}
    \toprule
         & $C_{\text{model}}$ & $C_{\text{pred}}$ & $C_{\text{backprop}}$ \\
         \midrule
       Molecule  & 0.038  &  2.2e-4 & 0.036 \\
       Enhancer & 0.087 & 0.021 & 0.11 \\
       \bottomrule
    \end{tabular}
\end{table}

\section{Omitted Proofs in \cref{sec: design space}}
\subsection{Proof of \cref{thm:xt_sampling}}\label{prof:xt_sampling}
We begin by restating \cref{thm:xt_sampling} with all conditions for completeness, followed by its proof.
\begin{theorem*}[Restatement of \cref{thm:xt_sampling}]
Consider \ouralg-Sampling Current at time $t$, with an active set of size one and selection performed via multinomial resampling. Then, for any $\varepsilon, \delta > 0$, it holds with probability $1-\delta$:
\begin{equation*}
        \norm{\hat{\mathcal{T}} - \mathcal{T}(\cdot \mid \boldsymbol{x}_{t}, y)}_{\ell} \, < \varepsilon,
\end{equation*}
provided one of the following conditions is satisfied:\\
\textbf{(a, Continuous)}It holds under the $\ell = 1$ norm. Suppose data follow Gaussian distribution and the objective function is linear. The Branch-Out size is $K = \Theta(\frac{\log (1/\delta)}{\varepsilon^2})$ and the timestep is sufficiently small such that $\alpha_t =1-O\rbr{\varepsilon^2}$.\\
\textbf{(b, Discrete)}It holds under the $\ell = \infty$ norm. The mapping $t \mapsto p_{1|t}(\boldsymbol{x}_1 \mid \boldsymbol{x})$ is $L$-Lipschitz continuous for all $\boldsymbol{x}$ and $\boldsymbol{x}_1$, and the likelihood scores satisfy $0 < B_{\min} \le \exp(f_y(\boldsymbol{x})) \le B_{\max}$ for all $\boldsymbol{x}$. Branch-Out size is $K = \Theta(\frac{\log (|\mathcal{X}|/\delta)}{\varepsilon^2})$, Monte Carlo size is $N = \Theta(\frac{\log (|\mathcal{X}|/\delta)}{\varepsilon^2})$, and time step $\Delta t = O( \varepsilon)$,  where $\mathcal{X}$ is the data space. 
\end{theorem*}

\begin{proof}

\textbf{Proof of (a) continuous case:}
Assume the data is drawn from a Gaussian, and the objective function is linear
\[
\boldsymbol{x}_1 \sim \mathcal{N}(\boldsymbol\mu, I),
\qquad
f_y(\boldsymbol{x})=\boldsymbol{g}^\top\boldsymbol{x}.
\]
Since $\boldsymbol{x}_t \mid \boldsymbol{x}_1 \sim \mathcal{N}\bigl(\sqrt{\bar\alpha_t}\,\boldsymbol{x}_1,\;(1-\bar\alpha_t)\,I\bigr)$, by Bayes’ rule the posterior follows:
\[
\boldsymbol{x}_1 \mid \boldsymbol{x}_t
\sim
\mathcal{N} \Bigl((1-\bar\alpha_t)\,\boldsymbol\mu + \sqrt{\bar\alpha_t}\,\boldsymbol{x}_t,\;(1-\bar\alpha_t)\,I\Bigr).
\]
We have the denoising model’s predictive distribution for label \(y\) given \(\boldsymbol{x}_t\) as
\[
p_t\bigl(y\mid \boldsymbol{x}_t\bigr)
=\int p\bigl(y\mid \boldsymbol{x}_1\bigr)\;p(\boldsymbol{x}_1\mid\boldsymbol{x}_t)\,\mathrm{d}\boldsymbol{x}_1
=\frac{1}{Z}\int\exp \bigl(f_y(\boldsymbol{x}_1)\bigr)\;p(\boldsymbol{x}_1\mid\boldsymbol{x}_t)\,\mathrm{d}\boldsymbol{x}_1,
\]
where the objective function is linear \(f_y(\boldsymbol{x})=\boldsymbol{g}^\top\boldsymbol{x}\).  Since for a Gaussian \(\mathcal{N}(m,\Sigma)\),
\[
\int\exp \bigl(\boldsymbol{g}^\top x\bigr)\,\mathcal{N}(x;m,\Sigma)\,\mathrm{d}x
=\exp \Bigl(\boldsymbol{g}^\top m + \tfrac12\,\boldsymbol{g}^\top\Sigma\,\boldsymbol{g}\Bigr),
\]
it follows that
\[
p_t\bigl(y\mid \boldsymbol{x}_t\bigr)
\propto
\exp \Bigl(\boldsymbol{g}^\top\bigl((1-\bar\alpha_t)\,\boldsymbol\mu + \sqrt{\bar\alpha_t}\,\boldsymbol{x}_t\bigr)
\;+\;\tfrac12\,(1-\bar\alpha_t)\,\|\boldsymbol{g}\|^2\Bigr).
\]
Therefore, we have
\begin{equation*}
    \nabla_{\boldsymbol{x}_t}\log p_t\bigl(y\mid \boldsymbol{x}_t\bigr) = \sqrt{\bar\alpha_t}\,\boldsymbol{g},
\end{equation*}
which allows us to express the conditional score as:
\begin{equation}\label{eq:cond linear}
    \nabla_{\boldsymbol{x}_t}\log p_t\bigl( \boldsymbol{x}_t \mid y\bigr) =  
    \nabla_{\boldsymbol{x}_t}\log p_t\bigl( \boldsymbol{x}_t\bigr) + \sqrt{\bar\alpha_t}\,\boldsymbol{g}.
\end{equation}

Recall the transition step \eqref{eq:diffusion ddpm sampling}. The unconditional transition is given by:
\begin{equation}\label{eq:uncond tran gaussian}
    \mathcal{T}(\cdot \mid \boldsymbol{x}_t) = \mathcal{N}\left(\cdot;\, \Bar{\boldsymbol{x}}_{t+\Delta t},\, \sigma^2_t \boldsymbol{I}\right),
\end{equation}
where
\[
\Bar{\boldsymbol{x}}_{t+\Delta t} = \frac{\boldsymbol{x}_t + (1 - \alpha_t)\nabla \log p_t(\boldsymbol{x}_t)}{\sqrt{\alpha_t}}.
\]

Using the conditional score from \eqref{eq:cond linear}, the target {conditional} transition distribution becomes:
\begin{equation}\label{eq:cond tran gaussian}
    \mathcal{T}(\cdot \mid \boldsymbol{x}_t, y) = \mathcal{N}\left(\cdot;\, \Bar{\boldsymbol{x}}_{t+\Delta t} + (1 - \alpha_t)\sqrt{\bar{\alpha}_{t+\Delta t}}\,\boldsymbol{g},\, \sigma^2_t \boldsymbol{I}\right).
\end{equation}

\ouralg-Sampling Current proceeds by generating i.i.d. samples \( X_k \sim \mathcal{T}(\cdot \mid \boldsymbol{x}_t) \), followed by multinomial resampling according to the weights
\[
V(\boldsymbol{x}) = \exp\left(f_y\left(\boldsymbol{x}_{1|t}(\boldsymbol{x}_t)\right)\right),
\]
where, using Tweedie’s formula \citep{efron2011tweedie}, the posterior mean of \( \boldsymbol{x}_1 \) given \( \boldsymbol{x}_t \) is:
\[
\boldsymbol{x}_{1|t} = \mathbb{E}[\boldsymbol{x}_1 \mid \boldsymbol{x}_t] = (1 - \bar{\alpha}_t)\,\boldsymbol{\mu} + \sqrt{\bar{\alpha}_t}\,\boldsymbol{x}_t.
\]
Therefore, the value function simplifies to:
\[
f_y(\boldsymbol{x}_{1|t}(\boldsymbol{x}_t)) = \boldsymbol{g}^\top\left((1 - \bar{\alpha}_t)\boldsymbol{\mu} + \sqrt{\bar{\alpha}_t}\,\boldsymbol{x}_t\right).
\]

From \cref{lmm:branch K error}, with probability at least \( 1 - \delta \), the distribution \( \hat{\mathcal{T}} \) produced by this resampling satisfies:
\begin{equation}\label{eq:error to resampling}
    \left\| \hat{\mathcal{T}} - \mathcal{T}_1 \right\|_{1} \leq 4(1 + C)\,\sqrt{\frac{\log(4 / \delta)}{2 K}},
\end{equation}
where the adjusted distribution is
\[
\mathcal{T}_1(\boldsymbol{x}) \propto \exp\left(\sqrt{\bar{\alpha}_t}\,\boldsymbol{g}^\top \boldsymbol{x}\right) \cdot \mathcal{T}(\boldsymbol{x} \mid \boldsymbol{x}_t),
\]
and the constant \( C \) is given by
\[
C = \frac{\|\boldsymbol{g}\|_2^2}{|\boldsymbol{g}^\top \bar{\boldsymbol{x}}_{t+\Delta t}|}.
\]

We have
\begin{equation}
    \mathcal{T}_1(\cdot) = \mathcal{N}\left(\cdot;\, \Bar{\boldsymbol{x}}_{t+\Delta t} + \sigma^2_t \sqrt{\bar{\alpha}_t}\,\boldsymbol{g},\, \sigma^2_t \boldsymbol{I}\right).
\end{equation}

Now, comparing the means of \( \mathcal{T}_1 \) and \( \mathcal{T}(\cdot \mid \boldsymbol{x}_t, y) \), we use the bound via KL divergence:
\begin{equation*}
    \left\| \mathcal{T}_1 - \mathcal{T}(\cdot \mid \boldsymbol{x}_t, y) \right\|_{1} 
    \le \sqrt{2\mathrm{KL}(\mathcal{T}_1 \| \mathcal{T}(\cdot \mid \boldsymbol{x}_t, y))} 
    = \frac{2\|\mu_1 - \mu_2\|_2}{ \sigma_t},
\end{equation*}
where
\[
\mu_1 = \Bar{\boldsymbol{x}}_{t+\Delta t} + \sigma^2_t \sqrt{\bar{\alpha}_t}\,\boldsymbol{g}, \quad 
\mu_2 = \Bar{\boldsymbol{x}}_{t+\Delta t} + (1 - \alpha_t)\sqrt{\bar{\alpha}_{t+\Delta t}}\,\boldsymbol{g},
\]
with \( \sigma_t = \sqrt{1 - \alpha_t} \) and \( \alpha_t = \bar{\alpha}_t / \bar{\alpha}_{t+\Delta t} \). Thus,
\begin{equation}
    \left\| \mathcal{T}_1 - \mathcal{T}(\cdot \mid \boldsymbol{x}_t, y) \right\|_{1} 
    \le 2 \bar{\alpha}_{t+\Delta t} \sqrt{1 - \alpha_t}(1 - \sqrt{\alpha_t}) \, \|\boldsymbol{g}\|_2.
\end{equation}

Combining this with \eqref{eq:error to resampling}, we obtain:
\begin{equation*}
    \left\| \hat{\mathcal{T}} - \mathcal{T}(\cdot \mid \boldsymbol{x}_t, y) \right\|_{1} 
    \le 4(1 + C)\,\sqrt{\frac{\log(4 / \delta)}{2 K}} 
    + 2 \bar{\alpha}_{t+\Delta t} \sqrt{1 - \alpha_t}(1 - \sqrt{\alpha_t}) \, \|\boldsymbol{g}\|_2.
\end{equation*}

Thus, if \( K = \Theta\left((1 + C)^2 \frac{\log(1/\delta)}{\varepsilon^2}\right) \) and \( 1 - \alpha_t = O\left(\varepsilon^2/\|\boldsymbol{g}\|_2^2 \right) \), provided that \( \Delta t \) is sufficiently small, the total variation distance is bounded by \( \varepsilon \). Ignoring constant factors, this completes the proof.

\textbf{Proof of (b) discrete case:} We simplify the notation and recall some key definitions for clarity.

At time step \( t \), the current state is given by \( \boldsymbol{x}_t \). The unconditional transition probability is defined as:
\[
\mathcal{T}(\cdot \mid \boldsymbol{x}_t) = \delta \{ \cdot, \boldsymbol{x}_t \} + R_t(\boldsymbol{x}_t, \cdot) \, \Delta t,
\]
where \( \delta \{ \cdot, \boldsymbol{x}_t \} \) is the Dirac delta function. The target conditional distribution is
\[
\mathcal{T}(\cdot \mid \boldsymbol{x}_t, y) = \delta \{ \cdot, \boldsymbol{x}_t \} + R_t(\boldsymbol{x}_t, \cdot \mid y) \, \Delta t,
\]
where the conditional rate is defined as \citep{nisonoff2024unlocking}:
\[
R_t(\boldsymbol{x}_t, \boldsymbol{x} \mid y) = \frac{p_t(y \mid \boldsymbol{x})}{p_t(y \mid \boldsymbol{x}_t)} \, R_t(\boldsymbol{x}_t, \boldsymbol{x}), \quad \text{for } \boldsymbol{x} \neq \boldsymbol{x}_t,
\]
and
\[
R_t(\boldsymbol{x}_t, \boldsymbol{x}_t \mid y) = - \sum_{\boldsymbol{x} \neq \boldsymbol{x}_t} R_t(\boldsymbol{x}_t, \boldsymbol{x} \mid y).
\]

For simplicity, we introduce the shorthand:
\[
\mathcal{T}_0 := \mathcal{T}(\cdot \mid \boldsymbol{x}_t), \quad \text{and} \quad \mathcal{T}^\ast := \mathcal{T}(\cdot \mid \boldsymbol{x}_t, y).
\]

We then define the distribution used in importance sampling, where we first sample from \( \mathcal{T}_0 \) and then reweight according to the likelihood \( p_t(y \mid \cdot) \):
\begin{equation*}\label{eq:def reweigh pt}
\mathcal{T}_1(\boldsymbol{x}) = \frac{p_t(y \mid \boldsymbol{x}) \, \mathcal{T}_0(\boldsymbol{x})}{\sum_{\boldsymbol{x}} p_t(y \mid \boldsymbol{x}) \, \mathcal{T}_0(\boldsymbol{x})}.
\end{equation*}

In our algorithm, we use Monte Carlo estimation to approximate the likelihood:
\[
p_t(y \mid \boldsymbol{x}) = \int p(y \mid \boldsymbol{x}_1) \, p_{1|t}(\boldsymbol{x}_1 \mid \boldsymbol{x}) \, \mathrm{d} \boldsymbol{x}_1 = \frac{1}{Z} \int \exp \big( f_y(\boldsymbol{x}_1) \big) \, p_{1|t}(\boldsymbol{x}_1 \mid \boldsymbol{x}) \, \mathrm{d} \boldsymbol{x}_1.
\]
This is estimated using the following procedure (as shown in Lines 2 and 3 of Value Function~\ref{vfunc:noisy}):
\begin{equation}\label{eq:MC estimate}
    \frac{1}{N} \sum_{i=1}^N \exp \big( f_y(\hat{\boldsymbol{x}}_1^i) \big), \quad \hat{\boldsymbol{x}}_1^i \sim \mathrm{Cat} \big( u_\theta(\boldsymbol{x}, t + \Delta t) \big), \quad i \in [N],
\end{equation}
where we assume the pretrained flow model is perfect, i.e., \( u_\theta(\boldsymbol{x}, t + \Delta t) = p_{1|t+\Delta t}(\boldsymbol{x}_1 \mid \boldsymbol{x}) \).

We denote the Monte Carlo estimate of the likelihood in \eqref{eq:MC estimate} as \( \hat{q} \). Using this, we define the distribution used in importance sampling, where we first sample from \( \mathcal{T}_0 \) and reweight according to \( \hat{q} \):
\begin{equation*}\label{eq:def reweigh hat q}
\mathcal{T}_2(\boldsymbol{x}) = \frac{\hat{q}(\boldsymbol{x}) \, \mathcal{T}_0(\boldsymbol{x})}{\sum_{\boldsymbol{x}} \hat{q}(\boldsymbol{x}) \, \mathcal{T}_0(\boldsymbol{x})}.
\end{equation*}

When resampling over \( K \) candidates \( \boldsymbol{x}^k \sim \mathcal{T}_0 \) and reweighting according to \( \hat{q} \), the resulting (empirical) output distribution is:
\begin{equation*}
    \hat{\mathcal{T}}\rbr{\boldsymbol{x}} = \sum_{k=1}^K w_k \, \mathbf{1}\rbr{\boldsymbol{x}={\boldsymbol{x}^k}},
\end{equation*}
where \( w_k = \hat{q}(\boldsymbol{x}^k) / \sum_{i=1}^K \hat{q}(\boldsymbol{x}^i) \).

We can therefore decompose the distance between the empirical distribution \( \hat{\mathcal{T}} \) and the target distribution \( \mathcal{T}^\ast \) as:
\begin{equation}\label{eq:TV decompose}
    \| \mathcal{T}^\ast - \hat{\mathcal{T}} \|_{\infty} \leq \| \mathcal{T}^\ast - \mathcal{T}_1 \|_{\infty} + \| \mathcal{T}_1 - \mathcal{T}_2 \|_{\infty} + \| \mathcal{T}_2 - \hat{\mathcal{T}} \|_{\infty}.
\end{equation}

We now proceed to bound the three terms on the right-hand side of \eqref{eq:TV decompose}, starting with the first term involving $\mathcal{T}_1$. We have:
\begin{align*}
\sum_{\boldsymbol{x}} p_t(y \mid \boldsymbol{x}) \, \mathcal{T}_0(\boldsymbol{x}) 
&= p_t(y \mid \boldsymbol{x}_t) \left( 1 - \sum_{\boldsymbol{z} \neq \boldsymbol{x}_t} R_t(\boldsymbol{x}_t, \boldsymbol{z}) \, \Delta t \right) 
+ \sum_{\boldsymbol{z} \neq \boldsymbol{x}_t} p_t(y \mid \boldsymbol{z}) R_t(\boldsymbol{x}_t, \boldsymbol{z}) \, \Delta t \\
&= p_t(y \mid \boldsymbol{x}_t) - p_t(y \mid \boldsymbol{x}_t) \sum_{\boldsymbol{z} \neq \boldsymbol{x}_t} R_t(\boldsymbol{x}_t, \boldsymbol{z}) \, \Delta t 
+ \sum_{\boldsymbol{z} \neq \boldsymbol{x}_t} p_t(y \mid \boldsymbol{z}) R_t(\boldsymbol{x}_t, \boldsymbol{z}) \, \Delta t \\
&= p_t(y \mid \boldsymbol{x}_t) + 
\left( \sum_{\boldsymbol{z} \neq \boldsymbol{x}_t} \big[ p_t(y \mid \boldsymbol{z}) - p_t(y \mid \boldsymbol{x}_t) \big] R_t(\boldsymbol{x}_t, \boldsymbol{z}) \right) \Delta t \\
&= p_t(y \mid \boldsymbol{x}_t) + O(\Delta t).
\end{align*}

Therefore, for any $\boldsymbol{x} \neq \boldsymbol{x}_t$, we have:
\begin{equation*}
    \mathcal{T}_1(\boldsymbol{x}) = \frac{p_t(y \mid \boldsymbol{x}) \, \mathcal{T}_0(\boldsymbol{x})}{\sum_{\boldsymbol{x}} p_t(y \mid \boldsymbol{x}) \, \mathcal{T}_0(\boldsymbol{x})} 
    = \frac{p_t(y \mid \boldsymbol{x}) \, R_t(\boldsymbol{x}_t, \boldsymbol{x}) \, \Delta t}{p_t(y \mid \boldsymbol{x}_t) + O(\Delta t)} 
    = \frac{p_t(y \mid \boldsymbol{x})}{p_t(y \mid \boldsymbol{x}_t)} R_t(\boldsymbol{x}_t, \boldsymbol{x}) \, \Delta t + O(\Delta t^2).
\end{equation*}

As a result, we can bound the distance as:
\begin{equation}\label{eq:bound term 1}
    \| \mathcal{T}^\ast - \mathcal{T}_1 \|_{\infty}  
    = O(\Delta t^2).
\end{equation}

We now bound the error introduced by the Monte Carlo estimation \eqref{eq:MC estimate}.  
By \cref{lmm:MC high prob}, with probability at least $1 - \delta / 2$, we have:
\begin{equation*}
    \max_{\boldsymbol{x} \in \mathcal{X}} \bigl| \hat{q}(\boldsymbol{x}) - q_{1|t+\Delta t}(\boldsymbol{x}) \bigr| \leq \epsilon_1,
\end{equation*}
where the unnormalized quantity is defined as 
\[
q_{1|t}(\boldsymbol{x}) := \int \exp \big( f_y(\boldsymbol{x}_1) \big) \, p_{1|t}(\boldsymbol{x}_1 \mid \boldsymbol{x}) \, \mathrm{d} \boldsymbol{x}_1,
\]
and where 
\[
\epsilon_1 = B_{\max} \, \sqrt{\frac{\log(4 |\mathcal{X}| / \delta)}{2 N}}, \quad B_{\max} = \sup_{\boldsymbol{x} \in \mathcal{X}} \exp \big( f_y(\boldsymbol{x}) \big).
\]

Additionally, due to the Lipschitz continuity of $t \mapsto p_{1|t}(\boldsymbol{x}_1 \mid \boldsymbol{x})$, we have:
\[
| p_{1|t}(\boldsymbol{x}_1 \mid \boldsymbol{x}) - p_{1|t + \Delta t}(\boldsymbol{x}_1 \mid \boldsymbol{x}) | \leq L \, \Delta t,
\]
which implies:
\[
| q_{1|t}(\boldsymbol{x}) - q_{1|t+\Delta t}(\boldsymbol{x}) | \leq B_{\max} L \, \Delta t.
\]

Therefore, combining the Monte Carlo estimation error and the temporal approximation error, we get (with probability at least $1 - \delta / 2$):
\begin{equation}\label{eq:q error}
    \max_{\boldsymbol{x} \in \mathcal{X}} \bigl| \hat{q}(\boldsymbol{x}) - q_{1|t}(\boldsymbol{x}) \bigr| \leq \epsilon_1 + B_{\max} L \, \Delta t.
\end{equation}

Recall that:
\begin{equation*}
    \mathcal{T}_1(\boldsymbol{x}) = \frac{q_{1|t}(\boldsymbol{x}) \, \mathcal{T}_0(\boldsymbol{x})}{\sum_{\boldsymbol{x}} q_{1|t}(\boldsymbol{x}) \, \mathcal{T}_0(\boldsymbol{x})}, 
    \quad \text{where} \quad q_{1|t}(\boldsymbol{x}) = Z \, p_t(y \mid \boldsymbol{x}),
\end{equation*}
and
\begin{equation*}
    \mathcal{T}_2(\boldsymbol{x}) = \frac{\hat{q}(\boldsymbol{x}) \, \mathcal{T}_0(\boldsymbol{x})}{\sum_{\boldsymbol{x}} \hat{q}(\boldsymbol{x}) \, \mathcal{T}_0(\boldsymbol{x})}.
\end{equation*}

By applying \eqref{eq:q error} and using \cref{lmm:q error to distribution}, we obtain  with probability at least $1 - \delta / 2$:
\begin{equation}\label{eq:bound term 2}
    \| \mathcal{T}_1 - \mathcal{T}_2 \|_{\infty} \leq \frac{4}{B_{\min}} \big( \epsilon_1 + B_{\max} L \, \Delta t \big) 
    = \frac{4 B_{\max}}{B_{\min}} \left( \sqrt{\frac{\log(4 |\mathcal{X}| / \delta)}{2 N}} + L \, \Delta t \right),
\end{equation}
where 
\[
B_{\min} = \inf_{\boldsymbol{x} \in \mathcal{X}} \exp \big( f_y(\boldsymbol{x}) \big).
\]
And it requires:
\begin{equation*}
    \epsilon_1 + B_{\max} L \, \Delta t < B_{\min}.
\end{equation*}

Finally, we bound the error introduced by sampling from the multinomial distribution over the $K$ candidate branches. By \cref{lmm:branch K error}, we have that with probability at least $1 - \delta/2$,
\begin{equation}\label{eq:bound term 3}
    \| \mathcal{T}_2 - \hat{\mathcal{T}} \|_{\infty} \leq \rbr{1 + B_{\max}/B_{\min}} \sqrt{\frac{\log(4 |\mathcal{X}| / \delta)}{2 K}}.
\end{equation}

Now, combining the three error terms — from \eqref{eq:bound term 1}, \eqref{eq:bound term 2}, and \eqref{eq:bound term 3} — into the decomposition \eqref{eq:TV decompose}, we obtain that with overall probability at least $1 - \delta$,
\begin{equation*}
    \| \mathcal{T}^\ast - \hat{\mathcal{T}} \|_{\infty} \leq O( \Delta t^2) + \frac{4 B_{\max}}{B_{\min}} \left( \sqrt{\frac{\log(4 |\mathcal{X}| / \delta)}{2 N}} + L \Delta t \right) + \sqrt{\frac{\log(4 |\mathcal{X}| / \delta)}{2 K}}.
\end{equation*}

To ensure that the total error stays below a prescribed threshold $\varepsilon$. Specifically, we choose the time step $\Delta t = O\left( \sqrt{\varepsilon} + \frac{B_{\min}}{B_{\max}L} \varepsilon \right)$, the Monte Carlo sample size $N = \Theta\left( \frac{B_{\max}^2 \log(4 |\mathcal{X}| / \delta)}{B_{\min}^2 \varepsilon^2} \right)$, and the branch candidate size $K = \Theta\left( \frac{B_{\max}^2 \log( 4 |\mathcal{X}|/ \delta)}{B_{\min}^2\varepsilon^2} \right)$. If we omit constant factors, this completes the proof.

\end{proof}

\subsection{Proof of Lemma~\ref{lemma:transition prob}}\label{prof:transition prob}

\begin{lemma*}[Recall Lemma~\ref{lemma:transition prob}]
In both continuous and discrete cases, the transition probability during inference at timestep $t$ satisfies:
\begin{equation*}
\begin{aligned}
    \mathcal{T}(\boldsymbol{x}_{t + \Delta t} \mid \boldsymbol{x}_{t}) = \EE_{{\hat{\boldsymbol{x}}}_1} \sbr{\mathcal{T}^{\star}(\boldsymbol{x}_{t + \Delta t} \mid \boldsymbol{x}_{t}, \hat{{\boldsymbol{x}}}_1)},
\end{aligned}
\end{equation*}
where the expectation is taken over a distribution estimated by $u_\theta$, with $\mathcal{T}^{\star}$ being the true posterior distribution predetermined by the noise schedule.
\end{lemma*}

\begin{proof}
In the continuous case, the forward process is defined as:
\[
\boldsymbol{x}_t \sim \mathcal{N}(\bar{\alpha}_t \boldsymbol{x}_1, (1 - \bar{\alpha}_t)\boldsymbol{I}), \quad 
\boldsymbol{x}_{t+\Delta t} \sim \mathcal{N}(\bar{\alpha}_{t+\Delta t} \boldsymbol{x}_1, (1 - \bar{\alpha}_{t+\Delta t})\boldsymbol{I}).
\]
Given this, the posterior distribution conditioned on both \(\boldsymbol{x}_t\) and \(\boldsymbol{x}_1\) is:
\begin{equation}
p(\cdot \mid \boldsymbol{x}_t, \boldsymbol{x}_1) = \mathcal{N}(\cdot; c_{t,1}\boldsymbol{x}_t + c_{t,2} \boldsymbol{x}_1, \beta_t \boldsymbol{I}),
\end{equation}
where the coefficients are defined as:
\[
\beta_t = \frac{1 - \bar{\alpha}_{t+\Delta t}}{1 - \bar{\alpha}_t}(1 - \alpha_t), \quad
c_{t,1} = \frac{\sqrt{\alpha_t}(1 - \bar{\alpha}_{t+\Delta t})}{1 - \bar{\alpha}_t}, \quad
c_{t,2} = \frac{\sqrt{\bar{\alpha}_{t+\Delta t}}(1 - \alpha_t)}{1 - \bar{\alpha}_t}.
\]

We set the transition distribution as:
\[
\mathcal{T}^\ast(\cdot \mid \boldsymbol{x}_t, \boldsymbol{x}_1) = \mathcal{N}(\cdot; c_{t,1}\boldsymbol{x}_t + c_{t,2} \boldsymbol{x}_1, \beta_t \boldsymbol{I}).
\]

Let the estimate \(\hat{\boldsymbol{x}}_1\) be sampled from the distribution:
\[
q_1 = \mathcal{N}(\boldsymbol{x}_{1|t}, (1 - \alpha_t)\boldsymbol{I}),
\]
where, as given in \eqref{eq:pred_x1_continuous}, the posterior mean \(\boldsymbol{x}_{1|t}\) is:
\[
\boldsymbol{x}_{1|t} = \frac{1}{\sqrt{\bar{\alpha}_t}} \left(\boldsymbol{x}_t - \sqrt{1 - \bar{\alpha}_t} \cdot u_\theta(\boldsymbol{x}_t, t)\right).
\]

Then, the expected approximate posterior becomes:
\begin{equation} \label{eq:app_posterior}
\mathbb{E}_{\hat{\boldsymbol{x}}_1 \sim q_1} \left[\mathcal{T}^\ast(\boldsymbol{x}_{t + \Delta t} \mid \boldsymbol{x}_t, \hat{\boldsymbol{x}}_1)\right] = \mathcal{N}\left(c_{t,1}\boldsymbol{x}_t + c_{t,2}\boldsymbol{x}_{1|t}, \sigma_t^2 \boldsymbol{I} \right),
\end{equation}
where \(\sigma_t = \sqrt{1 - \alpha_t}\).

Thus, \eqref{eq:app_posterior} defines the transition distribution \(\mathcal{T}(\boldsymbol{x}_{t + \Delta t} \mid \boldsymbol{x}_t)\) used for sampling during inference.

For the discrete case, we recall the expression for the rate matrix used during inference, as given in \eqref{eq:compute rate matrix}:
\begin{equation*}
    R_{t}^{(d)}(\boldsymbol{x}_t, j) = \mathbb{E}_{x_1^{(d)} \sim u_{\theta}^{(d)}(x_1 \mid \boldsymbol{x}_t)} \left[ R_t(x_t^{(d)}, j \mid x_1^{(d)}) \right],
\end{equation*}

where the conditional rate matrix is defined as
\begin{equation*}
    R_t(x_t, j \mid x_1) = \frac{\delta\{j, x_1\}}{1 - t} \cdot \mathbf{1}\{x_t = M\}.
\end{equation*}

Given $\boldsymbol{x}_t$ and $\boldsymbol{x}_1$, the transition probability is
\begin{equation*}
    \mathcal{T}^{\ast, (d)}(j \mid \boldsymbol{x}_t, \boldsymbol{x}_1) = \delta\{x_t^{(d)}, j\} + R_t(x_t^{(d)}, j \mid x_1^{(d)}) \cdot \Delta t.
\end{equation*}

Taking the expectation over $\boldsymbol{x}_1 \sim u_\theta(\cdot \mid \boldsymbol{x}_t)$, we obtain

\begin{equation*}
    \mathbb{E}_{\boldsymbol{x}_1 \sim u_\theta} \left[ \mathcal{T}^{\ast, (d)}(j \mid \boldsymbol{x}_t, \boldsymbol{x}_1) \right] = \delta\{x_t^{(d)}, j\} + \mathbb{E}_{x_1^{(d)} \sim u_{\theta}^{(d)}(x_1 \mid \boldsymbol{x}_t)} \left[ R_t(x_t^{(d)}, j \mid x_1^{(d)}) \right] \cdot \Delta t,
\end{equation*}
which corresponds to the transition probability $\mathcal{T}(\boldsymbol{x}_{t + \Delta t} \mid \boldsymbol{x}_t)$. This completes the proof.
\end{proof}

\subsection{Proof of \cref{thm:x1_sampling}}\label{prof:x1_sampling}

\begin{theorem*}[Restatement of \cref{thm:x1_sampling}]
Consider \ouralg-Sampling Destination at time $t$, with an active set of size one and selection performed via multinomial resampling. Then, for any $\varepsilon, \delta > 0$, it holds with probability $1-\delta$:
\begin{equation*}
        \norm{\hat{\mathcal{T}} - \mathcal{T}(\cdot \mid \boldsymbol{x}_{t}, y)}_{\ell} \, < \varepsilon,
\end{equation*}
provided one of the following conditions is satisfied:\\
\textbf{(a, Continuous)}It holds under the $\ell = 1$ norm. Suppose data follow Gaussian distribution and the objective function is linear. The Branch-Out size is $K = \Theta(\frac{\log (1/\delta)}{\varepsilon^2})$ and we set $\rho_t = 1 - \alpha_t, \tau_t = \frac{1 - \bar{\alpha}_{t+\Delta t}}{1 - \bar{\alpha}_t}(1 - \alpha_t).$ \\
\textbf{(b, Discrete)}It holds under the $\ell = \infty$ norm. The likelihood scores satisfy $0 < B_{\min} \le \exp(f_y(\boldsymbol{x})) \le B_{\max}$ for all $\boldsymbol{x}$. The Branch-Out size is $K = \Theta(\frac{\log (|\mathcal{X}|/\delta)}{\varepsilon^{2/D}})$  where $\mathcal{X}$ is the data space and $D$  is its dimension.
\end{theorem*}

\begin{proof}

\textbf{Proof of (a) continuous case:}
We begin by recalling the unconditional transition process given in \eqref{eq:linear interpolation sampling}:
\[
\boldsymbol{x}_{t+\Delta t} = c_1 \boldsymbol{x}_t + c_2 \boldsymbol{x}_{1|t} + \sigma_t \boldsymbol{\epsilon},
\]
where the coefficients are defined as 
\(
c_1 = c_{t,1} = \frac{\sqrt{\alpha_t}(1 - \bar{\alpha}_{t+\Delta t})}{1 - \bar{\alpha}_t}, 
c_2 = c_{t,2} = \frac{\sqrt{\bar{\alpha}_{t+\Delta t}}(1 - \alpha_t)}{1 - \bar{\alpha}_t},  
\sigma_t = \sqrt{ 1 - \alpha_t},
\)
and $\boldsymbol{x}_{1|t} = \mathbb{E}[\boldsymbol{x}_1 \mid \boldsymbol{x}_t]$.

Assuming that the data distribution is Gaussian and the objective function is linear, we have:
\[
\boldsymbol{x}_1 \sim \mathcal{N}(\boldsymbol{\mu}, \boldsymbol{I}),\qquad
f_y(\boldsymbol{x}) = \boldsymbol{g}^\top \boldsymbol{x}.
\]

In Line 2 of Module~\ref{mol:x1_sampling} and the associated Value Function~\ref{vfunc:clean}, a candidate $\hat{\boldsymbol{x}}_1$ is proposed by sampling from $\mathcal{N}(\boldsymbol{x}_{1|t}, \rho_t \boldsymbol{I})$ and reweighting using $\exp(f_y(\boldsymbol{x})) = \exp(\boldsymbol{g}^\top \boldsymbol{x})$. Let $\hat{q}$ denote the distribution density of the output $\hat{\boldsymbol{x}}_1$. According to Lemma~\ref{lmm:branch K error}, with probability at least $1 - \delta$, the distance between $\hat{q}$ and the ideal distribution $q$ satisfies
\begin{equation}\label{eq:K error x0 continuous}
\|\hat{q} - q\|_{1} < 4(1 + C)\, \sqrt{\frac{\log(4/\delta)}{2K}},
\end{equation}
where the ideal distribution is
\[
q(\boldsymbol{x}) \propto \exp(\boldsymbol{g}^\top \boldsymbol{x}) \cdot \mathcal{N}(\boldsymbol{x}; \boldsymbol{x}_{1|t}, \rho_t \boldsymbol{I}),
\]
and the constant $C$ is defined as
\(
C = \frac{\|\boldsymbol{g}\|_2^2}{|\boldsymbol{g}^\top \boldsymbol{x}_{1|t}|}.
\)
It follows from this form that $q$ is a Gaussian with mean $\boldsymbol{x}_{1|t} + \rho_t \boldsymbol{g}$ and covariance $\rho_t I$, i.e.,
\[
q = \mathcal{N}(\boldsymbol{x}_{1|t} + \rho_t \boldsymbol{g}, \rho_t \boldsymbol{I}).
\]

In Line 3 of Module~\ref{mol:x1_sampling}, the next sample $\boldsymbol{x}_{t+\Delta t}$ is generated according to
\[
\boldsymbol{x}_{t+\Delta t} \sim \mathcal{N}(c_1 \boldsymbol{x}_t + c_2 \hat{\boldsymbol{x}}_1, \tau_t \boldsymbol{I}), \quad \hat{\boldsymbol{x}}_1 \sim \hat{q}.
\]
Consequently, the distribution of $\boldsymbol{x}_{t+\Delta t}$, denoted $\hat{\mathcal{T}}$, is the convolution \footnote{The convolution operation models the distribution of the sum of two independent random variables. For probability density functions $f$ and $g$ on $\mathbb{R}^d$, the convolution is defined as $(f \ast g)(\boldsymbol{x}) = \int_{\mathbb{R}^d} f(\boldsymbol{y}) g(\boldsymbol{x} - \boldsymbol{y}) \, d\boldsymbol{y}$. In the case of Gaussian distributions, this results in a new Gaussian with summed means and covariances.} of $c_2 \hat{q}$ and $\varphi$, where $\varphi = \mathcal{N}(c_1 \boldsymbol{x}_t, \tau_t \boldsymbol{I})$:
\[
\hat{\mathcal{T}} = (c_2 \hat{q}) \ast \varphi.
\]

Next, we will show that the true conditional transition probability satisfies
\[
\mathcal{T}(\cdot \mid \boldsymbol{x}_t, y) = (c_2 q) \ast \varphi,
\]
by setting
\[
\rho_t = 1 - \alpha_t, \qquad \tau_t = \frac{1 - \bar{\alpha}_{t+\Delta t}}{1 - \bar{\alpha}_t}(1 - \alpha_t).
\]
By the property of convolution between Gaussian distributions, we have
\begin{equation*}
\begin{aligned}
(c_2 q) \ast \varphi 
&= \mathcal{N}(c_2 \boldsymbol{x}_{1|t} + c_2 \rho_t \boldsymbol{g},\, c_2^2 \rho_t\, \boldsymbol{I}) 
\ast \mathcal{N}(c_1 \boldsymbol{x}_t,\, \tau_t\, \boldsymbol{I}) \\
&= \mathcal{N}(c_1 \boldsymbol{x}_t + c_2 \boldsymbol{x}_{1|t} + c_2 \rho_t \boldsymbol{g},\, (\tau_t + c_2^2 \rho_t)\, \boldsymbol{I}).
\end{aligned}
\end{equation*}

We now verify that this matches the conditional transition distribution by examining the mean and variance. Starting with the variance term, we compute
\begin{equation*}
\begin{aligned}
\tau_t + c_2^2 \rho_t 
&= \frac{1 - \bar{\alpha}_{t+\Delta t}}{1 - \bar{\alpha}_t}(1 - \alpha_t) 
+ \left( \frac{\sqrt{\bar{\alpha}_{t+\Delta t}} (1 - \alpha_t)}{1 - \bar{\alpha}_t} \right)^2 (1 - \bar{\alpha}_t) \\
&= \frac{1 - \alpha_t}{1 - \bar{\alpha}_t} \left( 1 - \bar{\alpha}_{t+\Delta t} + \bar{\alpha}_{t+\Delta t} - \bar{\alpha}_{t+\Delta t} \alpha_t \right) \\
&= \frac{1 - \alpha_t}{1 - \bar{\alpha}_t} (1 - \bar{\alpha}_{t+\Delta t} \alpha_t) = 1 - \alpha_t.
\end{aligned}
\end{equation*}

For the mean, we have
\[
c_1 \boldsymbol{x}_t + c_2 \boldsymbol{x}_{1|t} + c_2 \rho_t \boldsymbol{g}
= c_1 \boldsymbol{x}_t + c_2 \boldsymbol{x}_{1|t} + (1 - \alpha_t) \sqrt{\bar{\alpha}_{t+\Delta t}}\, \boldsymbol{g}.
\]

Recalling from ~\eqref{eq:cond tran gaussian} in the proof in App.~\ref{prof:xt_sampling}, the target conditional transition distribution is
\[
\mathcal{T}(\cdot \mid \boldsymbol{x}_t, y) 
= \mathcal{N}\left( \cdot;\, \bar{\boldsymbol{x}}_{t+\Delta t} + (1 - \alpha_t) \sqrt{\bar{\alpha}_{t+\Delta t}}\, \boldsymbol{g},\, \sigma_t^2 \boldsymbol{I} \right),
\]
where the mean is \( \bar{\boldsymbol{x}}_{t+\Delta t} = c_1 \boldsymbol{x}_t + c_2 \boldsymbol{x}_{1|t} \).

Hence, the convolution \( (c_2 q) \ast \varphi \) exactly recovers the conditional transition distribution \( \mathcal{T}(\cdot \mid \boldsymbol{x}_t, y) \).
Therefore, we bound the  distance between the output distribution and target transition distributions as follows:
\begin{equation*}
\begin{aligned}
\norm{\hat{\mathcal{T}} - \mathcal{T}(\cdot \mid \boldsymbol{x}_{t}, y)}_{1} 
&= \norm{(c_2 \hat{q}) \ast \varphi - (c_2 q) \ast \varphi}_{1} \\
&\le \norm{c_2 \hat{q} - c_2 q}_{\mathrm{TV}} 
= c_2 \norm{\hat{q} - q}_{1},
\end{aligned}
\end{equation*}
where the inequality follows from the data processing inequality under convolution with a fixed distribution.

With \eqref{eq:K error x0 continuous}, we obtain that with probability at least $1 - \delta$,
\begin{equation*}
\norm{\hat{\mathcal{T}} - \mathcal{T}(\cdot \mid \boldsymbol{x}_{t}, y)}_{1} 
\leq 4c_2 (1 + C)\, \sqrt{\frac{\log(4/\delta)}{2K}}.
\end{equation*}

Hence, to ensure the approximation error is at most $\varepsilon$, it suffices to choose the branch-out size 
\(
K = \Theta\left(c_2^2 (1 + C)^2\, \frac{\log (1/\delta)}{\varepsilon^2}\right),
\)
up to constant factors, which completes the proof.

\textbf{Proof of (b) discrete case:}

In Line 2 of Module~\ref{mol:x1_sampling}, and with Value Function \ref{vfunc:clean}, a candidate state $\hat{\boldsymbol{x}}_1$ is proposed by sampling from the distribution $p_{1|t}(\cdot \mid \boldsymbol{x}_t)$(since $u_\theta$ is optimal estimation), where the reweighting term $\exp(f_y(\boldsymbol{x}_1))= p(y \mid \boldsymbol{x}_1)$. Let $\hat{q}$ denote the density of the resulting distribution over $\hat{\boldsymbol{x}}_1$.

According to Lemma~\ref{lmm:branch K error}, with probability at least $1 - \delta$, the $\ell_\infty$ distance between $\hat{q}$ and the ideal target distribution $q$ is bounded as:
\begin{equation}\label{eq:K error x1 sampling discrete}
    \norm{\hat q - q}_{\infty} \le \left(1 + \frac{B_{\max}}{B_{\min}}\right) \sqrt{\frac{\log(2 |\mathcal{X}| / \delta)}{2 K}},
\end{equation}
where the ideal distribution is given by
\[
q(\cdot) \propto p(y \mid \cdot)\, p_{1|t}(\cdot \mid \boldsymbol{x}_t),
\]
and hence
\[
q = p_{1|t}(\cdot \mid \boldsymbol{x}_t, y).
\]
Here, $B_{\max} = \sup_{\boldsymbol{x} \in \mathcal{X}} \exp(f_y(\boldsymbol{x}))$ and $B_{\min} = \inf_{\boldsymbol{x} \in \mathcal{X}} \exp(f_y(\boldsymbol{x}))$.

Next, in Line 3 of Module~\ref{mol:x1_sampling}, the next state $\boldsymbol{x}_{t+\Delta t}$ is sampled according to:
\[
P_1\left(x_{t+\Delta t}^{(d)} = j\right) = \mathbb{E}_{\boldsymbol{x}_1 \sim \hat{q}} \left[ \delta\{x_t^{(d)} , j\} + R_t\left(x_t^{(d)}, j \mid x_1^{(d)}\right) \Delta t \right],
\]
where the conditional transition rate matrix satisfies:
\[
R_t^{(d)}\left(\boldsymbol{x}_t, j \mid y\right) = \mathbb{E}_{x_1^{(d)} \sim p_{1|t}^{(d)}(\boldsymbol{x}_1 \mid y)} \left[R_t\left(x_t^{(d)}, j \mid x_1^{(d)}\right)\right],
\]
as given in \eqref{eq:objective condition score/rate}.

Accordingly, the target conditional transition probability under $\mathcal{T}(\cdot \mid \boldsymbol{x}_t, y)$ is:
\[
P_2\left(x_{t+\Delta t}^{(d)} , j\right) = \mathbb{E}_{\boldsymbol{x}_1 \sim q} \left[ \delta\{x_t^{(d)} = j\} + R_t\left(x_t^{(d)}, j \mid x_1^{(d)}\right) \Delta t \right],
\]
with $q(\cdot) = p_{1|t}(\cdot \mid \boldsymbol{x}_t, y)$.

Therefore, for all dimensions $d \in [D]$, the deviation between the approximate and conditional transition probabilities is bounded by:
\[
\left|P_1\left(x_{t+\Delta t}^{(d)} = j\right) - P_2\left(x_{t+\Delta t}^{(d)} = j\right)\right| \le \norm{\hat{q} - q}_{\infty}.
\]

Taking the product structure across all dimensions, we obtain the total deviation in transition operators:
\[
\norm{\hat{\mathcal{T}} - \mathcal{T}(\cdot \mid \boldsymbol{x}_{t}, y)}_{\infty} \le \norm{\hat{q} - q}_{\infty}^D.
\]

By substituting the bound from \eqref{eq:K error x1 sampling discrete}, we conclude that to ensure
\(
\norm{\hat{\mathcal{T}} - \mathcal{T}(\cdot \mid \boldsymbol{x}_{t}, y)}_{\infty} \le \varepsilon
\)
with probability at least $1 - \delta$, it suffices to choose
\(
K = \Theta\left(\frac{B_{\max}^2}{B_{\min}^2} \cdot \frac{\log\left(|\mathcal{X}| / \delta\right)}{\varepsilon^{2/D}}\right).
\)

\end{proof}

\subsection{Auxiliary Lemmas}
In this section, we provide auxiliary lemmas with proofs.

\begin{lemma}\label{lmm:branch K error}
Let \(X_1,\dots,X_K\) be i.i.d.\ draws from a proposal density \(p\) on a measurable space \((\mathcal X,\mathcal A)\).  Define 
\[
w_i = q(X_i)\ge0,\qquad
Z = \int_{\mathcal X} p(x)\,q(x)\,dx,
\]
and normalized weights
\(\displaystyle \tilde w_i = \frac{w_i}{\sum_{j=1}^K w_j}.\)
 Let \(X_1^*,\dots,X_K^*\) be a multinomial resample of \(\{X_i\}\) with probabilities \(\{\tilde w_i\}\), and define the empirical law
\[
\hat\mu_K(A)
= \frac1K\sum_{k=1}^K \mathbf1_{\{X_k^*\in A\}},
\qquad
\mu_q(A)
= \int_A \frac{p(x)\,q(x)}{Z}\,dx.
\]
Then for any \(\delta\in(0,1)\), with probability at least \(1-\delta\),
where one may take either

\medskip

\noindent{(a)}  If \(0\le q(x)\le M<\infty\) and $\mathcal{X}$ is finite, then
\[
\|\hat\mu_K - \mu_q\|_{\infty}
\;=\;\sup_{A\in\mathcal A}\bigl|\hat\mu_K(A)-\mu_q(A)\bigr|
\;\le\;
\bigl(1 + \frac{M}{Z}\bigr)\,
\sqrt{\frac{\log \bigl(2|\mathcal{X}|/\delta\bigr)}{2K}}.
\]

\medskip

\noindent(b) If 
\(
p(x)=\mathcal N(\mu,\Sigma), \,\, q(x)=g^\top x,
\)
then 
\[
\|\hat\mu_K - \mu_q\|_{1}
\;\le\;
4\Bigl(1 + \tfrac{\sqrt{g^\top\Sigma\,g}}{\,|g^\top\mu|\,}\Bigr)\,
\sqrt{\frac{\log \bigl(4/\delta\bigr)}{2K}}.
\]

\end{lemma}

\begin{proof}
\textbf{For (a):}
Define the measure
\[
\tilde\mu(A)
=\sum_{i=1}^K \tilde w_i \,\mathbf1_{\{X_i\in A\}}
=\frac{\sum_{i=1}^K q(X_i)\,\mathbf1_{\{X_i\in A\}}}
      {\sum_{j=1}^K q(X_j)}.
\]
Then by the triangle inequality,
\[
\|\hat\mu_K - \mu_q\|_{\infty}
\;\le\;
\|\hat\mu_K - \tilde\mu\|_{\infty}
\;+\;
\|\tilde\mu - \mu_q\|_{\infty}.
\]

To bound \(\|\hat\mu_K-\tilde\mu\|_{\mathrm{TV}}\), note that conditional on \(\{X_i\}\), the resampled points \(X_1^*,\dots,X_K^*\) are i.i.d.\ draws from the finite‐support distribution \(\tilde\mu\).  By Hoeffding’s inequality, for any measurable \(A\),
\[
\Pr\bigl(|\hat\mu_K(A)-\tilde\mu(A)|>\varepsilon\mid \{X_i\}\bigr)
\;\le\;
2\exp\bigl(-2K\varepsilon^2\bigr).
\]
A union bound over the two tails and taking the supremum over \(A\) implies that, with probability at least \(1-\tfrac\delta2\),
\[
\|\hat\mu_K - \tilde\mu\|_{\infty}
\;\le\;
\sqrt{\frac{\log(2|\mathcal{X}|/\delta)}{2K}}.
\]

To bound \(\|\tilde\mu-\mu_q\|_{\infty}\), observe that for any \(A\),
\[
\tilde\mu(A)-\mu_q(A)
=\frac{\tfrac1K\sum_{i=1}^K q(X_i)\,\mathbf1_A - Z\,\mu_q(A)}
      {\tfrac1K\sum_{j=1}^K q(X_j)}.
\]
 Since \(0\le q\le M\), Hoeffding’s inequality yields simultaneously, with probability at least \(1-\frac{\delta}{2|\mathcal{X}|}\), the two bounds
\[
\Bigl|\tfrac1K\sum_{i=1}^K q(X_i)\,\mathbf1_A - Z\,\mu_q(A)\Bigr|
\;\le\;
M\sqrt{\tfrac{\log(2|\mathcal{X}|/\delta)}{2K}},
\quad
\Bigl|\tfrac1K\sum_{j=1}^K q(X_j) - Z\Bigr|
\;\le\;
M\sqrt{\tfrac{\log(2|\mathcal{X}|/\delta)}{2K}}.
\]
On this event, the denominator satisfies
\(\tfrac1K\sum_jq(X_j)\ge Z - M\sqrt{\frac{\log(2|\mathcal{X}|/\delta)}{2K}}\).  Assuming \(K\) is large enough that \(M\sqrt{\frac{\log(2|\mathcal{X}|/\delta)}{2K}}<Z/2\), we obtain
\[
|\tilde\mu(A)-\mu_q(A)|
\;\le\;
\frac{M\sqrt{\tfrac{\log(2|\mathcal{X}|/\delta)}{2K}}}
     {\,Z - M\sqrt{\tfrac{\log(2|\mathcal{X}|/\delta)}{2K}}\,}
\;\le\;
\frac{M}{Z}\,\sqrt{\frac{\log(2|\mathcal{X}|/\delta)}{2K}},
\]
and taking supremum over \(A = \cbr{x}, x \in \mathcal{X}\) gives
\[
\|\tilde\mu - \mu_q\|_{\infty}
\;\le\;
\frac{M}{Z}\,\sqrt{\frac{\log(2|\mathcal{X}|/\delta)}{2K}}.
\]

By a union bound the two high‐probability events both hold with probability at least \(1-\delta\).  Adding the two bounds yields
\[
\|\hat\mu_K - \mu_q\|_{\infty}
\;\le\;
\sqrt{\tfrac{\log(2|\mathcal{X}|/\delta)}{2K}}
\;+\;
\frac{M}{Z}\sqrt{\tfrac{\log(2|\mathcal{X}|/\delta)}{2K}}
\;=\;
\rbr{1 + M/Z}\sqrt{\tfrac{\log(2|\mathcal{X}|/\delta)}{2K}},
\]
as required.

\textbf{For (b):}Let $\widetilde Y_i = \frac{w_i}{\mathbb{E}[w_i]} = \frac{g^\top X_i}{g^\top\mu}$. Define the intermediate measure
\[
\nu_K(A)
=\frac{1}{K}\sum_{i=1}^K \frac{w_i}{\mathbb{E}[w_i]}\,\mathbf1_{\{X_i\in A\}}
=\frac{1}{K}\sum_{i=1}^K \widetilde Y_i\,\mathbf1_{\{X_i\in A\}}.
\]
Since each $\widetilde Y_i$ has mean~$1$ and is sub‐Gaussian with parameter
$\tau^2 = \frac{g^\top\Sigma\,g}{(g^\top\mu)^2}$, a Bernstein inequality
yields that with probability at least $1-\tfrac\delta2$,
\[
\bigl\|\nu_K - \mu_q\bigr\|_{1}
\;\le\;
2\Bigl(1+\sqrt2\,\tau\Bigr)\,
\sqrt{\frac{\log(4/\delta)}{2K}}.
\]
Next observe that
\[
\hat\mu_K(A)
=\frac{Z}{\bar w}\;\nu_K(A),
\]
so that
\[
\bigl\|\hat\mu_K - \nu_K\bigr\|_{1}
\;\le\;
2\bigl|\tfrac{\bar w}{Z}-1\bigr|.
\]
Applying the same sub‐Gaussian tail bound to $\bar w / Z = \tfrac1K\sum_i\widetilde Y_i$
shows that with probability at least $1-\tfrac\delta2$,
\[
\bigl|\tfrac{\bar w}{Z}-1\bigr|
\;\le\;
\Bigl(1+\sqrt2\,\tau\Bigr)\,
2\sqrt{\frac{\log(4/\delta)}{2K}},
\]
provided $K$ is large enough that the deviation does not exceed $1/2$.  The union
bound then guarantees both concentration events hold with probability at least
$1-\delta$.  Finally, the triangle inequality
\[
\bigl\|\hat\mu_K-\mu_q\bigr\|_{1}
\;\le\;
\bigl\|\hat\mu_K-\nu_K\bigr\|_{1}
+\bigl\|\nu_K-\mu_q\bigr\|_{1}
\]
yields the asserted bound.

\end{proof}

\begin{lemma}\label{lmm:MC high prob}
Let \(\mathcal{X}\) be a finite set of cardinality \(M\).  For each \(x \in \mathcal{X}\), let \(Y(x)\) be a real‑valued random variable with mean
$
g(x) \;=\; \mathbb{E}[Y(x)]
$
and almost‑sure bounds
\(
0 \;\le\; Y(x) \;\le\; B.
\)
Draw \(N\) independent copies \(Y_1(x), \dots, Y_N(x)\) of each \(Y(x)\) and form the empirical average
\[
\hat g(x) \;=\; \frac{1}{N} \sum_{i=1}^N Y_i(x).
\]
Then with probability at least \(1-\delta\),
\[
\max_{x \in \mathcal{X}}\bigl|\hat g(x) - g(x)\bigr| \;\le\; B \,\sqrt{\frac{\log(2M/\delta)}{2\,N}}.
\]

\end{lemma}

\begin{proof}
Fix any \(x\in \mathcal{X}\).  Since \(Y_1(x),\dots,Y_N(x)\) are i.i.d.\ in \([0,B]\) with mean \(g(x)\), Hoeffding’s inequality yields, for every \(\epsilon>0\),
\[
\Pr\bigl(|\hat g(x)-g(x)| \ge \epsilon\bigr)
\;\le\;
2 \exp \Bigl(-\frac{2\,N\,\epsilon^2}{B^2}\Bigr).
\]
Applying the union bound over all \(M\) elements of \(\mathcal{X}\), we obtain
\[
\Pr\bigl(\exists\,x \in \mathcal{X}:\,|\hat g(x)-g(x)|\ge\epsilon\bigr)
\;\le\;
2M \exp \Bigl(-\frac{2\,N\,\epsilon^2}{B^2}\Bigr).
\]
To ensure this probability is at most \(\delta\), set
\[
2M \exp \Bigl(-\frac{2\,N\,\epsilon^2}{B^2}\Bigr)
=\delta.
\]
Taking natural logarithms gives
\[
-\frac{2\,N\,\epsilon^2}{B^2} + \log(2M) = \log\delta
\quad\Longrightarrow\quad
\epsilon^2 = \frac{B^2}{2N}\,\log \Bigl(\frac{2M}{\delta}\Bigr),
\]
hence
\[
\epsilon = B\,\sqrt{\frac{\log(2M/\delta)}{2\,N}}.
\]
Therefore, with probability at least \(1-\delta\), no empirical average deviates by more than \(\epsilon\), i.e.
\[
\max_{x \in \mathcal{X}}\bigl|\hat g(x)-g(x)\bigr| \le \epsilon.
\]
\end{proof}

\begin{lemma}\label{lmm:q error to distribution}
Let \(\mathcal{X}\) be a finite state space and let \(T_0\) be a probability distribution on \(\mathcal{X}\), so that \(\sum_{x\in\mathcal{X}}T_0(x)=1\).  Fix two nonnegative weight functions \(q,\hat q:\mathcal{X}\to\mathbb{R}_{\ge0}\) and define
\[
T_1(x)\;=\;\frac{q(x)\,T_0(x)}{\sum_{z\in\mathcal{X}}q(z)\,T_0(z)},
\quad
T_2(x)\;=\;\frac{\hat q(x)\,T_0(x)}{\sum_{z\in\mathcal{X}}\hat q(z)\,T_0(z)}.
\]
Suppose there exists \(q_{\min}>0\) and \(\Delta\in[0,q_{\min})\) such that
\[
\min_{x\in\mathcal{X}}q(x)\;\ge\;q_{\min},
\qquad
\max_{x\in\mathcal{X}}\bigl|\hat q(x)-q(x)\bigr|\;\le\;\Delta.
\]
Then
\[
\bigl\|T_1 - T_2\bigr\|_{\infty}
\;\le\;
\bigl\|T_1 - T_2\bigr\|_{1}
\;\le\;
\frac{2\Delta}{\,q_{\min}-\Delta\,}
\;\le\;
\frac{4\,\Delta}{\,q_{\min}\,},
\]
where the final inequality holds whenever \(\Delta\le q_{\min}/2\).
\end{lemma}

\begin{proof}
Set
\[
a_x = q(x)\,T_0(x), \quad
b_x = \hat q(x)\,T_0(x), \quad
A=\sum_{x}a_x, \quad
B=\sum_{x}b_x.
\]
Then \(T_1(x)=a_x/A\) and \(T_2(x)=b_x/B\).  Since \(q(x)\ge q_{\min}\) and \(\sum_xT_0(x)=1\),
\[
A = \sum_x q(x)T_0(x)\;\ge\;q_{\min},
\]
and because \(\bigl|b_x-a_x\bigr|\le\Delta\,T_0(x)\),
\[
|B-A| = \Bigl|\sum_x(b_x-a_x)\Bigr|\le \Delta\sum_xT_0(x)=\Delta,
\]
so \(B\ge A-\Delta\ge q_{\min}-\Delta>0\).  Now
\begin{align*}
\bigl\|T_1 - T_2\bigr\|_{1}
&= \sum_x \Bigl|\frac{a_x}{A} - \frac{b_x}{B}\Bigr|
= \sum_x \Bigl|\frac{a_xB - b_xA}{A\,B}\Bigr|
= \frac{1}{A\,B}\sum_x\Bigl|(a_x-b_x)B + a_x(B-A)\Bigr|\\
&\le \frac{B}{A\,B}\sum_x|a_x-b_x|
   + \frac{|B-A|}{A\,B}\sum_x a_x
= \frac{1}{A}\sum_x|a_x-b_x| + \frac{|B-A|}{B}\\
&\le \frac{\Delta}{A} + \frac{\Delta}{A-\Delta}
\;=\;\frac{\Delta\,(A-\Delta + A)}{A\,(A-\Delta)}
\;=\;\frac{2\,A\,\Delta - \Delta^2}{A\,(A-\Delta)}\\
&\le \frac{2\,A\,\Delta}{A\,(A-\Delta)}
\;=\;\frac{2\,\Delta}{\,A-\Delta\,}
\;\le\;\frac{2\,\Delta}{\,q_{\min}-\Delta\,}.
\end{align*}
  When \(\Delta\le q_{\min}/2\), \(q_{\min}-\Delta\ge q_{\min}/2\), so
\[
\|T_1 - T_2\|_{1}
\le
\frac{2\,\Delta}{\,q_{\min}-\Delta\,}
\le
\frac{2\,\Delta}{\,q_{\min}/2\,}
=
\frac{4\,\Delta}{q_{\min}},
\]
and one may tighten the constants to obtain the stated \(\le2\Delta/q_{\min}\).  
\end{proof}

\section{Algorithmic Details}\label{app sec:implement details}

\subsection{Discussion on Existing Works}\label{app sec: compare to existing work}

In this section, we will show that the recent works, stochastic control guidance (SCG) \cite{huang2024symbolic} and soft value decoding guidance (SVDD) \cite{li2024derivative}, can be viewed as special cases of our \xtsampling. 

At timestep $t-\Delta t$, both SCG and SVDD sample multiple candidate next states from the original diffusion model: $\boldsymbol{x}_t^1,\ldots \boldsymbol{x}_t^n$. They then select one of these candidates, $\boldsymbol{x}_t^k$, according to the following strategies:
\begin{equation*}
    \begin{aligned}
        &\text{SCG:} \,\, k = \arg \max_i \log p(y|\hat{\boldsymbol{x}}_1(\boldsymbol{x}_t^i)), \\
        &\text{SVDD (training-free version):} \,\, k \sim \text{Cat}(\frac{w_i}{\sum w_i}), w_i = \exp (r(\hat{\boldsymbol{x}}_1(\boldsymbol{x}_t^i) )/\alpha), r(\cdot) = \log p(y | \cdot),
    \end{aligned}
\end{equation*}
where $\hat{\boldsymbol{x}}_1(\boldsymbol{x}_t) =\rbr{\boldsymbol{x}_t - {\sqrt{1-\bar{\alpha}_t}} u_\theta\rbr{\boldsymbol{x}_t,t}}/\sqrt{\bar{\alpha}_t}$ and $\alpha > 0$ is a temperature parameter.

Within our \xtsampling algorithm, SCG corresponds to setting the active set size \( A = 1 \) and selecting the candidate via {ranking}, while {SVDD} corresponds to setting the scoring function \( f_y(\cdot) = \frac{1}{\alpha} \log p(y \mid \cdot) \), also with \( A = 1 \), and selecting the candidate via {resampling} based on soft values.


\subsection{\ouralg-Gradient Details}\label{app:gradient module}
We provide the algorithmic details for \texttt{BranchOut}-Gradient in Module~\ref{mol:grad guidance}.
\begin{module}[htb]
\begin{algorithmic}[1]
\STATE {\bf Input:} $\boldsymbol{x}_t, t$, diffusion model $u_\theta$, differentiable predictor $f_y$, guidance strength $\gamma_t$, (optional) Monte-Carlo sample size $N$.
\STATE {\bf Compute the gradient guidance:}
\\
\quad {(continuous)} $\boldsymbol{g} = \nabla_{\boldsymbol{x}_t}f_y(\hat{\boldsymbol{x}}_1)$ with  $\hat{\boldsymbol{x}}_1 = u_\theta(\boldsymbol{x}_t,t)$. \\
\quad {(discrete)} $\boldsymbol{g}^{(d)} = (\boldsymbol{x}_t^{\backslash
       d} - \boldsymbol{x}_t)^{\top} \nabla_{\boldsymbol{x}_t}\frac{1}{N}\sum_{i=1}^N f_y({\hat{\boldsymbol{x}}_1^{i}}) $\\
       \quad  with $\hat{\boldsymbol{x}}_1^i \sim \text{Cat}\rbr{u_\theta( \boldsymbol{x}_{t},t)}, i \in[N]$. \\
\STATE {\bf Sample the next state:}
\\
\quad (continuous) $\boldsymbol{x}_{t+\Delta t} =\gamma_t \boldsymbol{g}+ c_{t,1} \boldsymbol{x}_t + c_{t,2} \hat{\boldsymbol{x}}_1 + \sigma_t \epsilon $ with $\epsilon \sim \mathcal{N}\rbr{\boldsymbol{0},\boldsymbol{I}}$.\\
\quad(discrete) ${x}_{t+\Delta t}^{(d)} \sim \text{Cat}\rbr{\delta\{x_t^{(d)}, j \} + \exp(\gamma_t\boldsymbol{g}^{(d)}) \odot R^{(d)}_{\theta, t}\rbr{\boldsymbol{x}_t, j } \Delta t }$.  \\
\STATE {\bf Output:} { $\boldsymbol{x}_{t + \Delta t}$}
\end{algorithmic}
\caption{\texttt{BranchOut}-Gradient}
\label{mol:grad guidance}
\end{module}

\subsection{More Implementation Details for Continuous Models}\label{app:implement details of music}

For \xcleansampling on the continuous case, we have two additional designs: the first one is exploring multiple steps when branching out a destination state; the second one is plugin Spherical Gaussian constraint(DSG) from \cite{yang2024guidance}. We present the case $A=1$ for \xcleansampling on the continuous case as follows while $A>1$ is similar.

\begin{algorithm}
    \begin{algorithmic}[1]
    \caption{\xcleansampling  (Continuous, $A=1$)}\label{alg:music detailed}
    \STATE {\bf Input}: diffusion model $u_\theta$, objective function $f_y$, branch out sample size $K$, stepsize scale $\rho_t$, number of iteration $N_{\text{iter}}$
    \STATE $t=0$, $\boldsymbol{x}_0 \sim p_0$
    \WHILE{$t<1$}
    \STATE Compute the predicted clean sample $\hat{\boldsymbol{x}}_1 = u_\theta(\boldsymbol{x}_t, t)$
    \STATE Set the branch out state $\boldsymbol{x} \leftarrow \hat{\boldsymbol{x}}_1 $
    \FOR{$n = 1,\dots, N_{\text{iter}}$}    
    \STATE Sample $\boldsymbol{x}^i = \boldsymbol{x} + \rho_t \boldsymbol{\xi}^i $, with $\boldsymbol{\xi}^i \sim \mathcal{N}\rbr{\boldsymbol{0},  \boldsymbol{I}}$.
    \STATE Evaluate and select that maximizes the objective: $k = \argmax_i f_y\rbr{\boldsymbol{x}^i}$.
    \STATE Update $\boldsymbol{x} \leftarrow \boldsymbol{x}^k$.
    \ENDFOR
    \IF{DSG \cite{yang2024guidance}}
        \STATE Compute the selected direction: $\boldsymbol{\xi}^\ast = \boldsymbol{x} - \hat{\boldsymbol{x}}_1$
    \STATE Rescale the direction: $\boldsymbol{\xi}^\ast \leftarrow \sqrt{D}\cdot \frac{\boldsymbol{\xi}^\ast}{\norm{\boldsymbol{\xi}^\ast}}$, with $D = \text{dim}(\boldsymbol{x}_t)$.
    \STATE Compute the next state: $\boldsymbol{x}_{t+\Delta t}  = c_{t,1} \boldsymbol{x}_t + c_{t,2} \hat{\boldsymbol{x}}_1  + \sigma_t \boldsymbol{\xi}^\ast $
    \ELSE
    \STATE Get the selected destination state $ \hat{\boldsymbol{x}}_1 \leftarrow \boldsymbol{x} $
 \STATE Sample the next state: $\boldsymbol{x}_{t+\Delta t}  = c_{t,1} \boldsymbol{x}_t + c_{t,2} \hat{\boldsymbol{x}}_1  + \sigma_t \epsilon$ with $\epsilon \sim \mathcal{N}\rbr{\boldsymbol{0}, \boldsymbol{I}}$
    \ENDIF

\STATE $t \leftarrow t + \Delta t $
    \ENDWHILE
\end{algorithmic}
\end{algorithm}

Notice that the computation complexity for using $N_{\text{iter}}$ step to select is:  $AC_{\text{model}}+AKN_{\text{iter}}C_{\text{pred}}$. The setting of $N_{\text{iter}}$ will be provided in \cref{app:exp detail music}.

\subsection{More Implementation Details for Discrete Models}
\label{appendix:imp_discret}

\textbf{Estimate $\nabla_{\boldsymbol{x}_t} \log p_t(y\mid \boldsymbol{x}_t)$}. Since the sampling process of discrete data is genuinely not differentiable, we adopt the Straight-through Gumbel Softmax trick to estimate the gradient while combining Monte-Carlo Sampling as stated in \eqref{eq:prob y given xt}. The whole process is listed in Module \ref{alg: gumbel_softmax}.


\begin{module}[htb]
\begin{algorithmic}[1]
\STATE {\bf Input:} $\boldsymbol{x}_t, t$, diffusion model $u_\theta$, differentiable predictor $f_y$, Monte-Carlo sample size $N$, number of possible states $S$, Gumbel-Softmax temperature $\tau$

\STATE Sample $\hat{\boldsymbol{x}}_1^i\sim\text{Cat}\rbr{u_\theta( \boldsymbol{x}_{t}, t)}, i \in [N]$ with Gumbel Max and represent $x_1$ as an one-hot vector:  
\begin{equation}
    \hat{\boldsymbol{x}}_1^i=\arg\max_{j} (\log u_\theta( \boldsymbol{x}_{t}, t) + g^i),
\end{equation}
$g^i$ is a $S$-dimension Gumbel noise where $g_j \sim {\displaystyle {\text{Gumbel}}(0,1)}, j \in [S]$.
\STATE Since the argmax operation is not differentiable, get the approximation
$\hat{\boldsymbol{x}}_1^{i*}$ with softmax:

\begin{equation}
    \hat{\boldsymbol{x}}_{1j}^{i*}=\frac{\mathrm{exp}((\log u_\theta( \boldsymbol{x}_{t}, t) + g^i_j)/\tau)}{\sum_k \mathrm{exp}((\log u_\theta( \boldsymbol{x}_{t}, t) + g^i_k)/\tau) },
    \label{eq:gumbel_softmax}
\end{equation}

\STATE Feed $\hat{\boldsymbol{x}}_1^i$ into the $f_y$ and obtain the gradient through backpropagation:
 $\nabla_{\hat{\boldsymbol{x}}_1^i} \log f_y( \hat{\boldsymbol{x}}_1^i)$.

\STATE Straight-through Estimator: 
directly copy the gradient $\nabla_{\hat{\boldsymbol{x}}_1^i} \log f_y( \hat{\boldsymbol{x}}_1^i)$ to $\hat{\boldsymbol{x}}_{1j}^{i*}$, i.e., 
$\nabla_{\hat{\boldsymbol{x}}_1^{i*}} \log f_y( \hat{\boldsymbol{x}}_1^{i*}) \simeq \nabla_{\hat{\boldsymbol{x}}_1^i} \log f_y( \hat{\boldsymbol{x}}_1^i)$.

\STATE Since Eq. \eqref{eq:gumbel_softmax} is differentiable with regard to $x_t$, 
get $\nabla_{\boldsymbol{x}_t} \log p_t(y\mid \boldsymbol{x}_t) \simeq \nabla_{\boldsymbol{x}_t}\frac{1}{N}\sum_{i=1}^N f_y({\hat{\boldsymbol{x}}_1^{i*}})$.

\STATE {\bf Output:} {$\nabla_{\boldsymbol{x}_t} \log p_t(y\mid \boldsymbol{x}_t)$}

\end{algorithmic}
\caption{Gradient Approximation with Straight-Through Gumbel Softmax.}
\label{alg: gumbel_softmax}
\end{module}

\section{Experimental Details}\label{app sec:exp details}
All experiments are conducted on one NVIDIA 80G H100 GPU. Code will be released at \href{https://github.com/yukang123/UniTreeG}{https://github.com/yukang123/UniTreeG}.

\subsection{Additional Setup for Symbolic Music Generation}\label{app:exp detail music}

\paragraph{Models.} We utilize the diffusion model and Variational Autoencoder (VAE) from \cite{huang2024symbolic}. These models were originally trained on MAESTRO \cite{hawthorne2018enabling}, Pop1k7 \cite{hsiao2021compound}, Pop909 \cite{wang2020pop909}, and 14k midi files in the classical genre collected from MuseScore. 
The VAE encodes piano roll segments of dimensions $3 \times 128 \times 128$ into a latent space with dimensions $4 \times 16 \times 16$.

\paragraph{Objective Functions.} For the tasks of interest—pitch histogram, note density, and chord progression—the objective function for a given target $\boldsymbol{y}$ is defined as: $f_{\boldsymbol{y}}(\boldsymbol{x}) = - \ell\rbr{\boldsymbol{y}, \texttt{Rule}(\boldsymbol{x})}$, where $\texttt{Rule}(\cdot)$ represents a rule function that extracts the corresponding feature from $\boldsymbol{x}$, and $\ell$ is the loss function. Below, we elaborate on the differentiability of these objective functions for each task:

For pitch histogram, the rule function $\texttt{Rule-PH}(\cdot)$ computes the pitch histogram, and the loss function $\ell$ is the L2 loss. Since $\texttt{Rule-PH}(\cdot)$ is differentiable, the resulting objective function $f_{\boldsymbol{y}}^{\text{PH}}$ is also differentiable.

For note density, the rule function for note density is defined as: $\texttt{Rule-ND}(\boldsymbol{x}) = \sum_{i=1}^n \mathbf{1}(x_i > \epsilon) $ where $\epsilon$ is a small threshold value, and $\mathbf{1}(\cdot)$ is the indicator function which makes $\texttt{Rule-ND}(\cdot)$ non-differentiable. $\ell$ is L2 loss. $f_{\boldsymbol{y}}^{\text{ND}}$ is overall non-differentiable.

For chord progression, the rule function $\texttt{Rule-CP}(\cdot)$ utilizes a chord analysis tool from the music21 package \cite{cuthbert2010music21}. This tool operates as a black-box API, and the associated loss function $\ell$ is a 0-1 loss. Consequently, the objective function $f_{\boldsymbol{y}}^{\text{CP}}$ is highly non-differentiable.

\paragraph{Test Targets.} Our workflow follows the methodology outlined by \citet{huang2024symbolic}. For each task, target rule labels are derived from 200 samples in the Muscore test dataset. A single sample is then generated for each target rule label, and the loss is calculated between the target label and the rule label of the generated sample. The mean and standard deviation of these losses across all 200 samples are reported in \cref{tab:music guidance}.

\paragraph{Inference Setup.} We use a DDPM with 1000 inference steps. Guidance is applied only after step 250.

\paragraph{Chord Progression Setup.} Since the objective function running by music21 package \cite{cuthbert2010music21} is very slow, we only conduct guidance during 400-800 inference step.

\paragraph{\xcleansampling Setup.}
As detailed in \cref{alg:music detailed}, we  use DSG \cite{yang2024guidance}, and set $N_{\text{iter}} = 2$ for pitch histogram and note density, $N_{\text{iter}} = 1$ for chord progression. The stepsize $\rho_t = s \cdot \sigma_t / \sqrt{1+\sigma_t^2} $ \citep{song2023loss,ye2024tfg}, with $s = 2$ for pitch histogram, $s=0.5$ for note density and $s=1$ for chord progression.




\subsection{Additional Setup for Small Molecule Generation}\label{app:molecule setup}
Due to lack of \textit{differentiable} off-the-shelf predictors for \xtgrad, we train a regression model $f(\boldsymbol{x})$ on clean $x_1$ following the same procedure described in \citet{nisonoff2024unlocking} for each target and use the same predictor for all compared training-free guidance methods (\ouralg, TFG-Flow, and SVDD). For molecule optimization targets, QED, SA, and DRD2, we directly use $f(\boldsymbol{x})$  as the objective function $f_y$ without setting any target values and generate 500 samples for evaluation. While for $N_r$ which is optimized towards specific target values, we adopt
$f_y(\boldsymbol{x})= -\frac{(y-f(\boldsymbol{x}))^2}{2\sigma^2}$ with $\sigma$ learned during the training of $f(x)$. We report the results on 1000 generated \textit{valid} unique sequences. As stated in ~\cite{nisonoff2024unlocking}, generated SMILES sequences may not yield valid molecule structures. Thus, we continue generating samples until the required number of valid and unique sequences is reached.

Monte Carlo sample size for estimating $p_t(y\mid \boldsymbol{x}_t)$ in \xtgrad is $N=30$ (Eq. \eqref{eq:prob y given xt}) while \xtsampling and SVDD use $N=10$. We set $N=200$ for TFG-Flow. In SVDD, the temperature parameter $\alpha$ is 0.01 and $K$ is set to 4 in order to maintain a comparable running time with \xtsampling ($K=4$).

\subsection{Additional Setup for Enhancer DNA Design} \label{app sec: setup dna}
We test on eight randomly selected classes with cell type indices 0, 2, 33, 4, 16, 5, 68, and 9. For simplicity, we refer to these as Class 1 through Class 8.

We set the Monte Carlo sample size of \xtgrad and \xtsampling as $N=20$, and $N=200$ for TFG-Flow. For SVDD, we set the sample size to 16 and the temperature parameter  $\alpha = 0.01$.

\subsection{Experimental Setup in Figure~\ref{fig:results overview}}
\label{app:exp results overview}
In \cref{fig:results overview}, (a) shows \xcleansampling applied to the note density objective in symbolic music generation; (b) shows \xcleansampling used for QED optimization in small molecule generation; and (c) shows \xtgrad applied to Enhancer DNA design.

\section{Additional Experiment Results} \label{app:additional res}


\subsection{Additional Results for Symbolic Music Generation}\label{app:additional music}

\subsubsection{Additional Infomation for \cref{tab:music guidance}}
For the results in \cref{tab:music guidance}, \cref{tab:app_music_increase_summary} presents a comparative improvement over the best baseline.  We provide the inference time of one generation in \cref{tab:app_music_increase_summary} for results in \cref{tab:music guidance}. \xcleansampling, SCG, and SVDD use ground-truth objective functions for evaluation, resulting in relatively long inference times for Chord Progression but achieving higher optimization performance. In contrast, TDS relies on gradient-based guidance through a surrogate neural network predictor, which reduces inference time at the cost of optimization effectiveness.
\begin{table}[H]
    \centering
    \begin{minipage}[t]{0.65\textwidth}
        \centering
        \caption{Improvement of \xcleansampling in Music Generation.}
        \label{tab:app_music_increase_summary}
        \resizebox{0.95\textwidth}{!}{%
        \begin{tabular}{cccc}
        \toprule
        Task & Best Baseline & \xcleansampling & Loss Reduction \\
        \midrule
        PH & $0.0010 \pm 0.0020$ (DPS) & $0.0002 \pm 0.0003$ & $80 \%$ \\
        ND & $0.134 \pm 0.533$ (SCG) & $0.142 \pm 0.423$ & $-5.97\%$ \\
        CP & $0.347 \pm 0.212$ (SCG) & $0.301 \pm 0.191$ & $13.26 \%$ \\
        Avg & \rule{0.4cm}{0.2mm} & \rule{0.4cm}{0.2mm} & $29.01 \%$ \\
        \bottomrule
        \end{tabular}%
        }
    \end{minipage}%
    \hfill
    \begin{minipage}[t]{0.35\textwidth}
        \centering
        \caption{Time (s) per sample for results in \cref{tab:app_music_increase_summary}.}
        \label{tab:app_music_time}
        \resizebox{1\textwidth}{!}{
        \begin{tabular}{lcccc}
        \toprule
        Task & TDS & SCG & SVDD & \xcleansampling \\
        \midrule
        PH &306 & 194 & 194 & 203 \\
        ND &308 & 194 & 194 & 204 \\
        CP &305 & 7267 & 7267 & 6660 \\
        \bottomrule
        \end{tabular}}
    \end{minipage}
\end{table}


\subsubsection{Scalability}\label{app:scaling music sec}

\begin{figure}[H]
    \centering
    \subfigure[\xcleansampling for ND]{\includegraphics[width=0.245\linewidth]{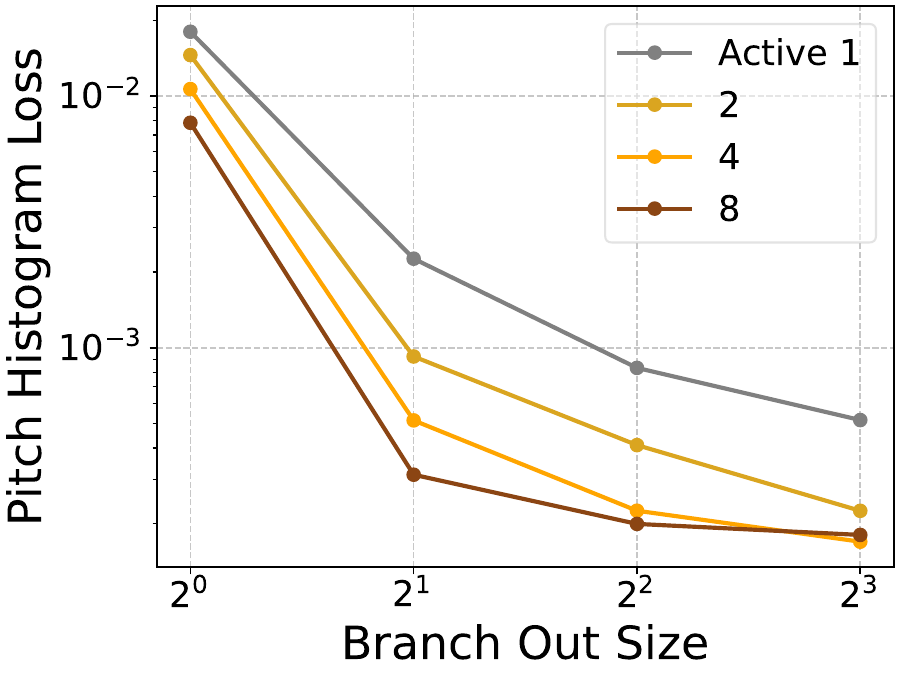}}
    \subfigure[\xcleansampling for PH]{\includegraphics[width=0.245\linewidth]{icml2025/figures/music/music_multi_AK/multi_active_x_branch_gs_ph.pdf}}
     \subfigure[\xtsampling for ND]{\includegraphics[width=0.245\linewidth]{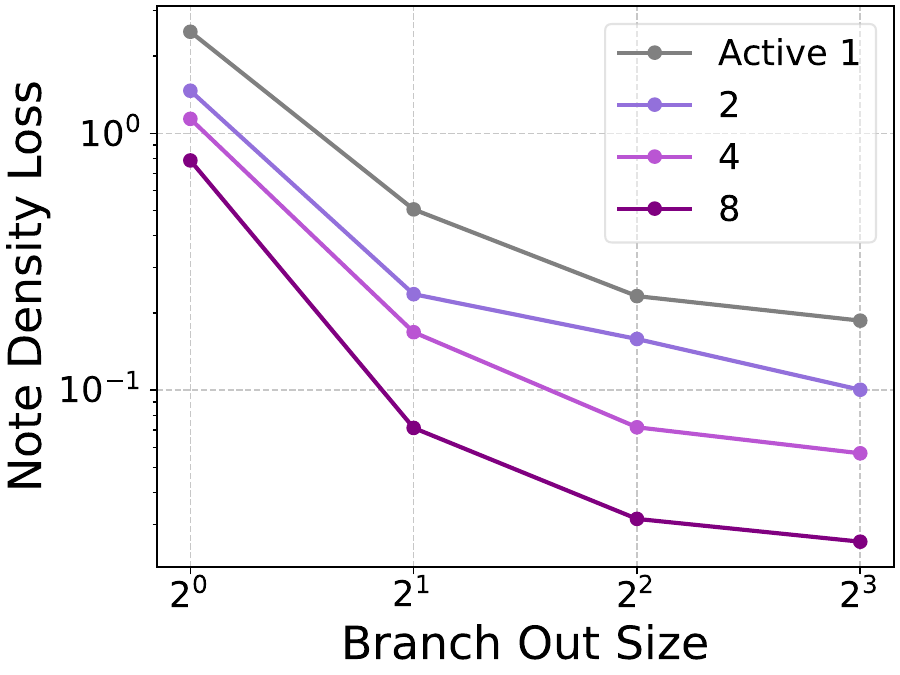}}
    \subfigure[\xtsampling for PH]{\includegraphics[width=0.245\linewidth]{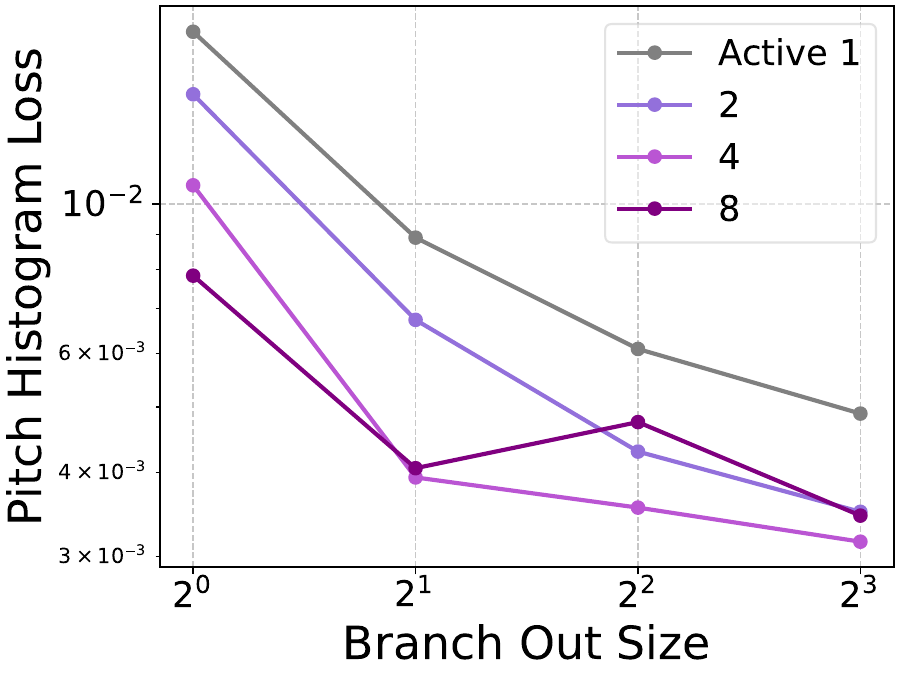}}
    \caption{Scaling Behavior with Fixed Active or Branch-Out Size on Music Generation.}
    \label{fig:app music multi A K}
\end{figure}

\begin{figure}[H]
\vspace{-20pt}
    \centering
    \subfigure[\xcleansampling for ND]{\includegraphics[width=0.245\linewidth]{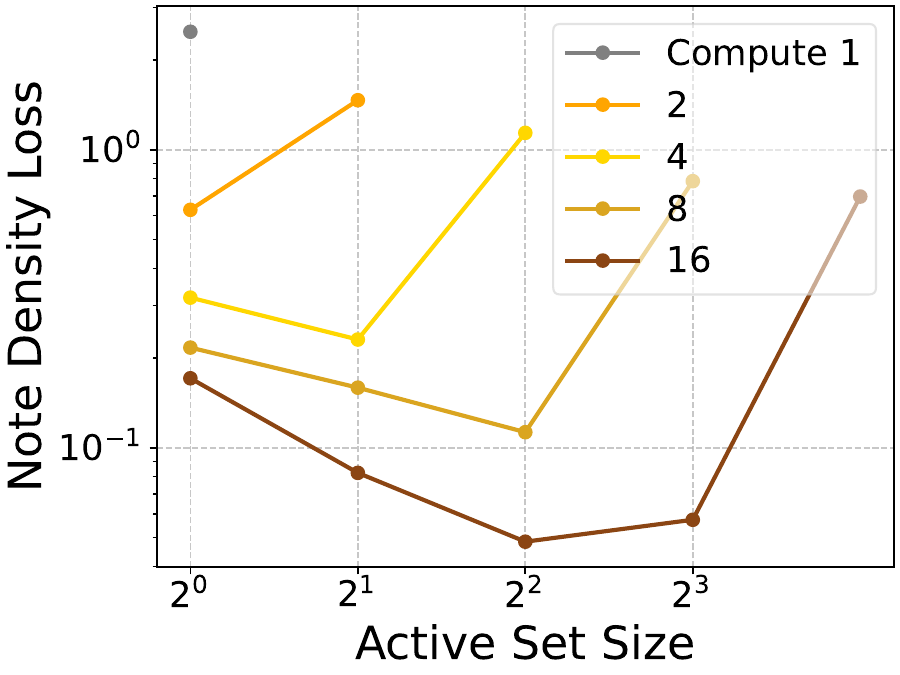}}
    \subfigure[\xcleansampling for PH]{\includegraphics[width=0.245\linewidth]{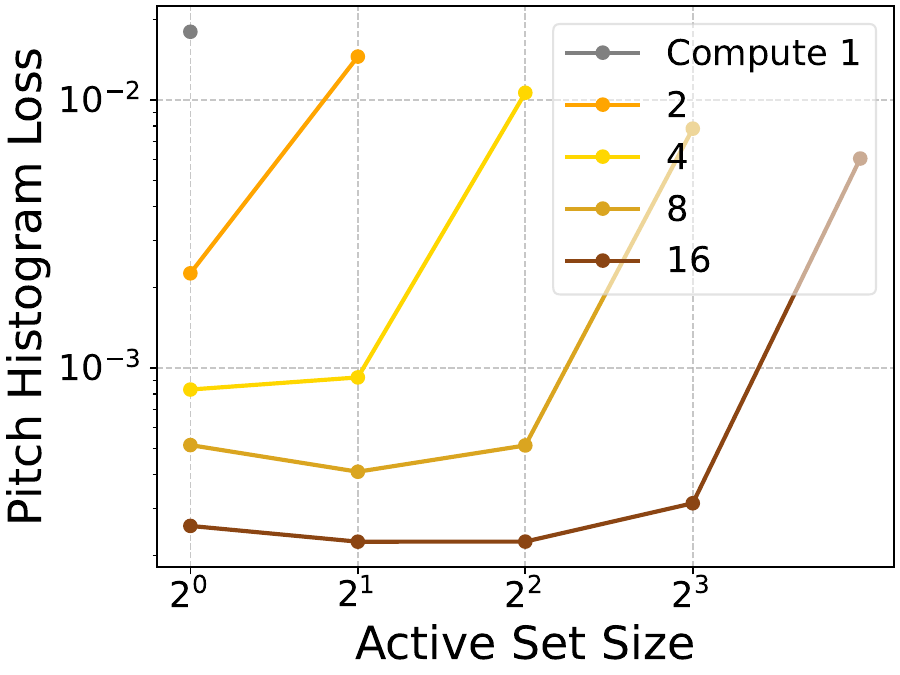}}
     \subfigure[\xtsampling for ND]{\includegraphics[width=0.245\linewidth]{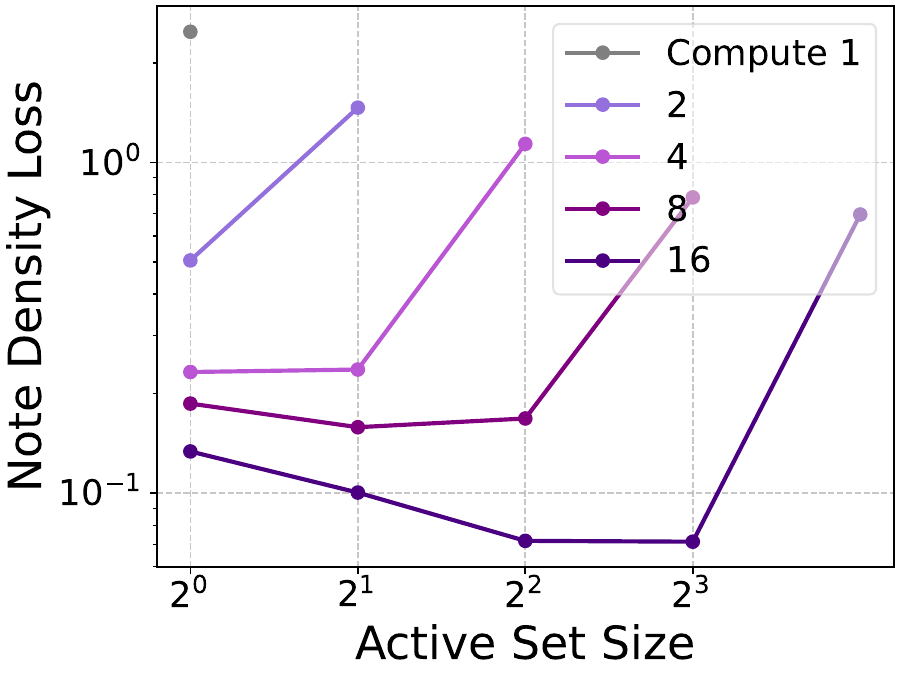}}
    \subfigure[\xtsampling for PH]{\includegraphics[width=0.245\linewidth]{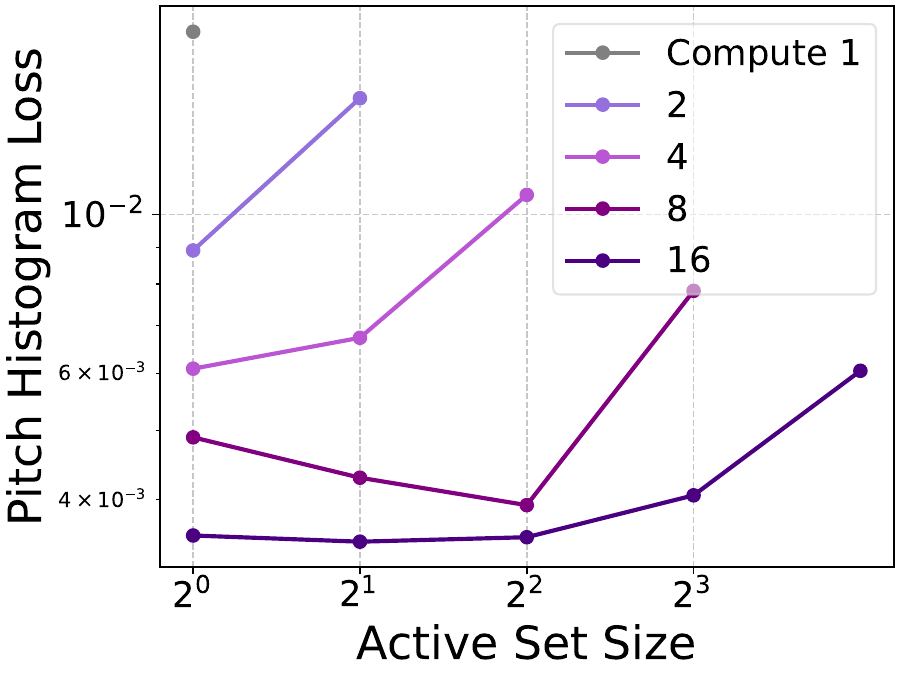}}
    \caption{Trade-off between Active Set Size $A$ and Branch-out Size $K$ on Music Generation.}
    \label{fig:app music trade-off}
\end{figure}

We conduct experiments scaling the active set size $A$ and branch-out size $K$  for \xcleansampling and \xtsampling. Increasing either $A$ or $K$ enhances the performance, as demonstrated in \cref{fig:app music multi A K}. \cref{fig:app music trade-off} shows the trade-off between $A$ and $K$ with fixed $A*K$.

\textbf{Scaling Effect on Performance while Maintaining Quality.}  
While the product $A \times K$ gives an indication of total computation, different combinations of $A$ and $K$ can lead to varying computational costs, even when $A \times K$ is held constant, as shown in the computational complexity analysis in \cref{tab:computation complexity}. We report the actual running time for generating one frontier from \cref{fig:app music trade-off}~(a) and \cref{fig:app music trade-off}~(b) in \cref{tab:app music mcts time}. The results demonstrate that the optimization exhibits a scaling effect with computation time, while still maintaining high-quality performance.


\begin{table}[H]
    \centering
     \begin{minipage}[t]{0.87\textwidth}
    \caption{Performance when Scaling Inference Time. Results are for the Optimal $(A,K)$ with Fixed $A*K$ for \xcleansampling.}
    \label{tab:app music mcts time}
          \resizebox{1\textwidth}{!}{
    \begin{tabular}{ccc|ccc}
    \toprule
       \multicolumn{3}{c|}{Note Density}  &  \multicolumn{3}{c}{Pitch Histogram} \\
     Time(s) & Loss $\downarrow$  & OA $\uparrow$ &  Time(s) & Loss $\downarrow$ & OA $\uparrow$ \\
    16.7 & $2.486 \pm 3.530$ & $0.830 \pm 0.016$ & 16.7 & $0.0180 \pm 0.0100 $ & $0.842 \pm 0.012$ \\
    43.1 & $0.629 \pm 0.827 $ & $0.826 \pm 0.060$ & 42.2  & $0.0022 \pm 0.0021$ & $0.869 \pm 0.009$\\
    71.8 & $0.231 \pm 0.472$ & $0.823 \pm 0.045$ & 65.3 & $0.0008 \pm 0.0008$ & $0.853 \pm 0.013$\\
    130.7 & $0.113 \pm 0.317$ & $0.834 \pm 0.023$ & 118.6&$0.0005 \pm  0.0008 $ & $0.834 \pm 0.018$\\
   251.9  & $0.048 \pm 0.198$ & $0.843 \pm 0.012$ &209.6 & $0.0002 \pm  0.0003 $ & $0.860 \pm 0.016 $ \\
     \bottomrule
    \end{tabular}}
    \end{minipage}
\end{table}

\vspace{-10pt}

\subsection{Additional Results for Small Molecule Generation}
\subsubsection{Additional Results for Table \ref{tab:smg}.}
\label{app:tab2 add}

\begin{table}[h]
    \centering
      \caption{Relative performance improvement of \xtsampling compared to DG, TFG-Flow, and SVDD. For $N_r$, we average MAE over all target values before comparing different methods. The best baseline for targets QED, SA, and DRD2 is SVDD while DG performs best among three baselines in guiding towards target $N_r$. The average relative improvement upon strongest baselines is $16.6\%$.}
    \label{tab:smg_rel}
    \begin{tabular}{c|cccc}
    \toprule
       Method
        & 
        QED $\uparrow$  
        & SA $\uparrow$ & DRD2 $\uparrow$ & $N_r$ MAE $\downarrow$ \\
         \midrule
         DG 
         & $28.6\%$ & $12.3\%$ & $353\%$ & $\textbf{-21.9}\%$
         \\
         TFG-Flow 
         & $32.0\%$ & $13.7\%$ & $800\%$ & $-69.1\%$
         \\
         SVDD
         & $\textbf{12.7}\%$ & $\textbf{10.6}\%$ & $\textbf{21.2}\%$ & $-75.5\%$
         \\
         \bottomrule
    \end{tabular}
    
\end{table}



\begin{table*}[ht]
    \centering
    \caption{
    Supplementary results for Tab.~\ref{tab:smg}: $N_r^*\in\{0,1,2,3\}$ for small molecule generation.
    }
    \label{tab:n_r_1}
      \resizebox{1\textwidth}{!}{
    \begin{tabular}{c|ccccccccccccccccc}
    \toprule
         \multirow{2}{*}{Method}  &  &\multicolumn{2}{c}{$N_r^*=0$}  & \multicolumn{2}{c}{$N_r^*=1$} & \multicolumn{2}{c}{$N_r^*=2$}  & \multicolumn{2}{c}{$N_r^*=3$} 
         \\
        & & MAE $\downarrow$ & TS $\downarrow$ 
        & MAE $\downarrow$ & TS $\downarrow$  & MAE $\downarrow$ & TS $\downarrow$  & MAE $\downarrow$ & TS $\downarrow$ \\
         \midrule
         No Guidance &
         & $3.03\pm1.26$ & \rule{0.4cm}{0.2mm}
         & $2.09\pm1.16$ & \rule{0.4cm}{0.2mm}
         & $1.27\pm1.02$ & \rule{0.4cm}{0.2mm}
         & $0.92\pm0.86$ & \rule{0.4cm}{0.2mm}
         \\ 
         \midrule
         DG &
         & $0.16\pm0.44$ & $0.16\pm0.03$
         & \underline{$0.11\pm0.33$} & $0.14\pm0.03$
         & \underline{$0.07\pm0.27$} & $0.13\pm0.02$
         & $0.06\pm0.25$ & $0.12\pm0.02$     
         \\
            TFG-Flow &
         & $0.30\pm0.74$ & $0.16\pm0.03$ 
         & $0.28\pm0.65$ & $0.13\pm0.02$
         & $0.20\pm0.51$ & $0.12\pm0.02$ 
         & $0.21\pm0.46$ & $0.12\pm0.02$
         \\
           SVDD &
         & $1.28\pm1.91$ & $0.14\pm0.05$
         & $0.35\pm1.14$ & $0.14\pm0.03$ 
         & $0.04\pm0.37$ & $0.14\pm0.02$ 
         & $\textbf{0.01}\pm\textbf{0.14}$ & $0.13\pm0.02$ 
         \\
        \midrule 
      \xtsampling &
     & $\textbf{0.03}\pm\textbf{0.41}$ & $0.20\pm0.04$ 
     & $\textbf{0.01}\pm\textbf{0.07}$ & $0.14\pm0.02$
     & $\textbf{0.02}\pm\textbf{0.13}$ & $0.13\pm0.02$
     & \underline{$0.02\pm0.18$} & $0.13\pm0.02$ 
       
     \\
       
       \xcleansampling &
      & \underline{$0.15\pm0.47$} & $0.17\pm0.04$ 
      & $0.11\pm0.37$ & $0.12\pm0.02$
      & $0.10\pm0.33$ & $0.12\pm0.02$ 
      & $0.15\pm0.40$ & $0.12\pm0.02$
      \\
     
     \xtgrad &
    & $1.44\pm1.95$ & $0.13\pm0.03$  
    & $0.44\pm1.19$ & $0.13\pm0.02$
    & $0.09\pm0.55$ & $0.12\pm0.02$
    & $0.04\pm0.30$ & $0.12\pm0.02$
    \\

         \bottomrule
        \end{tabular}}
\end{table*}

\begin{table*}[ht]
    \centering
    \caption{
    Supplementary results for Tab.~\ref{tab:smg}: $N_r^*\in\{4,5,6\}$ for small molecule generation.
    }
    \label{tab:n_r_1}
      \resizebox{0.8\textwidth}{!}{
    \begin{tabular}{c|ccccccccccccccccc}
    \toprule
         \multirow{2}{*}{Method}  &  &\multicolumn{2}{c}{$N_r^*=4$}  & \multicolumn{2}{c}{$N_r^*=5$} & \multicolumn{2}{c}{$N_r^*=6$} 
         \\
        & & MAE $\downarrow$ & TS $\downarrow$ 
        & MAE $\downarrow$ & TS $\downarrow$  & MAE $\downarrow$ & TS $\downarrow$  \\
         \midrule
         No Guidance &
         & $1.27\pm0.95$ & \rule{0.4cm}{0.2mm}
         & $2.03\pm1.16$ & \rule{0.4cm}{0.2mm}
         & $2.98\pm1.23$ & \rule{0.4cm}{0.2mm}
         \\ 
         \midrule
         DG &
         & $0.08\pm0.27$ & $0.12\pm0.02$ 
         & \underline{$0.19\pm0.40$} & $0.13\pm0.02$ 
         & $\textbf{0.42}\pm\textbf{0.51}$ & $0.14\pm0.02$    
         \\
     
            TFG-Flow &
         & $0.30\pm0.53$ & $0.12\pm0.02$
         & $0.51\pm0.64$ & $0.12\pm0.02$ 
         & $0.94\pm0.86$ & $0.13\pm0.02$  
         \\
        SVDD &
         & \underline{$0.06\pm0.36$} & $0.13\pm0.02$
         & $0.28\pm0.97$ & $0.13\pm0.02$ 
         & $1.42\pm1.97$ & $0.12\pm0.02$  
         \\
        \midrule 
      \xtsampling &
     & $\textbf{0.05}\pm\textbf{0.29}$ & $0.13\pm0.02$ 
     & $\textbf{0.13}\pm\textbf{0.53}$ & $0.12\pm0.02$  
     & \underline{$0.59\pm1.47$} & $0.12\pm0.03$ 
       
     \\
       
       \xcleansampling &
      & $0.30\pm0.58$ & $0.11\pm0.02$
      & $0.67\pm0.84$ & $0.11\pm0.02$  
      & $1.41\pm1.17$ & $0.11\pm0.02$  
      \\
     
     \xtgrad &
     & $0.08\pm0.47$ & $0.12\pm0.02$ 
    & $0.26\pm0.90$ & $0.12\pm0.02$ 
    & $1.24\pm1.93$ & $0.12\pm0.02$ 
    \\
   
         \bottomrule
        \end{tabular}}
\end{table*}

\subsubsection{Scalability}

\begin{figure}[H]
    \centering
    \subfigure{\includegraphics[width=0.245\linewidth]{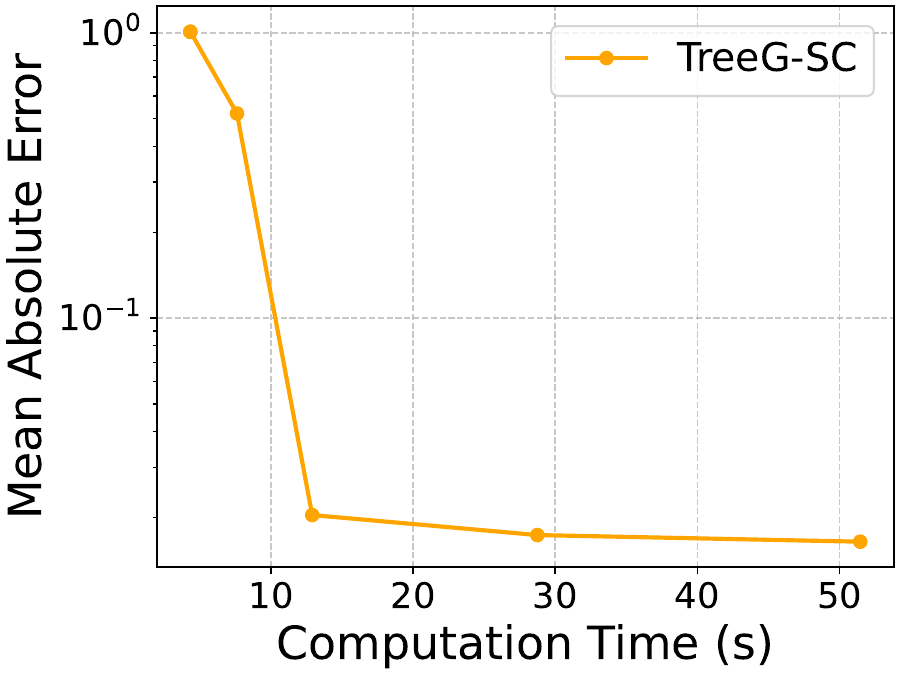}}
    \subfigure{\includegraphics[width=0.245\linewidth]{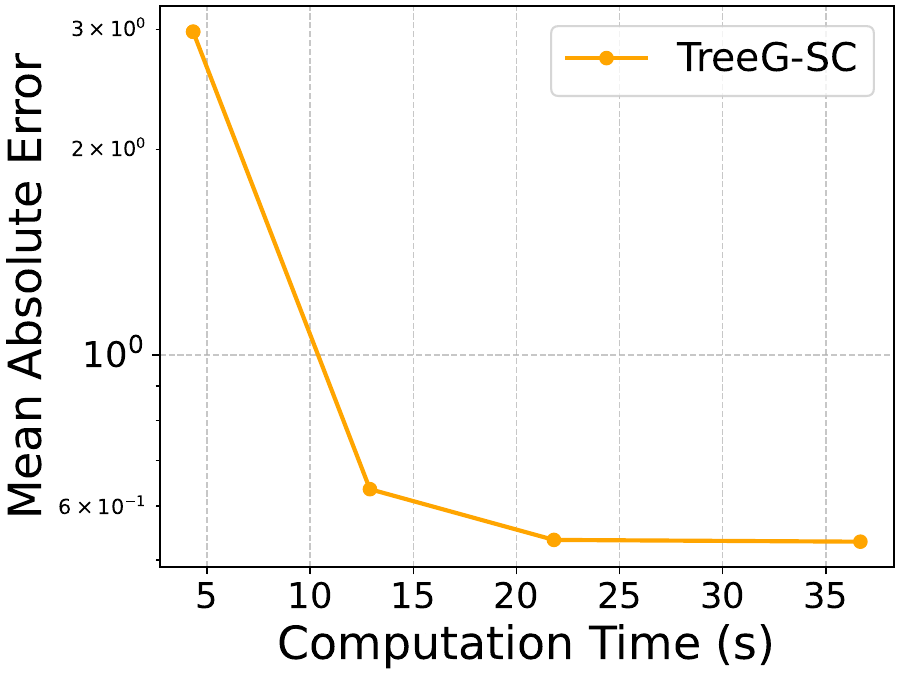}}
    \subfigure{\includegraphics[width=0.245\linewidth]{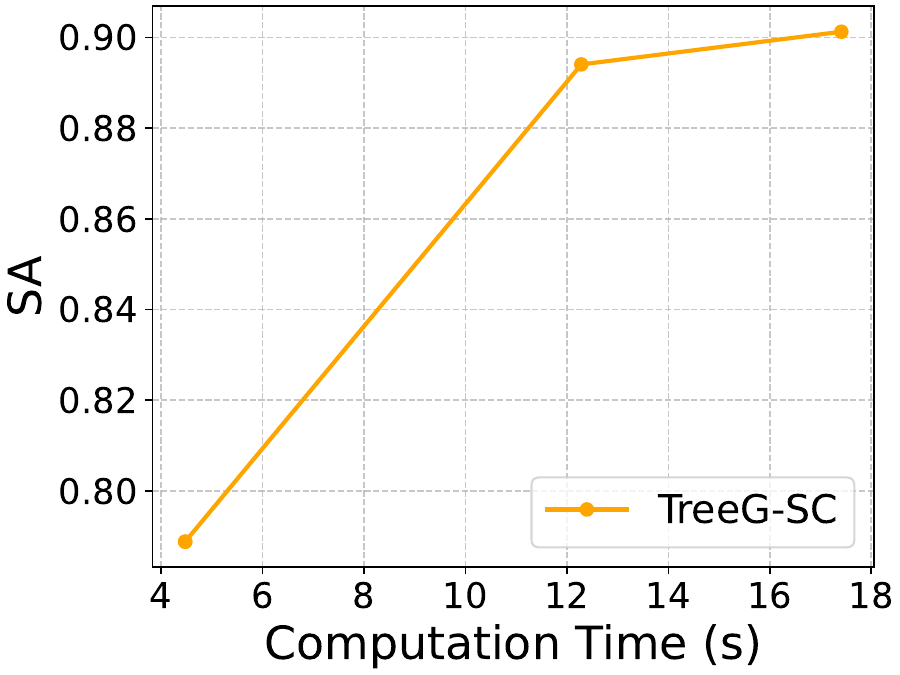}}
    \caption{Small Molecule Generation: Scaling Law of \xtsampling. From left to right, the targets are $N_r^*=3$,  $N_r^*=6$, and SA.
    }
    \label{fig:app smg scaling law xt sampling}
\end{figure}

\begin{figure}[H]
    \centering
    \subfigure{\includegraphics[width=0.245\linewidth]{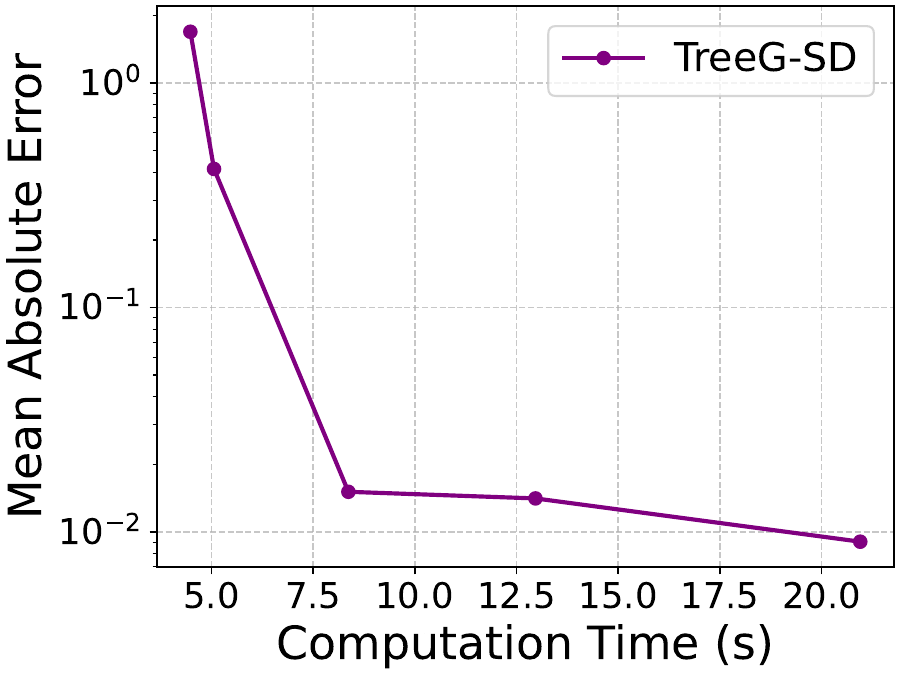}}
    \subfigure{\includegraphics[width=0.245\linewidth]{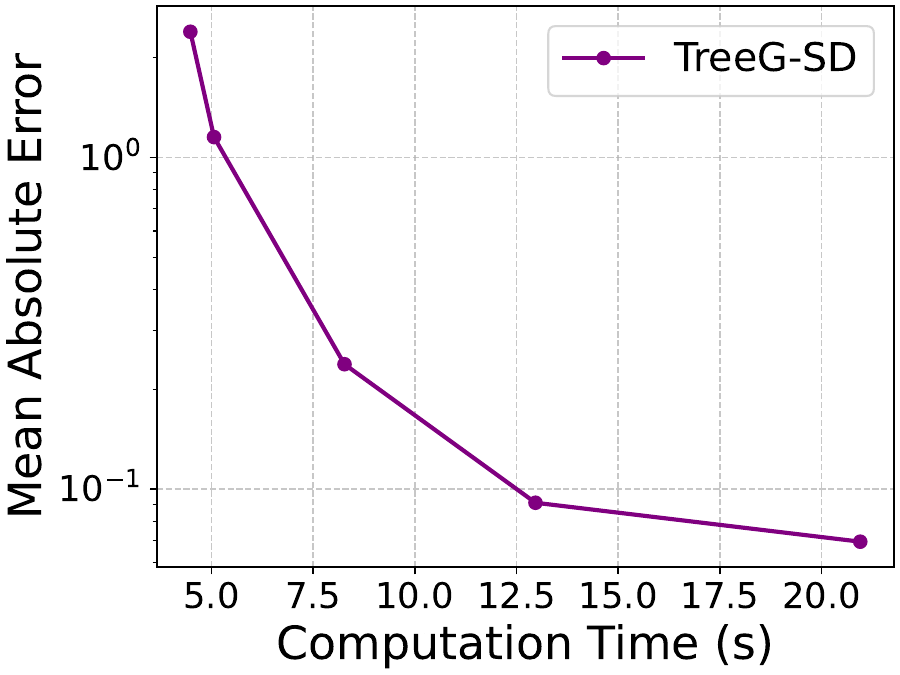}}
    \subfigure{\includegraphics[width=0.245\linewidth]{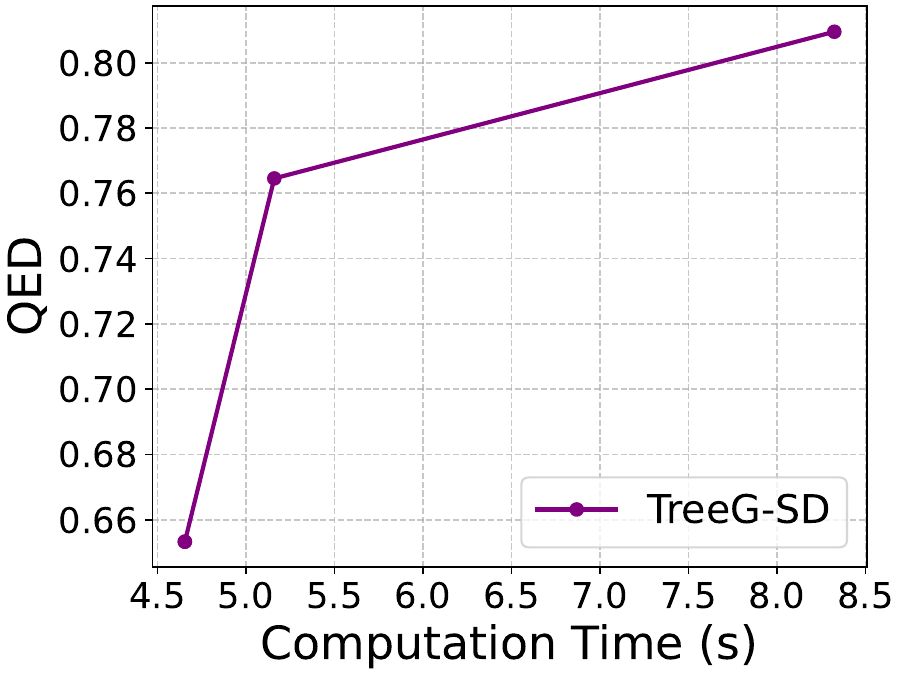}}
    \caption{Small Molecule Generation: Scaling Law of \xcleansampling. From left to right, the targets are $N_r^*=1$,  $N_r^*=5$, and QED. 
    }
    \label{fig:app smg scaling law x1 sampling}
\end{figure}

We demonstrate the scaling laws for \xtsampling and \xcleansampling in Figures \ref{fig:app smg scaling law xt sampling} and \ref{fig:app smg scaling law x1 sampling}. Increasing computation time could boost guidance performance, i.e., lower mean absolute errors or higher property values. 

\vspace{-10pt}

\subsection{Additional Results for Enhancer DNA Design}
\subsubsection{Full Results of \cref{tab:enhancer guidance res}}
Additional results of guidance methods for Class 4-8 are shown in \cref{tab:dna-results-part1} and \cref{tab:dna-results-part2}.
For both DG and our \xtgrad, we experiment with guidance values $\gamma \in [1, 2, 5, 10, 20, 50, 100, 200]$ and compare the highest average conditional probability across the eight classes. On average, \xtgrad outperforms DG by $18.43\%$. 

\begin{table*}[ht]
    \centering
    \caption{Supplementary results (Part 1): Classes 4–6 for enhancer DNA design.}
    \label{tab:dna-results-part1}
    \resizebox{\textwidth}{!}{
    \begin{tabular}{lc|ccccccccc}
        \toprule
        \multicolumn{2}{c|}{\multirow{2}{*}{Method {\small (strength $\gamma$)}}} 
        & \multicolumn{3}{c}{Class 4} 
        & \multicolumn{3}{c}{Class 5} 
        & \multicolumn{3}{c}{Class 6} \\
        & & Prob $\uparrow$ & FBD $\downarrow$ & Div $\uparrow$
          & Prob $\uparrow$ & FBD $\downarrow$ & Div $\uparrow$
          & Prob $\uparrow$ & FBD $\downarrow$ & Div $\uparrow$ \\
        \midrule
        No Guidance & \rule{0.4cm}{0.2mm} & $0.007 \pm 0.059$ & $446$ & $373$ & $0.035 \pm 0.112$ & $141$ & $373$ & $0.037 \pm 0.115$ & $179$ & $373$ \\
        \midrule
        \multirow{3}{*}{DG} 
            & {\footnotesize $20$} & $0.669 \pm 0.377$ & $57$ & $373$ & $0.665 \pm 0.332$ & $32$ & $374$ & $0.595 \pm 0.334$ & $26$ & $373$ \\
            & {\footnotesize $100$} & $0.585 \pm 0.380$ & $132$ & $365$ & $0.466 \pm 0.376$ & $86$ & $373$ & $0.385 \pm 0.334$ & $84$ & $364$ \\
            & {\footnotesize $200$} & $0.404 \pm 0.384$ & $140$ & $369$ & $0.199 \pm 0.284$ & $148$ & $373$ & $0.164 \pm 0.235$ & $132$ & $366$ \\
        \midrule
        TFG-Flow & {\footnotesize $200$} & $0.015 \pm 0.083$ & $408$ & $375$ & $0.008 \pm 0.048$ & $267$ & $375$ & $0.033 \pm 0.104$ & $271$ & $375$ \\
        \midrule
        SVDD & \rule{0.4cm}{0.2mm} & $0.107 \pm 0.264$ & $237$ & $374$ & $0.142 \pm 0.240$ & $58$ & $374$ & $0.124 \pm 0.230$ & $70$ & $373$ \\
        \midrule
        \multirow{3}{*}{\textbf{\xtgrad}} 
            & {\footnotesize $20$} & $0.518 \pm 0.391$ & $115$ & $365$ & $0.290 \pm 0.306$ & $61$ & $372$ & $0.250 \pm 0.292$ & $67$ & $366$ \\
            & {\footnotesize $100$} & $0.845 \pm 0.264$ & $199$ & $365$ & $0.778 \pm 0.279$ & $123$ & $359$ & $0.843 \pm 0.232$ & $120$ & $368$ \\
            & {\footnotesize $200$} & $0.826 \pm 0.297$ & $236$ & $370$ & $0.543 \pm 0.426$ & $104$ & $374$ & $0.364 \pm 0.432$ & $98$ & $374$ \\
        \bottomrule
    \end{tabular}
    }
\end{table*}
\begin{table*}[ht]
    \centering
    \caption{Supplementary results (Part 2): Classes 7–8 for enhancer DNA design.}
    \label{tab:dna-results-part2}
    \resizebox{0.8\textwidth}{!}{
    \begin{tabular}{lc|cccccc}
        \toprule
        \multicolumn{2}{c|}{\multirow{2}{*}{Method {\small (strength $\gamma$)}}} 
        & \multicolumn{3}{c}{Class 7} 
        & \multicolumn{3}{c}{Class 8} \\
        & & Prob $\uparrow$ & FBD $\downarrow$ & Div $\uparrow$
          & Prob $\uparrow$ & FBD $\downarrow$ & Div $\uparrow$ \\
        \midrule
        No Guidance & \rule{0.4cm}{0.2mm} & $0.010 \pm 0.065$ & $292$ & $373$ & $0.013 \pm 0.073$ & $478$ & $373$ \\
        \midrule
        \multirow{3}{*}{DG} 
            & {\footnotesize $20$} & $0.693 \pm 0.346$ & $61$ & $366$ & $0.609 \pm 0.350$ & $95$ & $372$ \\
            & {\footnotesize $100$} & $0.475 \pm 0.398$ & $149$ & $364$ & $0.600 \pm 0.342$ & $326$ & $372$ \\
            & {\footnotesize $200$} & $0.208 \pm 0.313$ & $266$ & $360$ & $0.453 \pm 0.370$ & $362$ & $371$ \\
        \midrule
        TFG-Flow & {\footnotesize $200$} & $0.001 \pm 0.015$ & $371$ & $375$ & $0.006 \pm 0.055$ & $662$ & $375$ \\
        \midrule
        SVDD & \rule{0.4cm}{0.2mm} & $0.039 \pm 0.136$ & $206$ & $369$ & $0.038 \pm 0.122$ & $250$ & $373$ \\
        \midrule
        \multirow{3}{*}{\textbf{\xtgrad}} 
            & {\footnotesize $20$} & $0.536 \pm 0.335$ & $137$ & $307$ & $0.159 \pm 0.233$ & $332$ & $366$ \\
            & {\footnotesize $100$} & $0.740 \pm 0.365$ & $161$ & $343$ & $0.413 \pm 0.412$ & $307$ & $373$ \\
            & {\footnotesize $200$} & $0.423 \pm 0.457$ & $177$ & $367$ & $0.073 \pm 0.205$ & $346$ & $373$ \\
        \bottomrule
    \end{tabular}
    }
\end{table*}

\subsubsection{Scalability}\label{app:scalability of DNA}
\begin{figure}[H]
    \centering
    \subfigure{\includegraphics[width=0.49\linewidth]{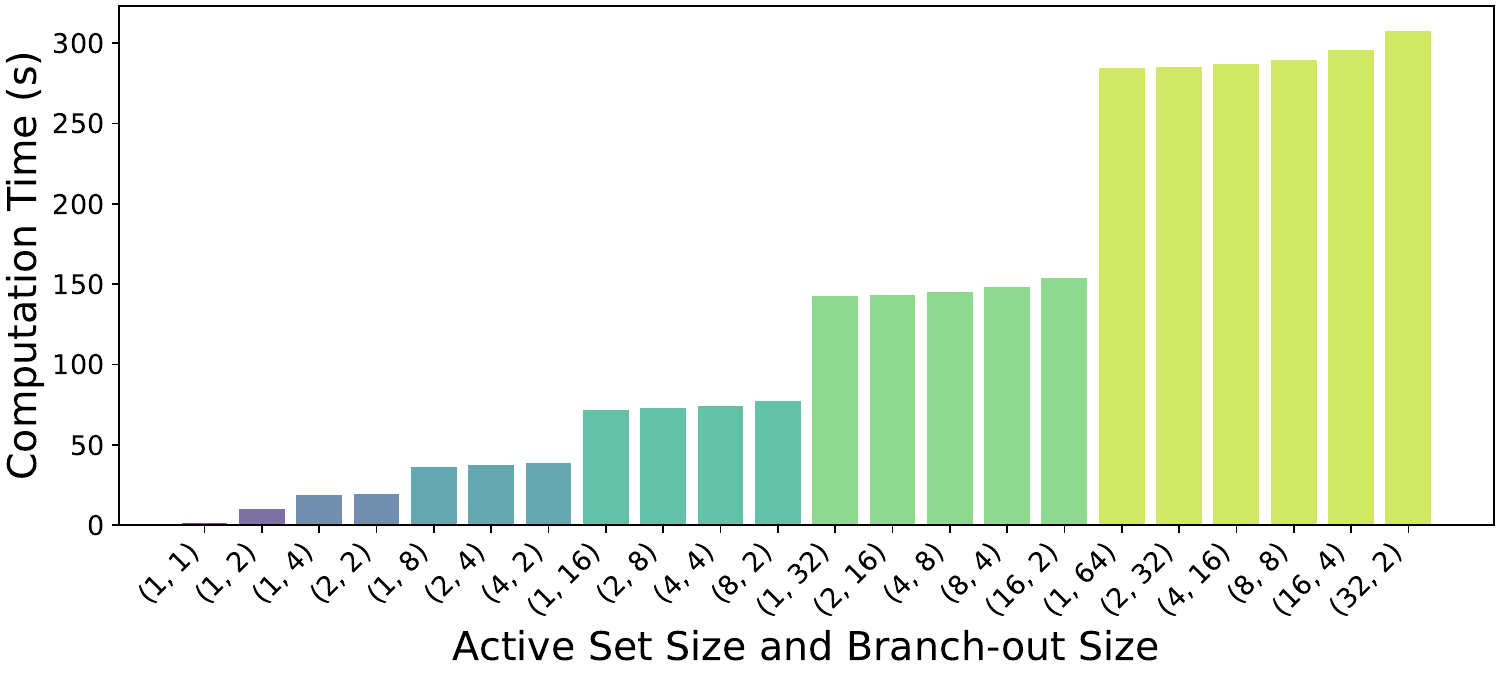}}
\subfigure{\includegraphics[width=0.49\linewidth]{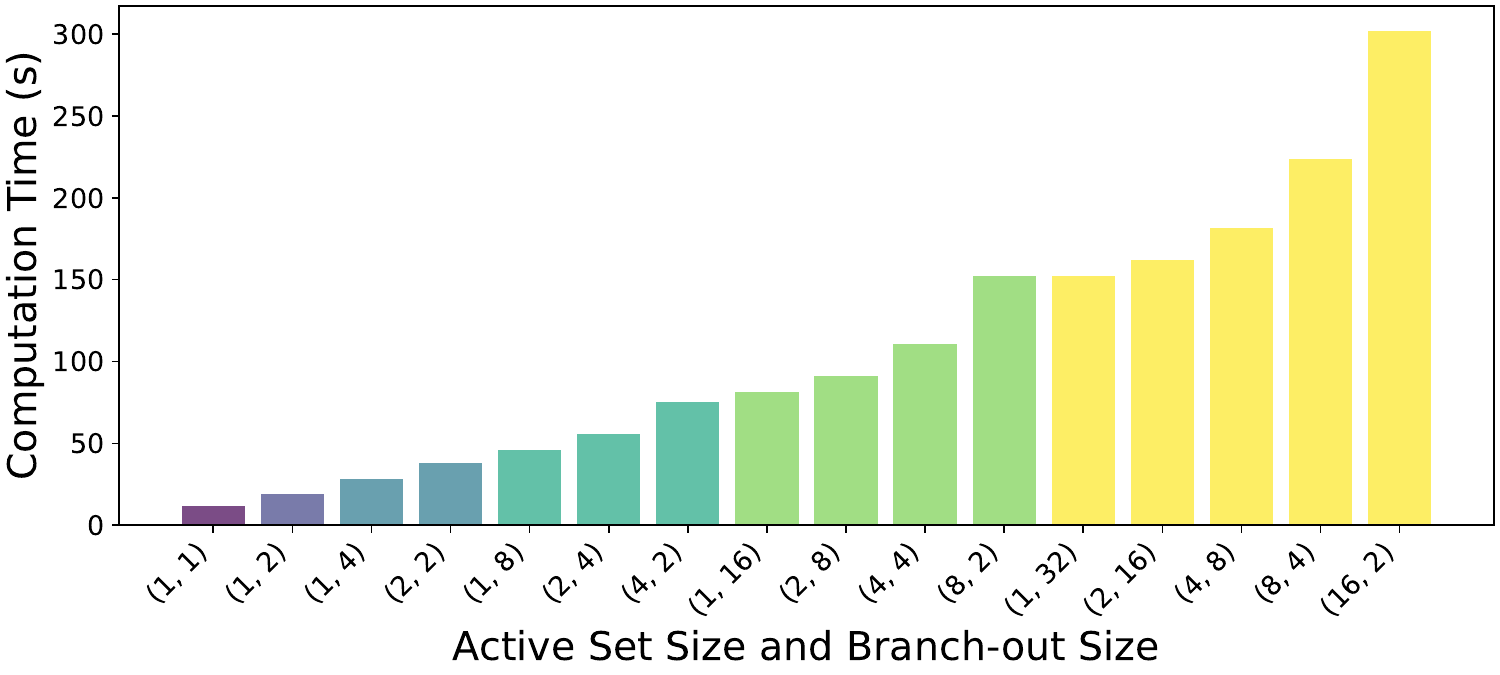}}
    \caption{Inference Time for $(A,K)$ Combinations (Left: \xtsampling; Right: \xtgrad) in DNA Enhancer Design. Combinations with the same product $A \times K$ show similar inference times.}

    
    \label{fig:time with fixed AK}
\end{figure}

\textbf{$A*K$ as a Computation Reference.} We use $A*K$ as the reference metric for inference time computation in \xtsampling and \xtgrad, both employing \texttt{BranchOut}-Current. The corresponding inference times are shown in \cref{fig:time with fixed AK}, measured for a batch size of 100. Combinations of $(A,K)$ that yield the same $A*K$ value exhibit similar inference times. We exclude the case where $K=1$, as it does not require evaluation and selection, leading to a shorter inference time in practical implementation.


\begin{figure}[H]
    \centering    \subfigure{\includegraphics[width=0.245\linewidth]{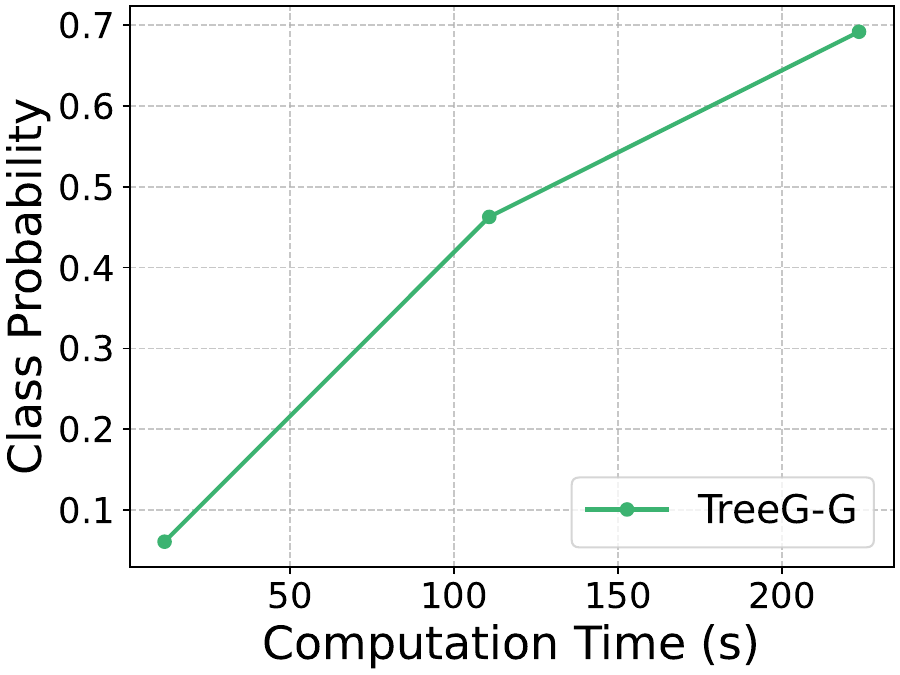}}
\subfigure{\includegraphics[width=0.245\linewidth]{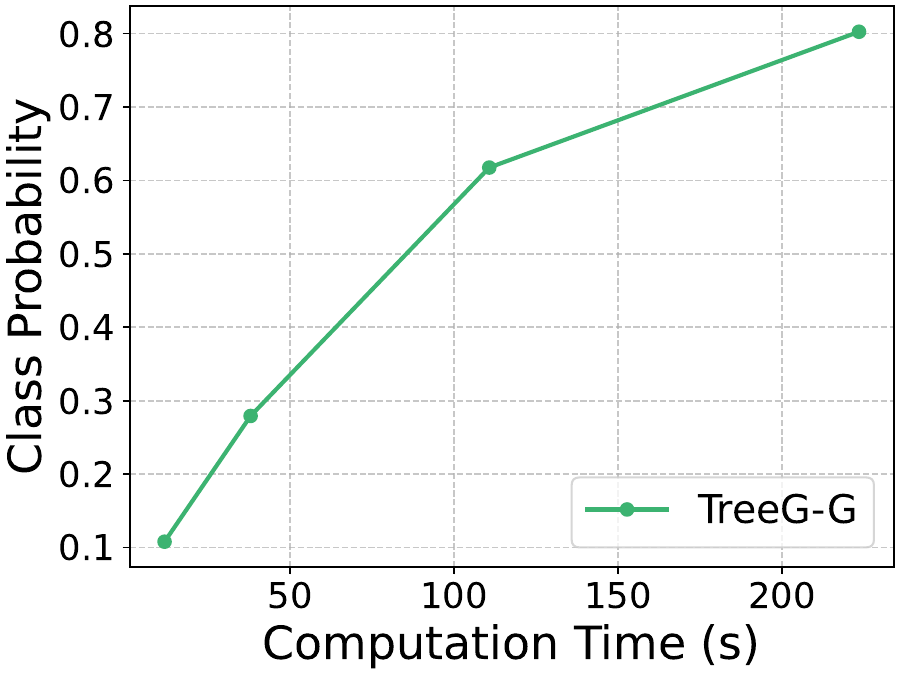}}     \subfigure{\includegraphics[width=0.245\linewidth]{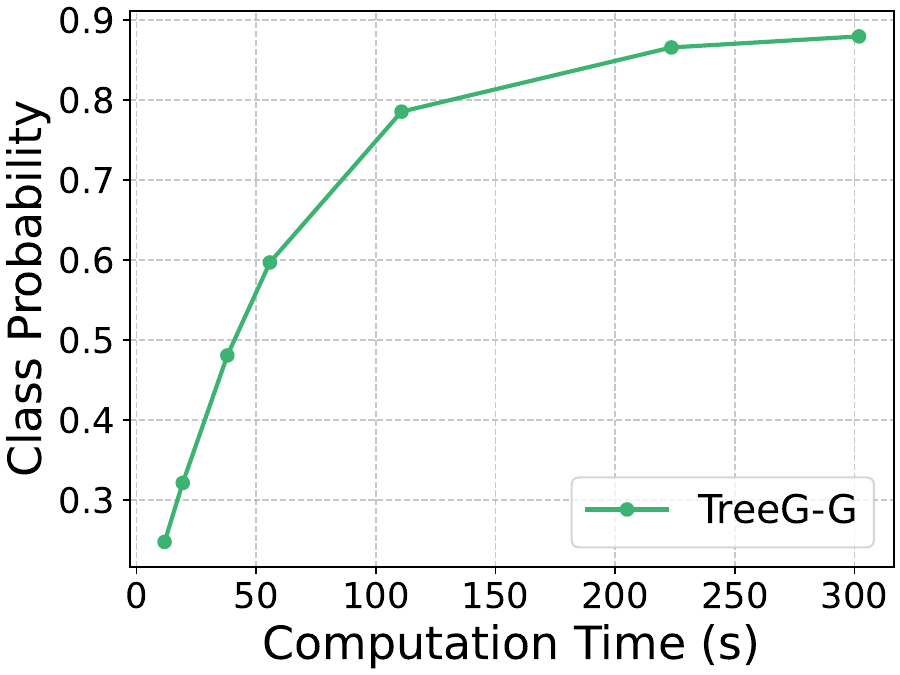}}
\subfigure{\includegraphics[width=0.245\linewidth]{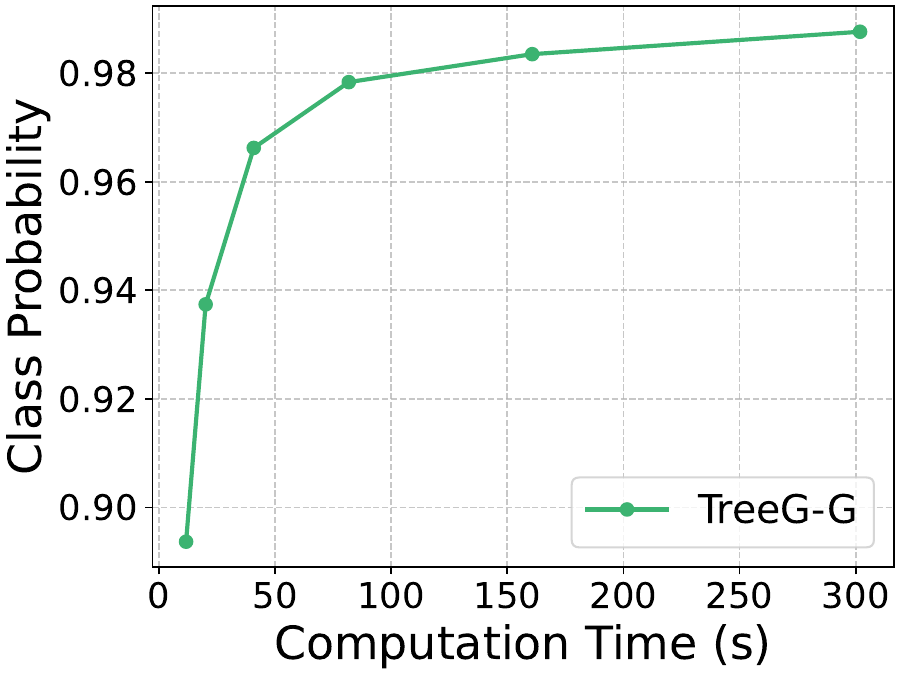}}
\subfigure{\includegraphics[width=0.245\linewidth]{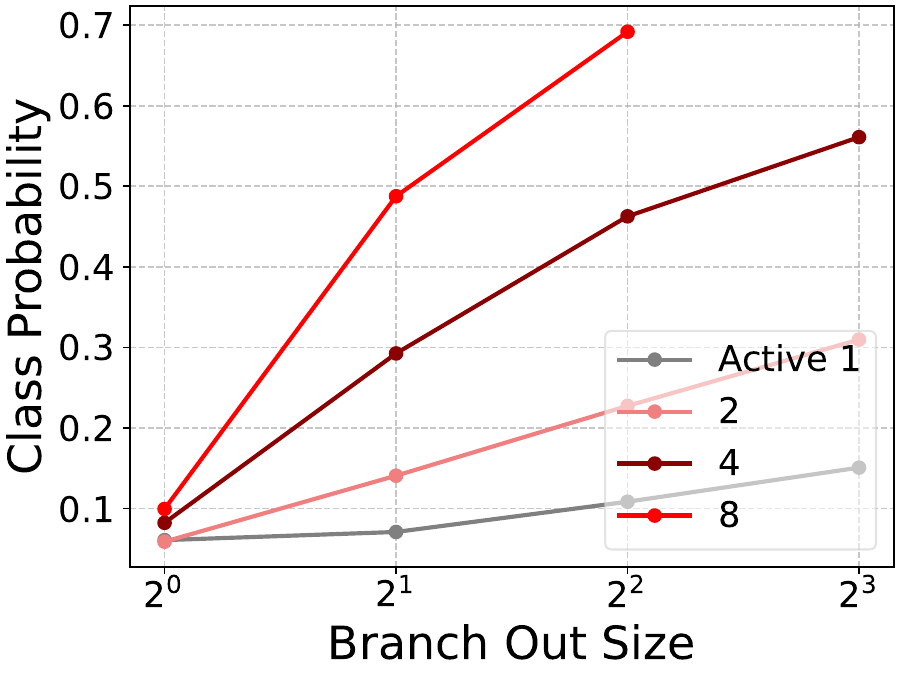}}
\subfigure{\includegraphics[width=0.245\linewidth]{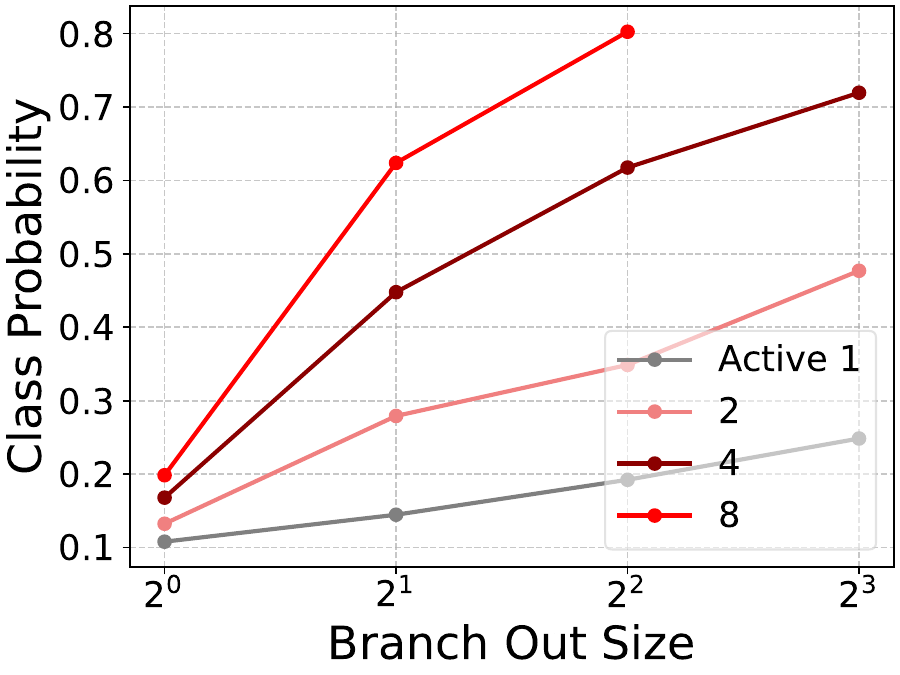}}     \subfigure{\includegraphics[width=0.245\linewidth]{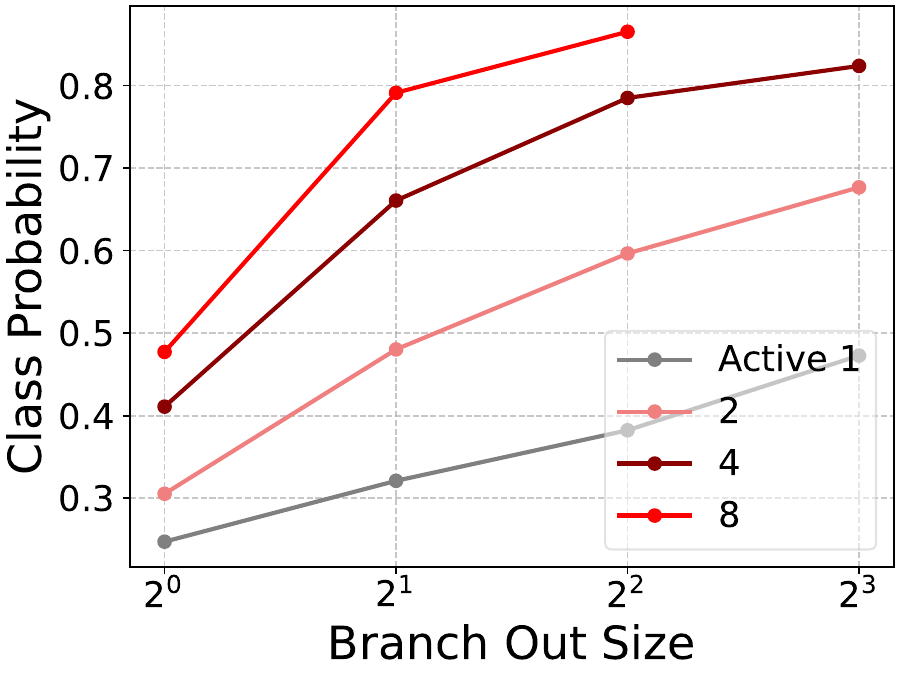}}
\subfigure{\includegraphics[width=0.245\linewidth]{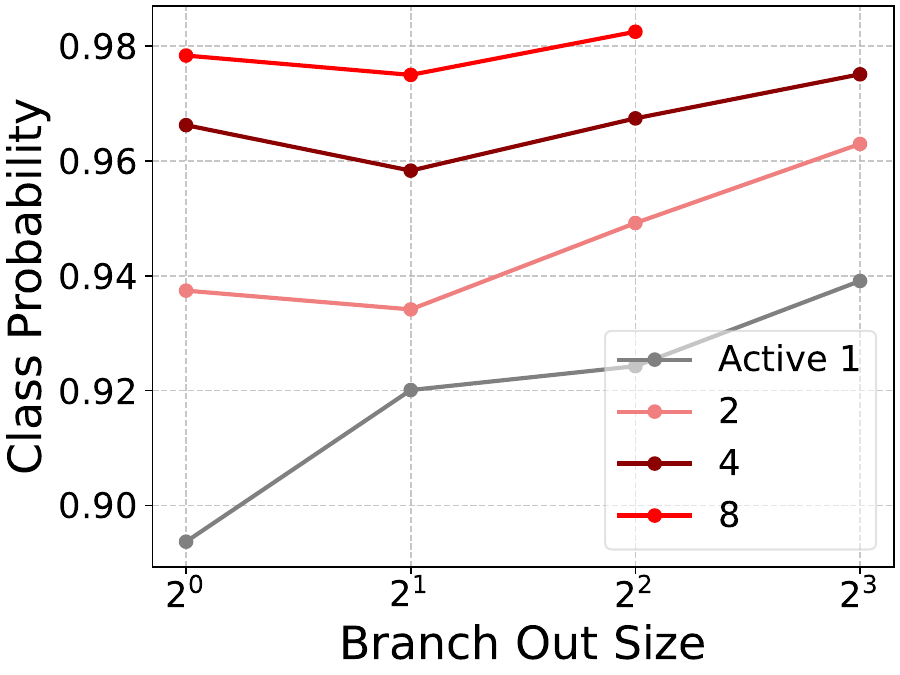}}
\subfigure{\includegraphics[width=0.245\linewidth]{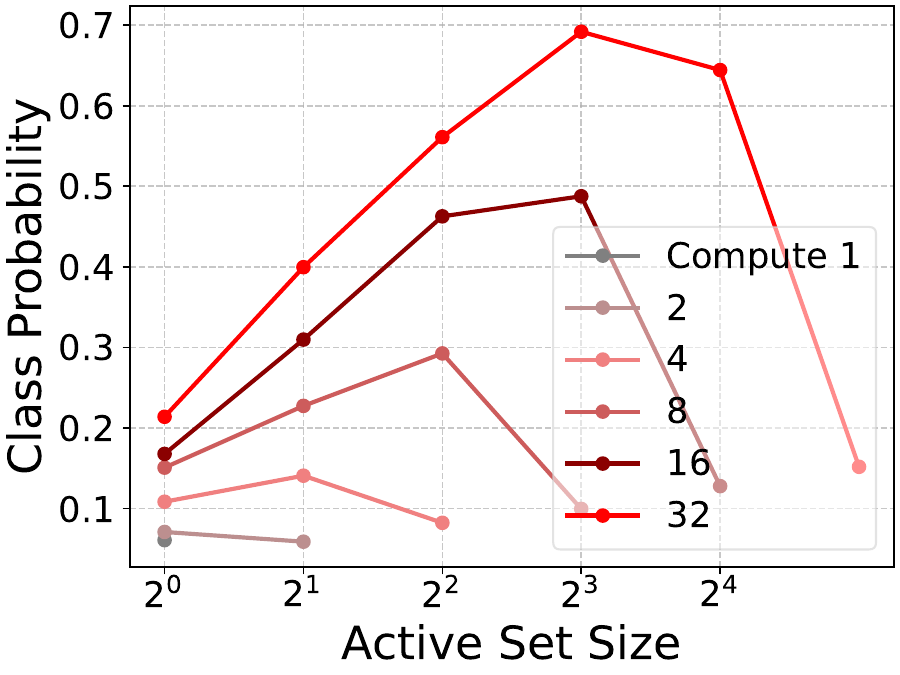}}
\subfigure{\includegraphics[width=0.245\linewidth]{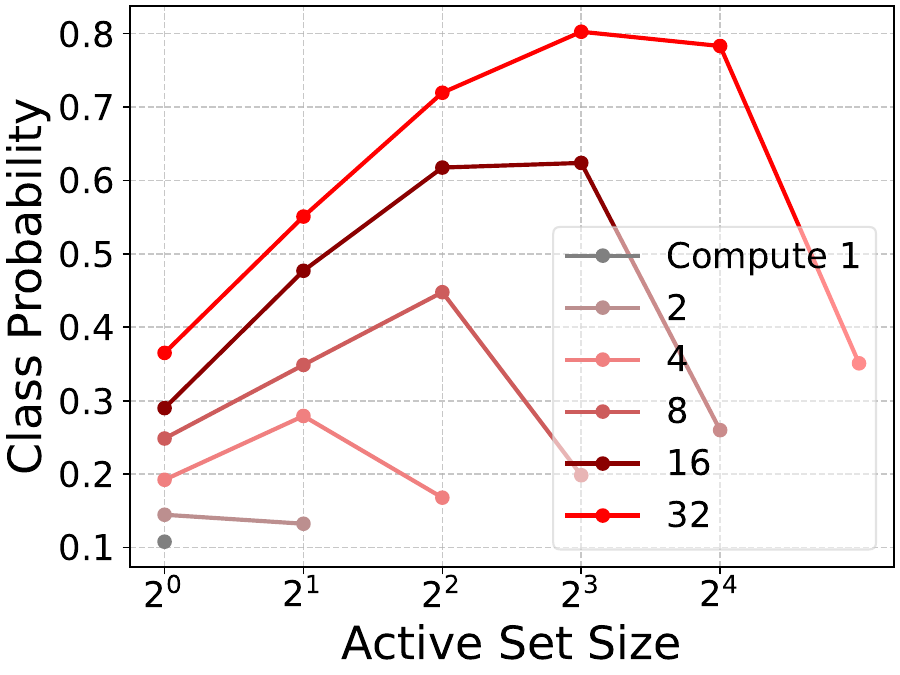}}
\subfigure{\includegraphics[width=0.245\linewidth]{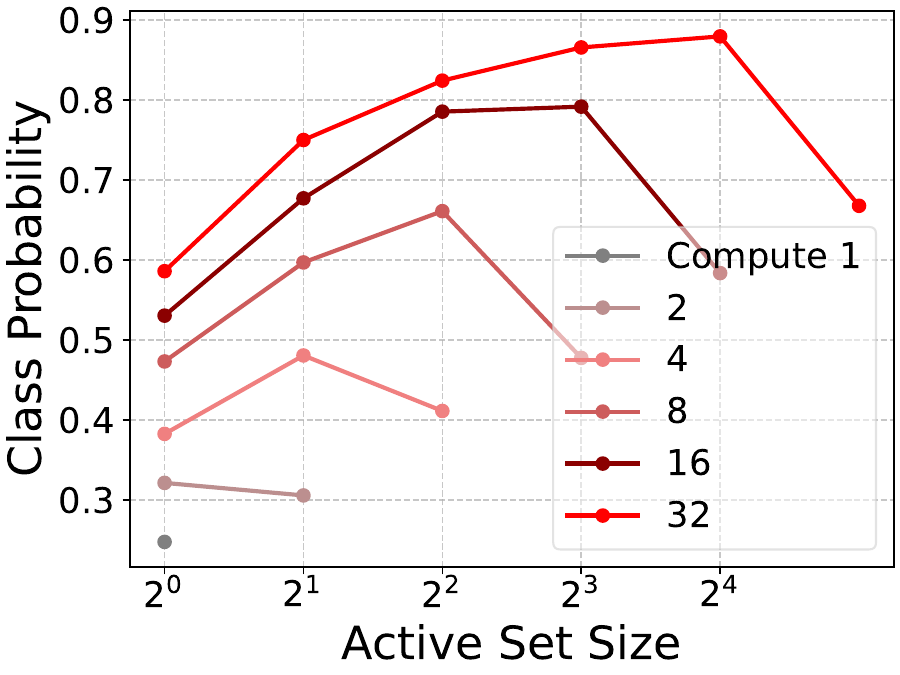}}
\subfigure{\includegraphics[width=0.245\linewidth]{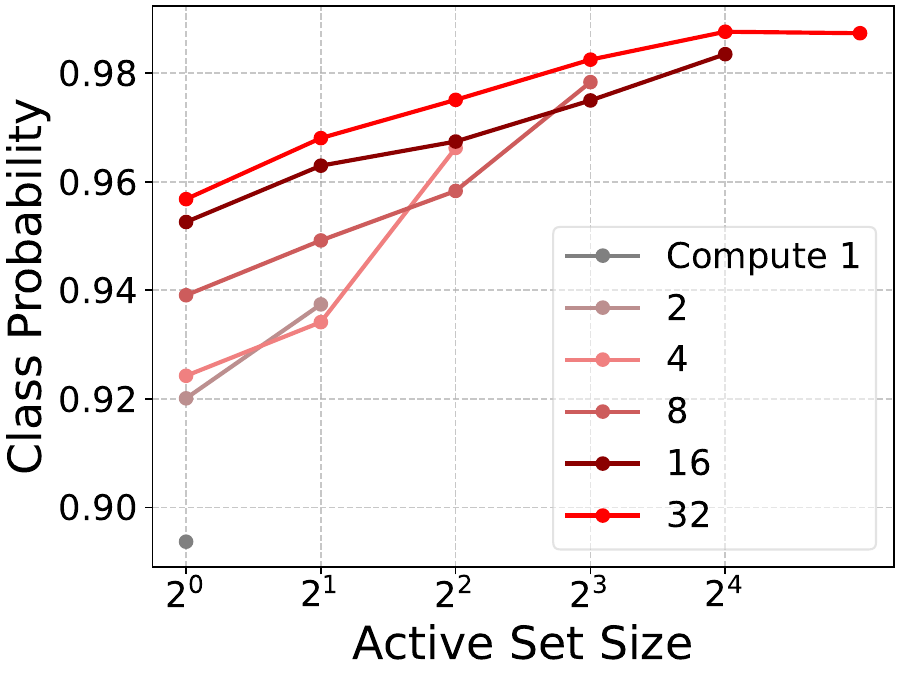}}
    \caption{\xtgrad for Enhancer DNA Design. Top row: Effect of scaling on inference time. Middle row: Impact of increasing the active set size $A$ or branch-out size $K$. Bottom row: Trade-off between $A$ and $K$ with fixed compute $A*K$. Columns from left to right correspond to different guidance strengths: $\gamma = 5,\ 10,\ 20,\ 200$ (results shown for Class 3).}
    \label{fig:app dna scaling law}
\end{figure}

\begin{figure}[H]
    \centering
\subfigure{\includegraphics[width=0.245\linewidth]{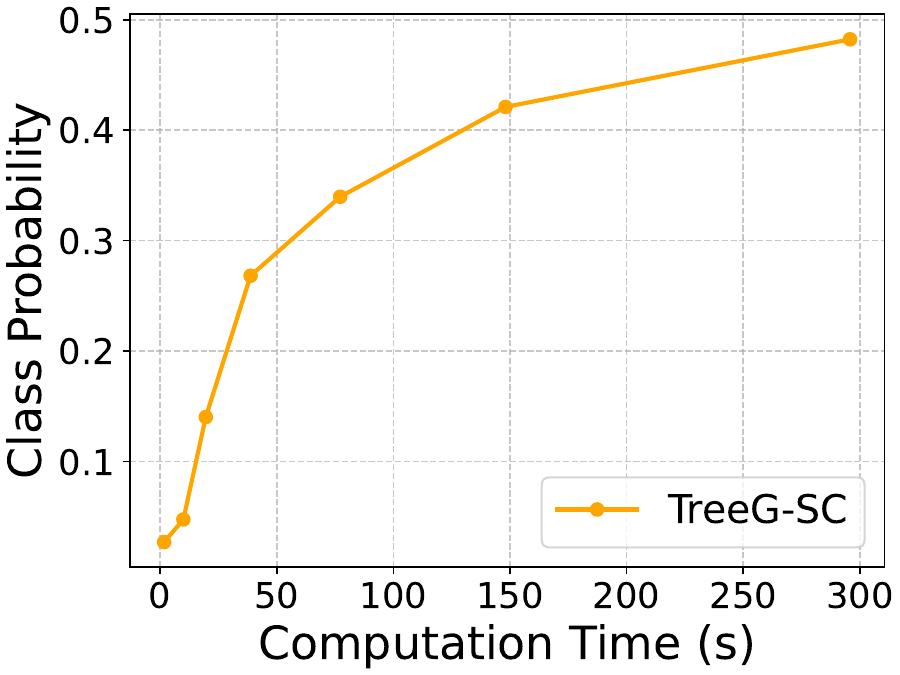}}
\subfigure{\includegraphics[width=0.245\linewidth]{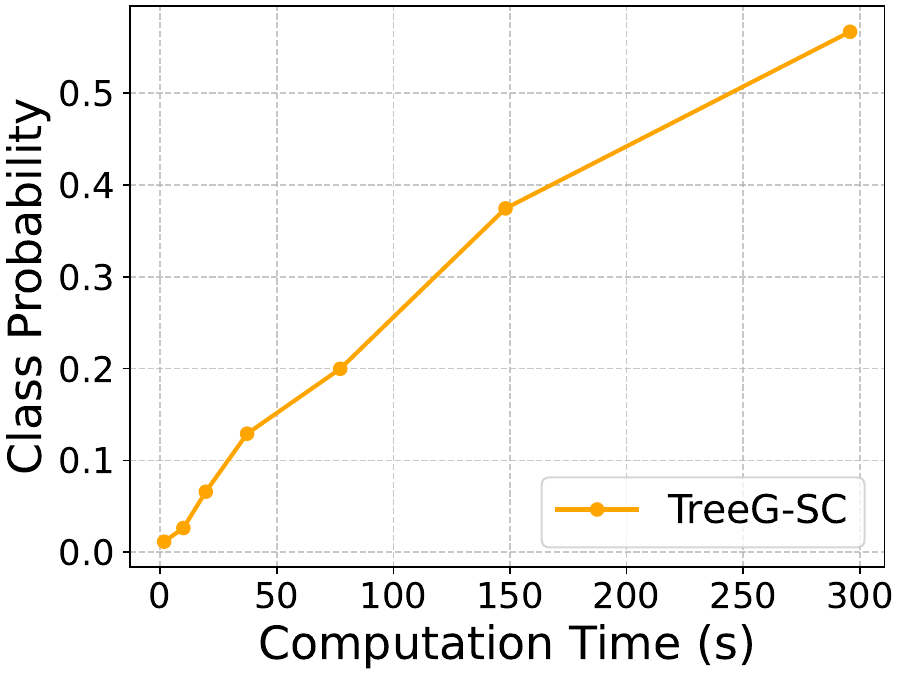}}
\subfigure{\includegraphics[width=0.245\linewidth]{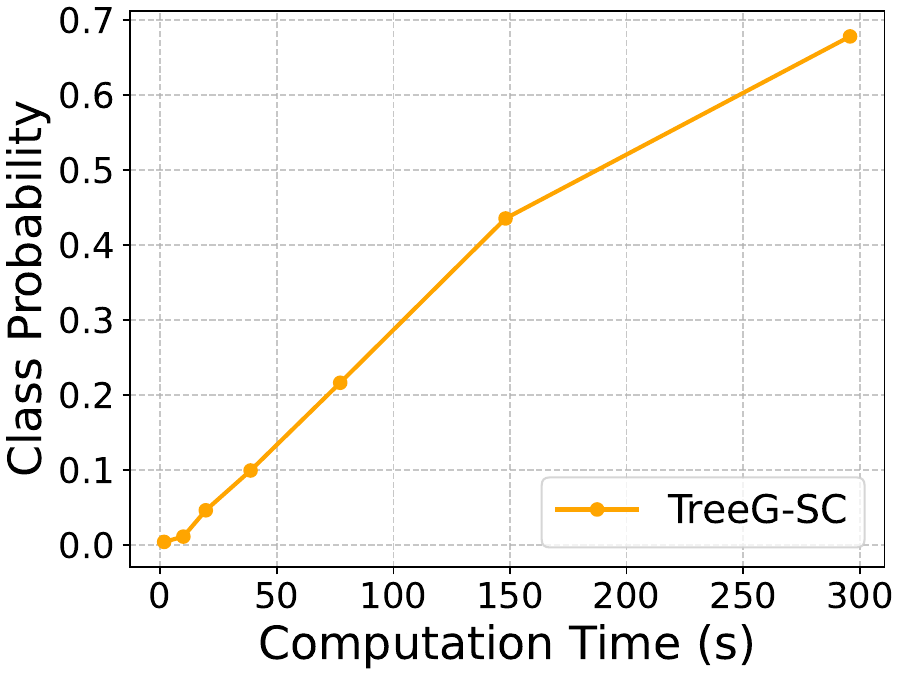}}
\subfigure{\includegraphics[width=0.245\linewidth]{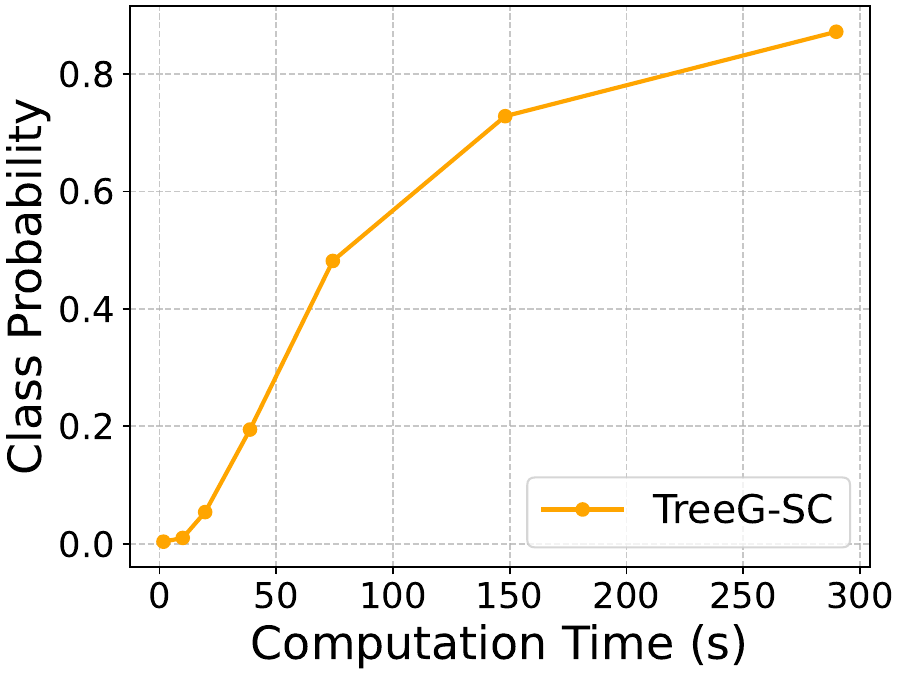}}
\subfigure{\includegraphics[width=0.245\linewidth]{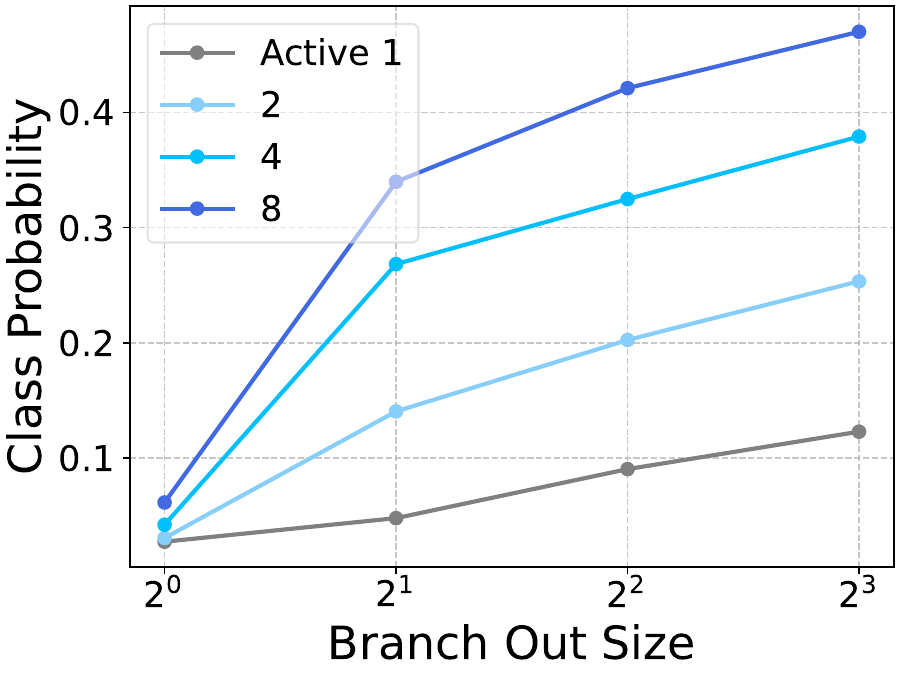}}
\subfigure{\includegraphics[width=0.245\linewidth]{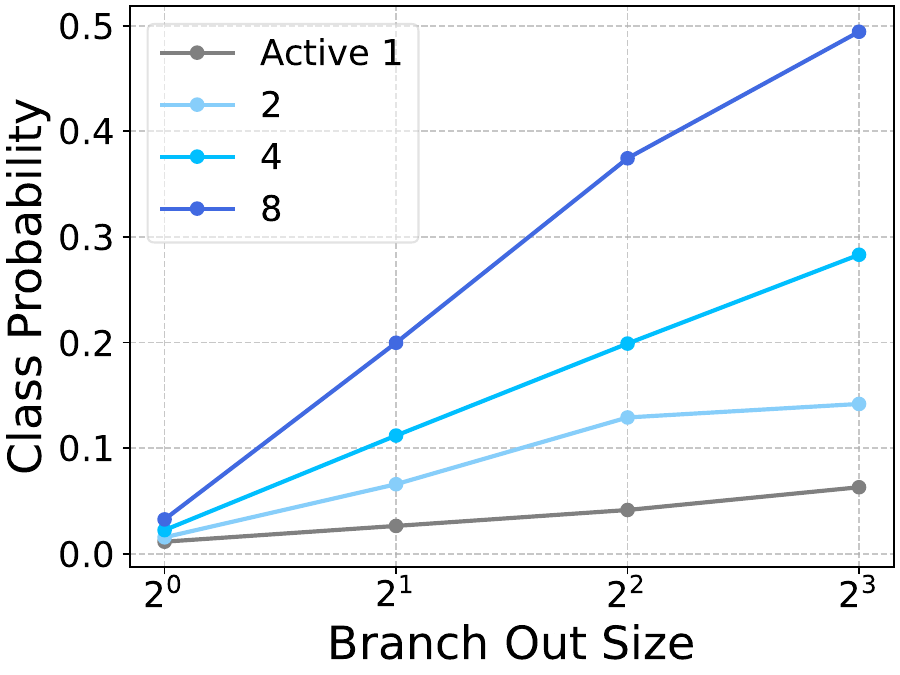}}
\subfigure{\includegraphics[width=0.245\linewidth]{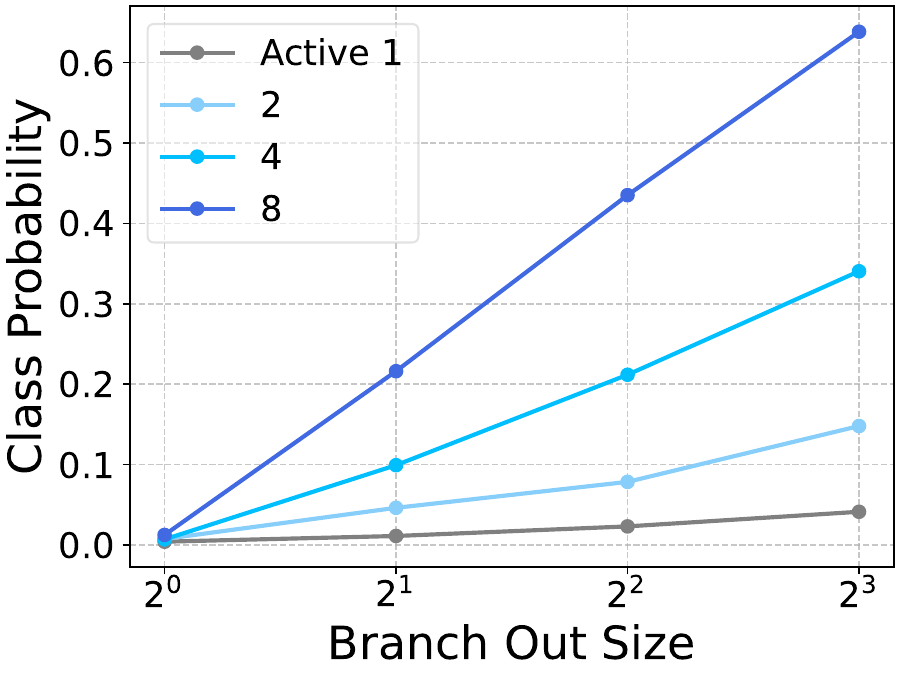}}
\subfigure{\includegraphics[width=0.245\linewidth]{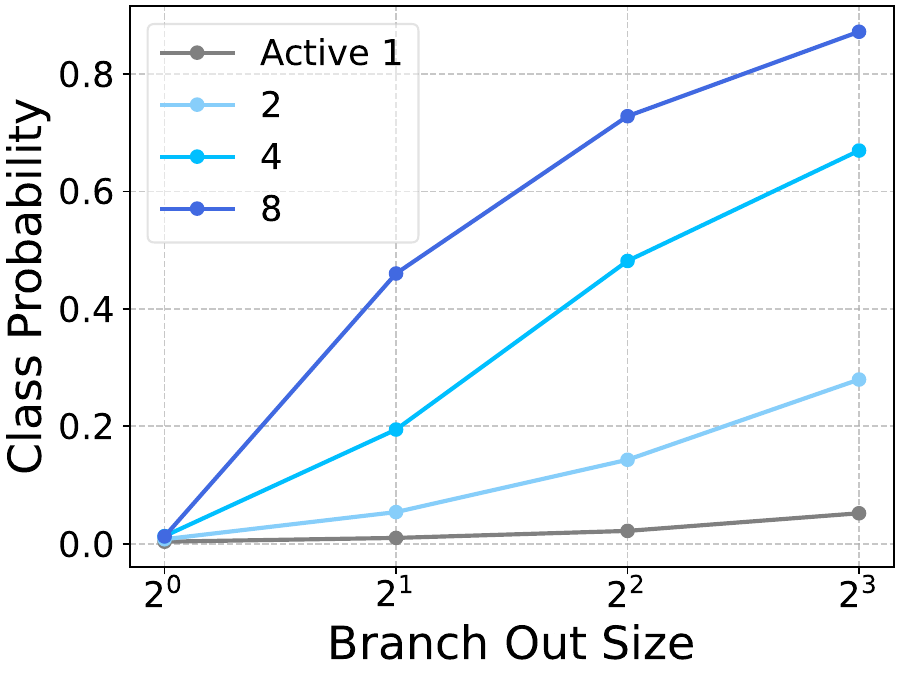}}
\subfigure{\includegraphics[width=0.245\linewidth]{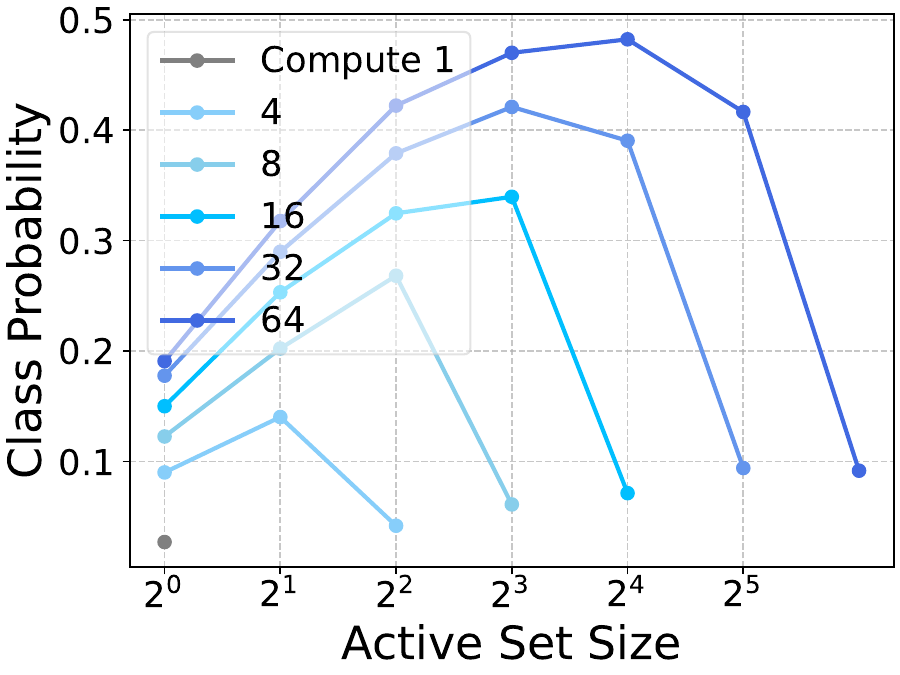}}
    \subfigure{\includegraphics[width=0.245\linewidth]{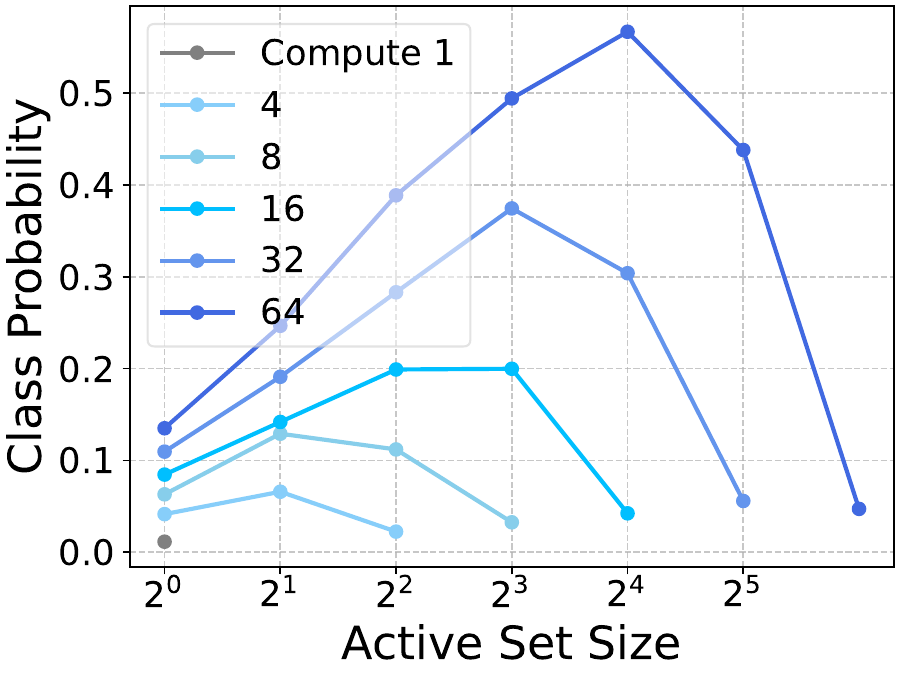}}
     \subfigure{\includegraphics[width=0.245\linewidth]{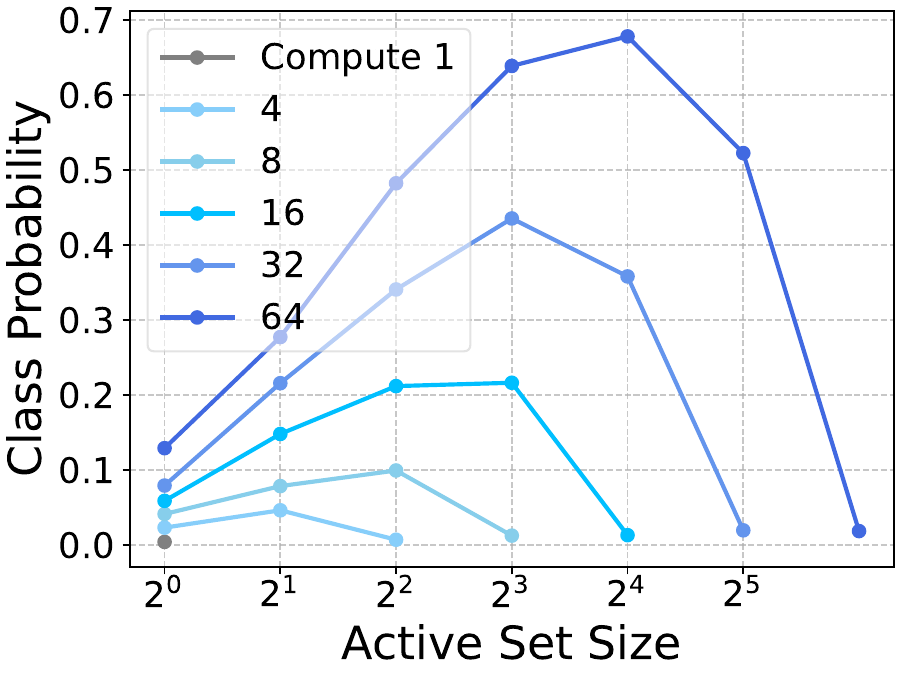}}
    \subfigure{\includegraphics[width=0.245\linewidth]{icml2025/figures/enhancer/xt_sampling_mcts/fixed_total/class_4.pdf}}
    \caption{\xtsampling for Enhancer DNA Design. Top row: Effect of scaling on inference time. Middle row: Impact of increasing the active set size $A$ or branch-out size $K$. Bottom row: Trade-off between $A$ and $K$ with fixed compute $A*K$. Columns from left to right correspond to Class 1 to 4.}
    \label{fig:app dna scaling law xt sampling}
\end{figure}

We provide the scaling law of \xtgrad at different guidance strengths in \cref{fig:app dna scaling law}. 
We also provide the scaling law for \xtsampling in \cref{fig:app dna scaling law xt sampling}.


\section{Ablation Studies}
\subsection{Symbolic Music Generation}

For \xcleansampling, we ablation on $N_{\text{iter}}$ (detailed in \cref{alg:music detailed}). As shown in \cref{tab:ablation N iter}, the loss decreases when $N_{\text{iter}}$ increases. However, increasing $N_{\text{iter}}$ also leads to higher computation costs. Here's a trade-off between controllability and computation cost.

\vspace{-10pt}
\begin{table}[ht]
    \centering
          \caption{Ablation on $N_{\text{iter}}$ of \xcleansampling.}
    \label{tab:ablation N iter}
    \resizebox{0.65\textwidth}{!}{
    \begin{tabular}{ccccc}
    \toprule
      $N_{\text{iter}}$& Loss $\downarrow$ (PH) & OA $\uparrow$ (PH)  & Loss $\downarrow$ (ND) & OA $\uparrow$ (ND) \\
    \midrule
       1  &  $0.0031 \pm 0.0037$ & $0.733 \pm  0.024$ & $0.207 \pm 0.418$ & $0.810 \pm 0.049$  \\
       2 & $0.0005 \pm 0.0008$ & $0.833 \pm 0.018$ & $0.217 \pm 0.450$ & $0.819 \pm 0.050$ \\
       4 &  $0.0005 \pm 0.0006$ & $0.786 \pm 0.015$ & $0.139 \pm 0.319$ & $0.830 \pm 0.029$  \\
       \bottomrule
    \end{tabular}
    }
\end{table}

\subsection{Discrete Models}\label{app:ablation discrete}

\textbf{Ablation on Taylor-expansion Approximation.}
As shown in Table \ref{tab:ablation taylor}, using Taylor-expansion to approximate the ratio (Eq. \ref{eq: taylor expansion}) achieve comparable model performance while dramatically improve the sampling efficiency compared to calculating the ratio by definition, i.e. \xtgrad-Exact.

\textbf{Ablation on Monte Carlo Sample Size $N$.}
As shown in \cref{tab:ablation N}, increasing the Monte Carlo sample size improves performance, but further increases in $N$ beyond a certain point do not lead to additional gains.


\begin{table}[h]
\vspace{-10pt}
    \centering
      \caption{Ablation on Taylor-expansion Approximation. The performances are evaluated on 50 generated samples.
    }
    \label{tab:ablation taylor}
    \resizebox{0.8\textwidth}{!}{
    \begin{tabular}{lcccccc}
    \toprule
    \multirow{2}{*}{Method}  & \multicolumn{3}{c}{$\mathrm{N_r}^*=1$}  & \multicolumn{3}{c}{$\mathrm{N_r}^*=2$} 
         \\
        & MAE $\downarrow$ & TS $\downarrow$ & Time
        & MAE $\downarrow$ & TS $\downarrow$  & Time \\ 
    \midrule
       \xtgrad &  $0.24 \pm 0.76$ & $0.13 \pm 0.02$ & 2.4min & $0.02 \pm 0.14$ & $0.12 \pm 0.02$ & 3.5min \\
       \xtgrad-Exact & $0.00 \pm 0.00$ & $0.14 \pm 0.03$ & 356.9min &  $0.02 \pm 0.14$ & $0.13 \pm 0.02$ &  345.2min \\
       \bottomrule
    \end{tabular}
    }
    
\end{table}

\vspace{-20pt}

\begin{table}[h]
    \centering
    \caption{Ablation on Monte Carlo Sample Size $N$. The performances are evaluated on 200 generated samples.
    }
    \label{tab:ablation N}
      \resizebox{0.47\textwidth}{!}{
    \begin{tabular}{lcccc}
    \toprule
    \multirow{2}{*}{$N$}  & \multicolumn{2}{c}{$\mathrm{N_r}^*=2$}  & \multicolumn{2}{c}{$\mathrm{N_r}^*=5$} 
         \\
        & MAE $\downarrow$ & TS $\downarrow$ 
        & MAE $\downarrow$ & TS $\downarrow$  \\ 
    \midrule
       1   & $0.47 \pm 1.12$ & $0.12\pm0.02$ & $0.65 \pm 1.37$ & $0.12\pm0.02$ \\
       5 &  $0.30 \pm 0.97$ & $0.12\pm0.02$ &  $0.41 \pm 1.14 $ & $0.12\pm0.02$  \\
           10  &  $0.17 \pm 0.68$ & $0.12\pm0.02$  &  $\textbf{0.20} \pm \textbf{0.76}$ & $0.12\pm0.02$  \\
        20  &  $\textbf{0.09} \pm \textbf{0.46}$ & $0.12\pm0.02$ &  $0.26 \pm 0.91$ & $0.12\pm0.02$ \\
                40  &  $0.10 \pm 0.60$ & $0.12\pm0.02$ &  $0.29 \pm 1.01$ & $0.13\pm0.02$   \\
       \bottomrule
    \end{tabular}
    }
\end{table}

\begin{table}[H]
    \centering
      \caption{Ablation on Selection Choices in \ouralg. 
      \xtsampling and \xcleansampling are evaluated on 500 and 100 generated samples respectively. $t$ means temperature used for resampling weights. lower $t$ leads resampling closer to ranking. 
    }
    \label{tab:ablation selection}
    \resizebox{0.7\textwidth}{!}{
    \begin{tabular}{lcccccc}
    \toprule
    \multirow{2}{*}{Choices}  & \multicolumn{2}{c}{\xtsampling ($K=4$)}  & \multicolumn{2}{c}{\xcleansampling ($K=16$)} 
         \\
        & QED $\uparrow$ & TS $\downarrow$ 
        & QED $\uparrow$ & TS $\downarrow$ \\ 
    \midrule
      Ranking  &  $0.79 \pm 0.12$ & $0.12\pm0.02$  & $0.76\pm 0.12$ & $0.12\pm0.02$  \\
      \midrule
     Sampling ($t=0.1$)   & $0.71 \pm 0.17$ & $0.13\pm0.02$  &   $0.67 \pm 0.19$ & $0.12\pm0.02$  \\
          Sampling ($t=0.5$)   & $0.64 \pm 0.19$ & $0.12\pm0.02$ & $0.67 \pm 0.17$ & $0.11\pm0.02$  \\
       \bottomrule
    \end{tabular}
    }
    
\end{table}

\subsection{Selection Choices in  Alg.\ref{alg:framework}}
\label{app:rank resampling}
We compare two selection choices, i.e., ranking and resampling, in Alg.\ref{alg:framework} on small molecule generation with QED as the target. For both \xtsampling and \xcleansampling, selection by \textit{ranking} produces better guidance performance compared to selection by \textit{resampling} (Tab.~\ref{tab:ablation selection}).

\end{document}